\documentclass{article} % For LaTeX2e
\usepackage{iclr2019_conference,times}

\usepackage{amsmath,amsfonts,bm}

\def\eqref#1{equation~\ref{#1}}
\def\1{\bm{1}}

\def\bphi{{\bm{\phi}}}
\DeclareMathAlphabet{\mathsfit}{\encodingdefault}{\sfdefault}{m}{sl}
\SetMathAlphabet{\mathsfit}{bold}{\encodingdefault}{\sfdefault}{bx}{n}

\DeclareMathOperator*{\argmax}{arg\,max}

\usepackage{amsmath,amsfonts,bm}

\newcommand{\be}{\begin{equation}}
\newcommand{\ee}{\end{equation}}
\newcommand{\ba}{\begin{eqnarray}}
\newcommand{\ea}{\end{eqnarray}}
\newcommand{\nn}{\nonumber \\}

\newcommand{\x}{{\pmb{x}}}
\newcommand{\y}{{\pmb{y}}}

\newcommand{\z}{{\pmb{z}}}

\newcommand{\q}{{\pmb{q}}}

\newcommand{\btheta}{{\pmb{\theta}}}
\newcommand{\bzeta}{{\pmb{\zeta}}}

\newcommand{\brho}{{\pmb{\rho}}}

\newcommand{\U}{{\mathcal{U}}}

\newcommand{\E}{{\mathbb{E}}}

\def\p{\partial}

\usepackage{hyperref}
\usepackage{url}

\usepackage{float}
\usepackage{subfig}
\usepackage{graphicx}
\usepackage[makeroom]{cancel}

\title{Improved Gradient-Based Optimization Over Discrete Distributions}

\author{Evgeny Andriyash, Arash Vahdat \& Bill Macready \\
Quadrant.ai, D-Wave Systems Inc.\\
  Burnaby, BC, Canada \\
  \texttt{\{evgeny,arash,bill\}@quadrant.ai}
}

\iclrfinalcopy % Uncomment for camera-ready version, but NOT for submission.
\begin{document}

\maketitle

\begin{abstract}
%Discrete stochastic variables are an important component of modern ML. 
In many applications we seek to maximize an expectation with respect to a distribution over discrete variables. Estimating gradients of such objectives with respect to the distribution parameters is a challenging problem. We analyze existing solutions including finite-difference (FD) estimators and continuous relaxation (CR) estimators in terms of bias and variance. We show that the commonly used Gumbel-Softmax estimator is biased and propose a simple method to reduce it. We also derive a simpler piece-wise linear continuous relaxation that also possesses reduced bias. We demonstrate empirically that reduced bias leads to a better performance in variational inference and on binary optimization tasks.
\end{abstract}

\section{Introduction}

Discrete stochastic variables arise naturally for certain types of data, and distributions over discrete variables can be important components of probabilistic models. Practitioners developing discrete probabilistic models must often minimize an expectation of some function of discrete random variables with respect to the distribution parameters:
\be\label{eq:object_gen}
\min_{\bphi} {\cal L}[{\bphi}], \quad {\rm where} \:\: {\cal L}[{\bphi}] = \mathbb{E}_{q_\bphi}\bigl[f(\z)\bigr] = \sum_\z q_{\bphi}(\z) f(\z).
\ee
Here, $\z$ is a vector of discrete variables under a ${\bphi}$-parameterized distribution $q_{\bphi}(\z)$.  In variational inference, $f(\z)$ is the variational lower bound and $q_\bphi(\z)$ is the approximating posterior distribution. In the optimization context, $f(\z)$ is an objective function and $q_\bphi(\z)$ is typically a simple distribution that during training collapses to a single point $\z$ indicating a local minimum of $f(\z)$.

Eq.~(\ref{eq:object_gen}) is commonly minimized by gradient-based methods, which require estimating the gradient $\p_\bphi {\cal L}[\bphi]$. The two main approaches to this problem are score function estimators and pathwise derivative estimators(see \cite{schulman2015gradient} for an overview). Pathwise derivative estimators are applicable to cases in which the stochastic variables can be reparameterized as a function of other parameter-independent random variables, \textit{i.e.} $z = g_\bphi(\epsilon)$ where $\epsilon \sim p$ and $p$ is independent of $\bphi$.  The derivative is then estimated as 
\be\label{eq:rep_trick}
\p_\bphi {\cal L} = \mathbb{E}_{\epsilon \sim p}\bigl[  \p_\bphi f\bigl(z = g_\bphi(\epsilon)\bigr) \bigr].
\ee
This {\it reparameterization trick} (\cite{kingmaICLR15}, \cite{rezende2014stochastic}) in Eq.~(\ref{eq:rep_trick})
has low variance, and has been widely used for continuous variational inference. However, for discrete variables the cumulative distribution function (CDF) is discontinuous and reparameterization is not possible.
There have been several proposals to address this problem that we divide into finite-difference (FD) estimators, and continuous relaxation (CR) estimators. 

Score function esimators are based on the identity $\p_\bphi {\cal L}[\bphi] = \sum_\z q_{\bphi}(\z) \p_\bphi \log q_{\bphi}(\z) f(\z)$ and are applicable to any distribution $ q_{\bphi}(\z)$. This estimator has high variance and one typically has to use control variate technique to reduce it. In this paper we discuss the two recent proposals REBAR \cite{tucker2017rebar} and RELAX \cite{grathwohl2017backpropagation}.

We make several contributions:
\begin{enumerate}
\item In sections \ref{sec:RAM}, \ref{sec:ARGMAX}, and \ref{sec:ARM} we collect and summarize a number of recent FD estimators noting their bias-variance tradeoff and computational complexity.
\item In section \ref{sec:sampleRAM} we show how to improve the complexity of an unbiased FD estimator by trading decreased computation for increased variance.
\item We turn to CR estimators in section \ref{sec:CREstimators} and derive less-biased versions of these estimators in section \ref{sec:improvedCR}. In particular, we show how to decrease the bias of the popular Gumbel-Softmax \cite{jang2016categorical}, \cite{maddison2016concrete} relaxation for both binary (section \ref{sec:gumbel}) and categorical variables (section \ref{sec:categoricalGumbel}). We develop simple piecewise-linear estimators for both binary (section \ref{sec:pwl}) and categorical variables (section \ref{sec:categoricalPWL}).
\item
We discuss Score Function estimators in Section~\ref{sec:SF}. We derive a simple expression for REBAR and highlight it's relation to the ICR estimators (section~\ref{subsec:rebar}). We propose an alternative version of RELAX estimator, termed RELAX+, that enjoys lower complexity and increased performance (section~\ref{subsec:relax}).
\item Lastly, in section \ref{sec:exp} we provide empirical evidence that these improved estimators allow for faster optimization when training variational autoencoders and when optimizing a continuous relaxation of a combinatorial optimization problem.
\end{enumerate}

\subsection{Related work}

The most generic approximator of $\p_\bphi {\cal L}[\bphi]$ is the score function (SF) estimator (a.k.a. REINFORCE \cite{williams1992simple}, \cite{glynn1990likelihood}). SF suffers from high variance and many remedies have been proposed to reduce this variance  (\cite{mnih2014neural, gregorICML14, gu2015muprop, mnih2016variational, tucker2017rebar, grathwohl2017backpropagation}). Unbiased estimators that require multiple function evaluations have also been proposed \cite{tokui2017evaluating}, \cite{aueb2015local}, \cite{lorberbom2018direct}, \cite{yin2018arm}. These estimators have lower variance but can be computationally demanding. CR estimators often trade bias for variance and previous CR proposals include straight-through estimators (\cite{bengio2013estimating, raiko2014techniques}), the Gumbel-Softmax estimator (\cite{jang2016categorical, maddison2016concrete}), and overlapping smoothing
(\cite{VahdatMBKA18, vahdat2018dvae, rolfe2016discrete}).

\section{Finite-Difference Estimators}
\label{sec:FDEstimators}

\subsection{The Reparameterization and Marginalization Estimator}
\label{sec:RAM}

We begin a review of FD approaches with the reparameterization and marginalization (RAM) estimator \cite{tokui2017evaluating}. For a single binary stochastic variable the expectation in Eq.~(\ref{eq:object_gen}) is enumerated as ${\cal L} = \sum_z q_\phi(z) f(z) = q f(1) + (1-q) f(0)$, where $q_\phi \equiv q_\phi(z=1)$. The gradient is: 
\be\label{eq:grad_1}
\p_\phi {\cal L} = \p_\bphi q_\phi (f(1) - f(0)) = \p_\phi l_\phi  \: q (1 - q) (f(1) - f(0)),
\ee
where $q_\phi = \sigma(l_\phi) = (1 + e^{-l_\phi})^{-1}$ and $l_\phi = \textrm{logit}(q_\phi)$. This derivative involves two function evaluations and contains a finite-difference  of $f(z)$. Eq.~(\ref{eq:grad_1}) is an unbiased zero-variance estimate since the summation over $z$ is done explicitly. The generalization to $M$ factorially-distributed random variables $q_\bphi(\z) = \prod_i q_{\bphi,i}(z_i)$ for $1\le i \le M$ is
\be\label{eq:grad_fact}
\p_\bphi {\cal L} = \sum_i \p_\bphi q_{\bphi,i} \sum_{\z_{\setminus i}} q_\bphi(\z_{\setminus i}) (f(z_i=1, \z_{\setminus i}) - f(z_i=0, \z_{\setminus i})),
\ee
where again the summation over $z_i$ is performed exactly and $ q_{\bphi,i} \equiv q_{\bphi,i}(z_i=1)$. The summation over $\z_{\setminus i}$ can be estimated with a single sample but the derivative requires $M+1$ function evaluations. This limits the applicability of RAM to moderately-sized models. Note that both $f(z_i=0, \z_{\setminus i})$ and $f(z_i=1, \z_{\setminus i})$ are evaluated at the same $\z_{\setminus i}$ which leads to a low-variance estimator. For hierarchical $q_\bphi(\z) = \prod_i q_{\bphi,i}(z_i | \z_{<i})$ the derivative takes the form:
\be\label{eq:grad_bayes}
\p_{\bphi} {\cal L} = \sum_i  \sum_{\z_{<i}} q_{\bphi}(\z_{<i}) \p_{\bphi} q_{\bphi,i}  \left[ \sum_{\z_{>i}} q_\bphi(\z_{>i}|1, \z_{<i}) f(\z_{>i},1, \z_{<i}) - \sum_{\z_{>i}} q_\bphi(\z_{>i}|0, \z_{<i}) f(\z_{>i},0, \z_{<i}) \right]. 
\ee
where $q_{\bphi,i} \equiv q_{\bphi,i}(1|\z_{<i})$. The key insight of \cite{tokui2017evaluating} is to use common random variates $\z_{>i}$ when sampling from $q_\bphi(\z_{>i}|1, \z_{<i})$ and $q_\bphi(\z_{>i}|0, \z_{<i})$. This reduces both the variance and the computational cost. 
%For example, for binary variables the samples can be obtained by sampling uniform variables $\epsilon_{>i}$ and using reparameterization $\z_{>i}^{a} = \Theta(q_\phi(\z_{>i}|a, \z_{<i})- \epsilon_{>i}), a=0, 1$. 
\cite{tokui2017evaluating} show that this estimator is optimal because it exactly sums over the binary variables whose probability distribution is being differentiated.

A RAM estimator can also be constructed for categorical variables. For a single one-hot encoded categorical variable $y = (y^0,...y^{A-1}), y^a \in \{0,1\}, \sum_a y^a = 1$, the derivative of Eq.~(\ref{eq:object_gen}) is 
\be\label{eq:grad_cat}
\p_{\bphi} {\cal L} = \p_{\bphi}  \sum_{y} q_\bphi(y) f(y) = \sum_a  (\p_{\bphi} l^a_\bphi) \sum_{b} q^a_\bphi q^{b}_\bphi (f^a - f^{b}),
\ee
where $f^a = f(y^a=1)$, and $q^a_\bphi = e^{l^a_\bphi} / \sum_{b} e^{l^{b}_\bphi}$.\footnote{This expression depends on the function values $f^a$ at all $A$ vertices of the simplex. However, we can rewrite Eq.~(\ref{eq:grad_cat}) more suggestively as
\begin{equation*}
\p_{\bphi} {\cal L} =  \sum_a    (\p_{\bphi} l^a_\bphi) q^a_\bphi \biggl[ f^a - \sum_{b} q^{b}_\bphi  f^{b} \biggr] =  \E_{a \sim q_\bphi^a}\bigl[ (\p_{\bphi} l^a_\bphi) \bigl(f^a -  \E_{b \sim q_\bphi^b}[f^{b}]\bigr)\bigr].
\end{equation*}
Each expectation can be computed by sampling thereby trading off computational effort for increased variance.} The generalization to many categorical variables proceeds as for binary variables. For example, the derivative of a factorial categorical distribution over $\y = \{y_i^a \mid 0\le a<A, 1\le i\le M\}$ is
\be\label{eq:grad_cat_fact}
\p_{\bphi} {\cal L} = \sum_i  \sum_{\y_{\setminus i}} q_\bphi(\y_{\setminus i})    \sum_{a,b}  (\p_{\bphi} l^a_{\bphi,i})  q^a_{\bphi,i} q^{b}_{\bphi,i} \bigl(f(y^a_i=1, \y_{\setminus i}) - f(y^{b}_i=1, \y_{\setminus i})\bigr).
\ee
This derivative can again be estimated with a single sample but requires $M A$ function evaluations. Since RAM is unbiased and has the minimal variance (due to the explicit summation over the differentiated variable), we use it as a baseline to evaluate computationally cheaper alternatives.

\subsection{Sampled Reparameterization and Marginalization}
\label{sec:sampleRAM}

We propose a modification to RAM that allows us to trade decreased computational cost for increased variance. For binary variables, $\p_\bphi q_{\bphi,i} = q_{\bphi,i} (1 - q_{\bphi,i}) \p_\bphi l_{\bphi,i}$ where $q_{\bphi,i} = \sigma(l_{\bphi,i})$, so that each term in Eq.~(\ref{eq:grad_fact}) or (\ref{eq:grad_bayes}) is proportional to $q_{\bphi,i} (1 - q_{\bphi,i})$. In many applications, we observe that the distribution of a large number of variables ($q_{\bphi,i}$) are drawn to 0 or 1 early during optimization. Such variables have negligible contribution to the full derivative. We exploit this observation to reduce the computational cost by including variable $z_i$ in the full gradient with probability $p_i = 4 q_{\bphi,i} (1-q_{\bphi,i}) / \beta$, where $\beta$ is an adjustable hyperparameter. This means that we replace the derivative in Eq.~(\ref{eq:grad_fact}) with 
\be\label{eq:grad_fact_sampled}
\p_\bphi {\cal L} = E_{\xi \sim p} \Bigl[ \sum_i (\p_\bphi l_{\bphi,i}) \frac{\beta \xi_i}{4} \sum_{\z_{\setminus i}} q_\bphi(\z_{\setminus i}) \bigl(f(z_i=1, \z_{\setminus i}) - f(z_i=0, \z_{\setminus i})\bigr) \Bigr],
\ee
where $\xi_i \in \{0,1\}$ are Bernoulli variables with probabilities $p_i$ indicating whether $z_i$ is included or not. We evaluate Eq.~(\ref{eq:grad_fact_sampled}) by sampling $\xi$ and only then evaluating non-zero terms.  In Section \ref{sec:exp} we show that in the context of variational inference on MNIST the number of function evaluations is reduced by an order of magnitude while still allowing for effective optimization. As always, this computational saving comes at the cost of increased gradient variance which slows training. 

In the categorical case, each term in Eq.~(\ref{eq:grad_cat_fact}) is accompanied by $q^a_{\bphi,i} q^b_{\bphi,i}$ that can be used to assign importance to the $(a, b)$ edge of the simplex for variable $y_i$. The computational cost $M A$ can be reduced by keeping each term with probability $p_i^{a,b} = 4 q_{\bphi,i}^a q_{\bphi,i}^b / \beta$. Alternatively, one can only evaluate derivatives for a subset of categorical variables selecting them with probability $p_i \sim \sum_{a \ne b} q^a_{\bphi,i} q^b_{\bphi,i} = [1 - \sum_a (q^a_{\bphi,i})^2]$.

\subsection{ARGMAX}
\label{sec:ARGMAX}

Recently, \cite{lorberbom2018direct} proposed another FD estimator that we refer to as ARGMAX. For a single variable, ARGMAX relies on an identity that approximates $\p_\bphi {\cal L} = \p_\bphi q_\bphi [f(1) - f(0)] = (\p_\bphi l_\bphi) q_\bphi [1-q_\bphi ] [f(1) - f(0)]$. With $q_\bphi = \sigma(l_\bphi)$, \cite{lorberbom2018direct} show that
\be\label{eq:biased_grad_eps}
\sigma(l_\bphi) [1 - \sigma(l_\bphi)]  [f(1) - f(0)]  = \lim_{\epsilon \to 0} \E_{\rho} \biggl[\frac{\Theta\bigl(\epsilon f(1) + l_\bphi + \sigma^{-1}(\rho) \bigr)  - \Theta\bigl(\epsilon f(0) + l_\bphi + \sigma^{-1}(\rho)  \bigr)}{\epsilon}\biggr]
\ee
where $\Theta$ is the Heaviside step function and $\rho \sim \U[0,1]$.\footnote{We motivate this identity by noting that the non-zero contribution comes from the region $\sigma( - \epsilon f(1)-l_\bphi) \le \rho \le \sigma(- \epsilon f(0) -l_\bphi)$ assuming $f(1) > f(0)$. For small $\epsilon$, samples land within this region with probability $p \sim \sigma(l_\bphi) (1-\sigma(l_\bphi)) \epsilon [f(1) - f(0)]$ giving rise to the identity. The variance of Eq.~(\ref{eq:biased_grad_eps}) is $\mathbb{V}\textrm{ar}\bigl(\textrm{Ber}(p)\bigr)/\epsilon^2 \propto \sigma(l_\bphi) (1-\sigma(l_\bphi)) [f(1) - f(0)]/\epsilon$ to leading order in $1/\epsilon$.} \cite{lorberbom2018direct} approximate the right side of Eq.~(\ref{eq:biased_grad_eps}) by sampling $\rho$ and evaluating at finite $\epsilon$ which introduces bias and variance. Decreasing $\epsilon$ decreases bias but increases variance. 

In the categorical case the gradient of Eq.~(\ref{eq:grad_cat}) can correspondingly be written as
\be
\p_{\bphi} {\cal L} = \sum_a  \p_{\bphi} l_{\bphi}^a \lim_{\epsilon \to 0} \frac{1}{\epsilon}  \E_\rho \left[  \bigl[a=\argmax_b(\epsilon f^b + l^b_\bphi + \gamma^b)\bigr] - \bigl[a=\argmax_b( l^b_\bphi + \gamma^b) \bigr ] \right],
\ee
where $\gamma_b = -\log( - \log(\rho^b))$ are Gumbel variables and $[pred]$ is the indicator function (1 if $pred$ is true and 0 otherwise). This derivative requires $A$ function evaluations, similar to RAM. \cite{lorberbom2018direct} extend the single variable result to multivariate distributions similar to Eqs.~(\ref{eq:grad_fact}) and (\ref{eq:grad_bayes}). 
ARGMAX has the same computational complexity as RAM, but is biased and has higher variance than RAM. Thus, it is suboptimal to Eqs.~(\ref{eq:grad_fact}) and (\ref{eq:grad_bayes}) and for this reason we do not perform experiments with ARGMAX.

\subsection{The Augment-REINFORCE-Merge Estimator}
\label{sec:ARM}

FD estimators are computationally expensive and require multiple function evaluations per gradient. A notable exception is the Augment-REINFORCE-Merge (ARM) estimator introduced in \cite{yin2018arm}. ARM provides an unbiased estimate using only two function evaluations for the factorized multivariate distribution, regardless of the number of variables. For a single binary variable, the ARM derivative is given by 
\begin{gather}\label{eq:grad_arm_bin_1}
\p_\phi {\cal L} = \p_\phi l \: q_\phi (1-q_\phi) (f(1) - f(0)) = \p_\phi l \: \E_{\rho \sim \U[0,1]}  [(f(z^{(2)}) - f(z^{(1)}) )(\rho - 0.5)], \nn
 z^{(1)} = \Theta(q_\phi - \rho) \quad z^{(2)} = \Theta(\rho - 1 + q_\phi).
\end{gather}
The expectation is approximated with a single sample $\rho$. This estimator has a significantly lower variance than REINFORCE since the expectation contains the difference $f(z^{(2)}) - f(z^{(1)})$ rather then the function itself. For multivariate $f$ and factorial $q_\bphi(\z)$ the derivative is 
\be\label{eq:grad_arm_bin_mult}
 \p_\bphi {\cal L} = \sum_i \p_\bphi l_{\bphi,i} \: \E_{\brho \sim \U[0,1]}  [(f(z_i^{(2)}, \z_{/i}^{(1)}) - f(z_i^{(1)} , \z_{/i}^{(1)}) )(\rho_i - 0.5)],
\ee
where $z_i^{(1)} = \Theta(q_{\bphi,i} - \rho_i)$ and $z_i^{(2)} = \Theta(\rho_i - 1 + q_{\bphi,i})$. \cite{yin2018arm} observed that one can replace $f(z_i^{(2)}, \z_{/i}^{(1)}) \to f(z_i^{(2)}, \z_{/i}^{(2)})$ without changing the expectation which allows evaluation by a single sample  and two function evaluations $f(\z^{(1)}),f(\z^{(2)})$ regardless of $M$:
\be\label{eq:grad_arm_bin}
\p_\bphi {\cal L} = \sum_i \p_\bphi l_{\bphi,i} \: \E_{\brho \sim \U[0,1]}  [(f(\z^{(2)}) - f(\z^{(1)}) )(\rho_i - 0.5)].
\ee
However, this change comes at the cost of higher variance of the expectation\footnote{Denoting $g^{(a)}(\z_{/i}) = f(z_i^{(a)}, \z_{/i}), a=1,2$, we can write
\ba\label{eq:var_covar}
&& {\rm VAR}_{\z_{/i}}[g^{(2)}(\z_{/i})-g^{(1)}(\z_{/i})] = {\rm VAR}_{\z_{/i}}[g^{(2)}(\z_{/i})] + {\rm VAR}_{\z_{/i}}[g^{(1)}(\z_{/i})]  - 2 {\rm COV}_{\z_{/i}}[g^{(1)}(\z_{/i}), g^{(2)}(\z_{/i})],\nn
&& {\rm VAR}_{\z_{/i}, \z_{/i}'}[g^{(2)}(\z_{/i})-g^{(1)}(\z_{/i}')] = {\rm VAR}_{\z_{/i}}[g^{(2)}(\z_{/i})] + {\rm VAR}_{\z_{/i}}[g^{(1)}(\z_{/i})].
\ea
If the functions $g^{(a)}(\z_{/i})$ are highly correlated the ARM estimator will have a much higher variance than RAM.}. Thus, while ARM estimator provides a low-variance gradient estimate for a single variable, it introduces high variance for multivariate functions compared to RAM. The ARM estimator has straightforward extensions to hierarchical $q(\z)$ (similar to Eq.~(\ref{eq:grad_bayes})) and to categorical variables \cite{yin2018arm}. 

\section{Continuous Relaxation Estimators}
\label{sec:CREstimators}

Unlike FD estimators that use multiple function evaluations to approximate the gradient, continuous relaxation estimators extend the reparameterization trick to discrete variables by approximating them with continuous variables: $z \to \zeta  = \zeta_\beta(\rho, q_{\bphi})$, where $\beta$ is a parameter that controls the approximation, $\rho \in \U[0,1]$ is uniform random variable and $q_{\bphi}$ are parameterized probabilities. We use the same symbol $\zeta$ to denote the random variable and its reparameterization by $\rho$. The objective function ${\cal L}[\bphi] = \sum_\z q_\bphi(\z) f(\z)$ is replaced with $\tilde {\cal L}[\bphi] = \E_{\brho}\left[ f\bigl(\bzeta(\rho, q_\bphi)\bigr)\right]$
%
%\be\label{eq:object_relaxed}
%{\cal L}[\bphi] = \sum_\z q_\bphi(\z) f(\z) \to \tilde {\cal L}[\bphi] = \E_{\brho}\left[ f\bigl(\bzeta= g_\phi^\beta(\rho)\bigr)\right]
%\ee
%
and its gradients are computed using the chain rule
\be\label{eq:grad_relaxed}
\p_\bphi \tilde {\cal L}[\bphi] =  \E_{\brho}\left[\sum_i \p_{\zeta_i} f(\bzeta) (\p_{q_i} \zeta_i) (\p_\bphi q_i) \Bigr|_{\substack{\zeta_i(\rho_i,  q_{\bphi,i}(1| \bzeta_{<i})} } \right]
\ee
%
%\E_{\brho}\left[\sum_i \p_{\zeta_i} f(\bzeta) \p_\bphi \zeta_i \Bigr|_{\zeta_i(\rho_i, q_{\bphi,i}(1| \bzeta_{<i}))} \right] 
These gradients can be computed efficiently by automatic differentiation libraries. However, because the objective function is changed, CR estimators in Eq.~(\ref{eq:grad_relaxed}) are biased. Nevertheless, in practice the bias is often small enough to allow for effective optimization. 

\subsection{The Gumbel-Softmax Estimator}
\label{sec:gumbel}

\begin{figure}[t]
    \centering
    \vspace{-1,5cm}
     \subfloat[Gumbel-Softmax]{  \includegraphics[scale=0.29]{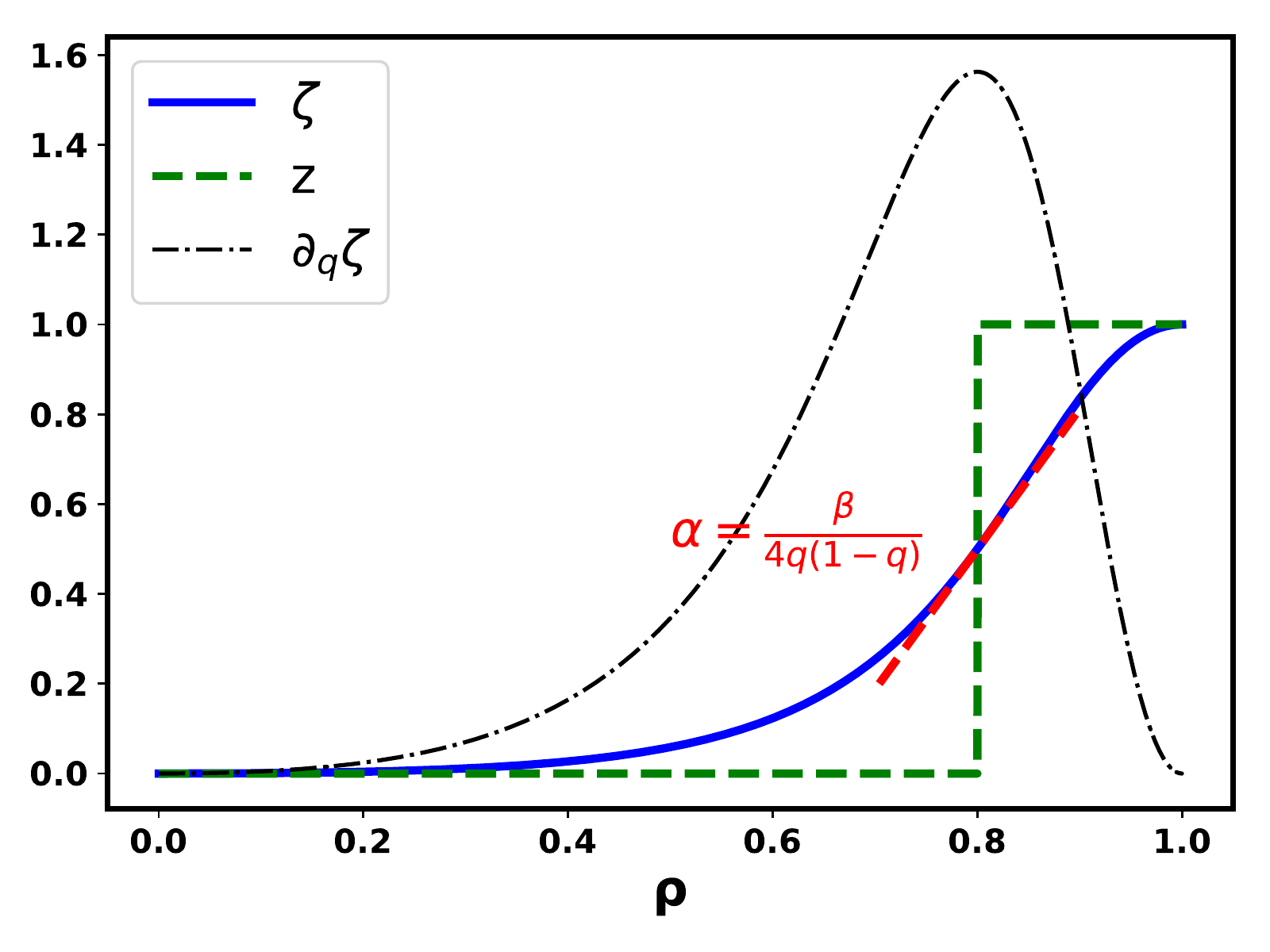}} 
          \subfloat[Piecewise Linear]{ \includegraphics[scale=0.29]{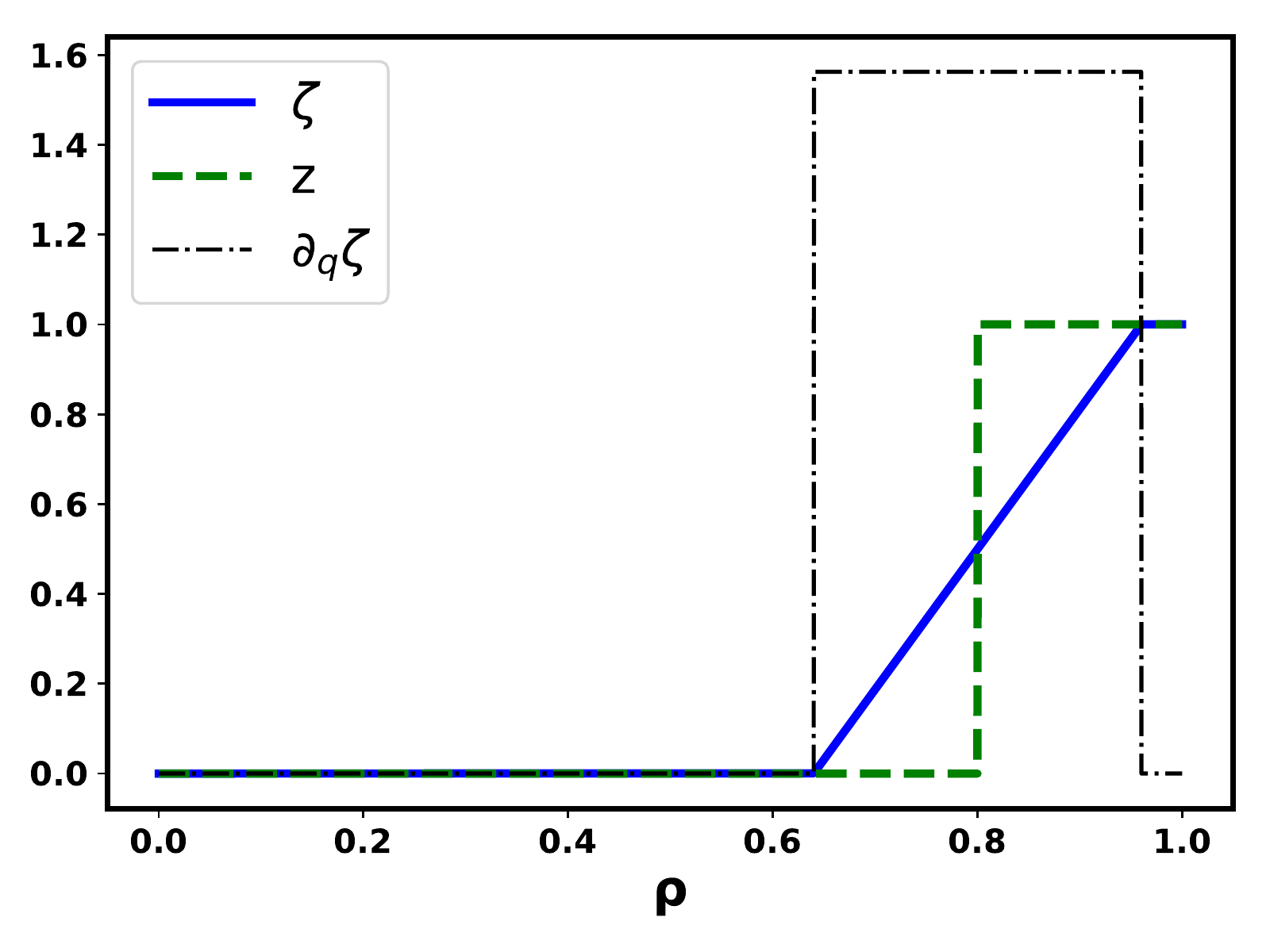}}
               \subfloat[$\p_q \zeta$]{ \includegraphics[scale=0.29]{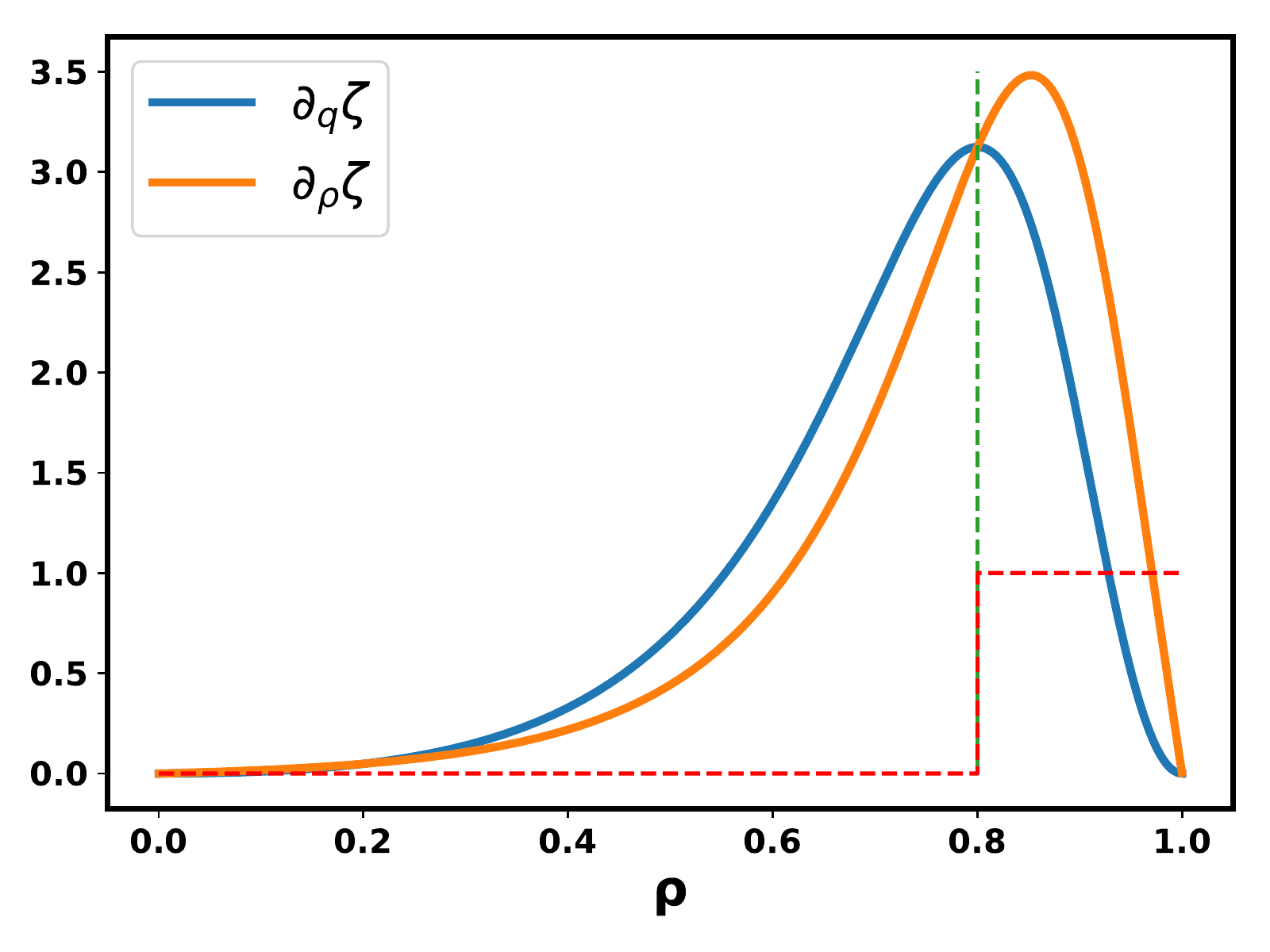}}
\caption{Continuous relaxations at $\beta=2$ to a Bernoulli random variable Ber$(q_\phi=0.8)$: (a) shows the GSM approximation and (b) shows a piece-wise linear approximation to the discontinuous CDF. (c) shows the effect of a modification to GSM that reduces its bias.} \label{fig:cr}
\end{figure}

Gumbel-Softmax (GSM), introduced in \cite{jang2016categorical,maddison2016concrete}\footnote{More precisely, \cite{maddison2016concrete} considered problems in variational inference where $f(\z)$ is relaxed by replacing a generative distribution with its continuous relaxation. In contrast, \cite{jang2016categorical} directly relaxed discrete variables $z \to \zeta$ without changing the objective, thus replacing $f(\z) \to f(\bzeta)$. This does not produce a consistent objective for a probabilistic model, but as \cite{jang2016categorical} have shown, works well in practice. We work with generic $f(\z)$ and thus only consider the approach of \cite{jang2016categorical}.}, is the most popular CR estimator.  For binary variables this relaxation takes a simple form:
\be\label{eq:gsm_relax}
\zeta_i(\rho_i, q_i) = \sigma\left(\beta \bigl[ \sigma^{-1}\bigl(q_i\bigr) + \sigma^{-1}(\rho_i)\bigr] \right),
\ee
where $q_i \equiv q_{\phi,i}(1|\bzeta_{<i})$. This relaxation and its derivative $\p_q\zeta$ are shown in Fig.~\ref{fig:cr}(a). $\beta$ controls the sharpness of the relaxation and tunes the trade-off between the bias (closeness of $\zeta$ to $z$) and variance of $\p_q \zeta$. We note that the slope of the relaxed $\zeta(\rho)$ at  $\zeta=1/2$ is $\alpha \equiv \beta /[4 q (1-q)]$ and thus becomes large when $q$ approaches $0$ or $1$. 

\subsection{Improved Continuous Relaxation Estimators}
\label{sec:improvedCR}

In this section, we analyze the bias introduced by Eq.~(\ref{eq:grad_relaxed}) and propose a simple method to reduce it. The bias of $\E_{\brho}\left[ \p_{\zeta_i} f(\bzeta) \p_\bphi \zeta_i \right]$ has two sources: (a) the relaxation of  $\zeta_j$ for $j \ne i$ and (b) the relaxation of $\zeta_i$. To characterize the latter bias, we start with a single binary variable and write the gradient as the following integral:
\be\label{eq:grad_1_int}
\p_\phi {\cal L} = \p_\phi q_\phi \bigl(f(1) - f(0)\bigr) = \p_\phi q_\phi \int_0^1 d \xi \, \p_\xi f(\xi) = \p_\phi q_\phi \int_0^1 d \rho \, \frac{\p \xi}{\p \rho} \p_\xi f(\xi).
\ee
Here, $\xi(\rho)$ is \emph{any} continuous function satisfying $\xi(0)=0$ and $\xi(1)=1$. For a non-decreasing function $\xi(\rho)$, we can view $\xi$ as a random variable with inverse CDF $\xi(\rho)$ and $\rho$ as a uniform random variable $\rho \in \U[0,1]$. With this interpretation the gradient can be written as an expectation
\be\label{eq:deriv_1}
\p_\phi {\cal L} = \p_\phi q_\phi \, \E_{\rho}\left[ \frac{\p \xi}{\p \rho}  \p_\xi f(\xi)\right],
\ee
which can be estimated by sampling. If $\partial_\xi f(\xi)$ does not vary significantly in the interval $\xi \in [0, 1]$ then the variance of this estimate is controlled by ${\rm var}(\p \xi / \p \rho) = \int_0^1 d \rho \, \left(  \p \xi/\p \rho \right)^2 - 1$. Thus, the more non-linear $\xi(\rho)$ is, the higher will be the variance of estimate Eq.~(\ref{eq:deriv_1}).
This idea can be extended to factorial $q_\bphi(\z) = \prod_{i=1}^M q_{\bphi,i}(z_i)$ as in Eq.~(\ref{eq:grad_fact}):
\be\label{eq:grad_fact_int}
\p_{\bphi} {\cal L} = \sum_i \p_\bphi q_{\bphi,i} \sum_{\z_{\setminus i}} q_\bphi(\z_{\setminus i}) \int d \rho_i \frac{\p \xi}{\p \rho_i}  \p_{\xi_i} f(\xi, \z_{\setminus i}) .
\ee
At this point there is no relationship between $\xi_i$ and $z_i$, and Eq.~(\ref{eq:grad_fact_int}) is just a higher variance version of Eq.~(\ref{eq:grad_fact}). However, if we relax $\z_{\setminus i} \to \bzeta_{\setminus i}(\rho, q_\bphi)$ and choose $\xi_i(\rho_i) = \zeta_i(\rho_i,   q_{\bphi,i})$  we obtain a \emph{biased} estimator
\be\label{eq:grad_fact_unbiased}
\p_{\bphi} {\cal L} \approx   \sum_i \E_{\brho} \left[   \p_{\zeta_i} f(\bzeta)   \frac{\p \zeta_i}{\p \rho_i}  \p_\bphi q_{\bphi,i} \right],
\ee
where the bias comes from the deviation of $\bzeta_{\setminus i}$ from $\z_{\setminus i}$ in the function evaluations. In other words, Eq.~(\ref{eq:grad_fact_unbiased}) uses an unbiased form for the differentiated variable $\zeta_i$ and the only bias comes from relaxing the remaining variables $\bzeta_{\setminus i}$. 

The variance of each term is controlled  by ${\rm var}\!\left(\frac{\p \zeta_i}{\p \rho_i}\right)$ and is reduced by making $\zeta_i(\rho_i,q_{\bphi,i})$ more linear. The bias is reduced by making $\bzeta$ closer $\z$. Varying $\zeta_i(\rho_i,q_{\bphi,i})$ between linear (low variance) and step function (low bias) allows for a controllable trade-off.  We call Eq.~(\ref{eq:grad_fact_unbiased}) the improved continuous relaxation (ICR) estimator. 
The original CR estimator of Eq.~(\ref{eq:grad_relaxed}) for factorial $q_\bphi$ has the form 
\be\label{eq:grad_relaxed_fact}
\p_\bphi \tilde {\cal L}[\bphi] =\sum_i \E_{\brho} \left[   \p_{\zeta_i} f(\bzeta)   \frac{\p \zeta_i}{\p q_i}  \p_\bphi q_{\bphi,i} \right],
\ee
and comparing this to the ICR estimate in Eq.~(\ref{eq:grad_fact_unbiased}), we see that CR can be transformed into ICR by the replacing $ \p_{q_i} \zeta_i$ with $\p_{\rho_i} \zeta_i$. This change can be simply implemented using TensorFlow's ${\rm stop\_gradient} \equiv {\bf sg}$ notation by replacing
\be\label{eq:qrho_trick}
\zeta_i(\rho_i, q_i ) \quad \text{with} \quad \zeta_i\bigl(\rho_i + q_i - {\bf sg}(q_i), {\bf sg}(q_i) \bigr)
\ee
in Eq.~(\ref{eq:grad_relaxed}). We emphasize that the ICR estimator in Eq.~(\ref{eq:grad_fact_unbiased}) is less biased than the direct CR estimator because, in the case of a single variable, ICR is unbiased while CR is not. Further, this decrease in bias is not accompanied by an increase in variance. We show in Appendix~\ref{app:icr_bayes} that similar benefits are obtained for hierarchical $q_\bphi$. Finally, we note a conceptual relationship between sampled RAM and ICR: both estimators evaluate the gradient only through a subset of variables $z_i$. Sampled RAM chooses this subset explicitly and evaluates the gradients with FD while ICR samples relaxed variables where only a subset of them will possess non-negligible gradients.

\subsection{Piece-wise Linear Relaxation}
\label{sec:pwl}
Inspired by ICR and a better understanding of bias-variance trade-off, we propose a piece-wise linear relaxation (PWL) depicted in  Fig.~\ref{fig:cr}(b). The linear part is centered at $\rho=1-q$ so that the corresponding binary variable is obtained by $z = {\rm round}(\zeta)$. The slope is given by $\alpha = \beta/[4 q (1-q)]$ similar to the Gumbel-Softmax slope \footnote{More generally, the slope of the linear part $\alpha$ can be chosen arbitrarily as long as $\alpha \ge 0.5/{\min(q, 1-q)}$ so that both $\zeta=0$ and $\zeta=1$ have non-zero probability.}. The explicit expression for PWL smoothing is
\be\label{eq:pwl_relax}
\zeta(\rho, q) = \bigl[ 0.5 + \alpha (\rho - (1-q)) \bigr]_0^1,
\ee
where $\left[ x \right]_0^1 \equiv \min\left(1, \max\left[ 0, x \right] \right)$ is the hard sigmoid function (note that $\partial_q \alpha=0$). This relaxation has several attractive properties. Firstly, we have $\p_{q_i} \zeta_i =  \p_{\rho_i} \zeta_i$, which means that the CR and ICR estimators coincide for PWL. Secondly, PWL has easily interpretable expressions for bias and variance. In the case of a function of a single variable the variance is given by ${\rm var}(\p_{\rho} \zeta) = \alpha - 1$ while the bias in computing expectation $\sum_z f(z)$ over the relaxed distribution is $\int d \rho f(\zeta(\rho)) - \sum_z f(z) = \{ \int_0^1 d x f(x) - [f(0)+f(1)]/2 \}/\alpha$. This clearly shows that $\alpha$ trades bias for variance.
%$p(\zeta \geq 0.5) = q$.

Finally, the PWL relaxation defined in Eq.~(\ref{eq:pwl_relax}) can be considered as the inverse CDF of the PDF
\be\label{eq:pdf_pwl}
q(\zeta) = \left(1-q-\frac{\epsilon}{2}\right) \delta(\zeta) + \left(q-\frac{\epsilon}{2}\right) \delta(\zeta - 1)  + \epsilon\ \U[0,1], \ \ \text{where} \ \ \epsilon = [4 q (1-q)]/\beta.
\ee
Eq.~(\ref{eq:pdf_pwl}) is a mixture of two delta distributions 
centered at 0 and 1, and a uniform distribution defined in the interval $[0, 1]$.

\subsection{Categorical Piece-wise Linear Relaxation}
\label{sec:categoricalPWL}

We now extend ICR estimators to categorical variables. For a single categorical variable we apply the integral representation to each edge $(a,b)$ of the simplex in Eq.~(\ref{eq:grad_cat}) and relax this pair of variables using PWL: $y \to y^{a,b} = \{y^a = \left[0.5+\alpha^{a,b}\bigl(\rho^{a,b} - q^b/(q^a + q^b)\bigr)\right]_0^1, y^b = 1 - y^a, y^{c \ne a,b}=0\}$ where  $\rho^{a,b} \sim \U[0,1]$ and $\alpha^{a,b}$ is the slope. We replace the summation over the edges of the simplex by sampling one edge at a time with probability $p^{a,b} = (q^a + q^b)/(A-1)$. Details are found in Appendix \ref{app:pwl_cat} and we provide the final result:
\be\label{eq:grad_1hot_int_sample}
{\cal L} = \E_{(a,b) \sim p^{a,b}} \left[ \E_{\rho \in \U[0,1]} \left[  f(\tilde y^{a,b}) \right] \right],
\ee
where $\tilde y^{a,b}$ has the same value as $y^{a,b}$ but has the gradient scaled by $\gamma^{a,b} = (A-1)(q^a + q^b)$.\footnote{In Tensorflow notation $\tilde y^{a,b} \equiv {\bf sg}(y^{a,b}) + {\bf sg}(\gamma^{a,b}) ( y^{a,b} - {\bf sg}(y^{a,b}))$.} The probabilities of edge selection and the scale factor are chosen to give correct values for the objective and its gradient. Extension of this categorical PWL estimator to multivariate distributions is straightforward; For example, for factorial distributions one relaxes each categorical variable $y_{i}$ by sampling an edge with probability $p^{a,b}_i = (q_i^a + q_i^b)(A-1)$ and uses a single function evaluation.  The resulting gradient is unbiased for a single variable and introduces a bias in the multivariate case which makes it an ICR estimator.

\subsection{Improved Categorical Gumbel-Softmax Estimator}
\label{sec:categoricalGumbel}

The sampling done in the categorical PWL estimator leads to increased variance of the gradients. As an alternative, we suggest an improved version of the categorical Gumbel-Softmax estimator.  Recall that the Gumbel-Softmax estimator for the categorical case has the form:
 \ba\label{eq:gsm_1hot}
&& \zeta^a(\rho; \q) = {\rm softmax} (\beta \left( \log q^a  - \log \rho^a\right)) \nn
&&  \rho^a = \frac{\log u^a}{\sum_b \log u^b}, \: u^b \sim \U[0,1].
\ea
We propose a new version of this estimator by applying the same trick as in Eq.~(\ref{eq:qrho_trick}):
\be\label{eq:gsm_trick_bin}
\zeta(\rho,\q) \to  \zeta(\rho - \q + {\bf sg}(q), {\bf sg}(q)).
\ee
This estimator remains biased but we empirically demonstrate that its bias is reduced.

\section{Score Function estimators}\label{sec:SF}

The most generic estimator of $\p_\bphi {\cal L}[\bphi]$ is the score function (SF) estimator (a.k.a. REINFORCE \cite{williams1992simple}, \cite{glynn1990likelihood}). This estimator is unbiased but posesses high variance and in what follows we discuss two recent proposals on reducing it.

\subsection{REBAR}\label{subsec:rebar}

In this section, we derive a simple expression for REBAR gradients \cite{tucker2017rebar} and describe its connection to ICR estimators. REBAR gradients of the objective (\ref{eq:object_gen}) can be written as:
\be\label{eq:grad_rebar}
\partial_\phi {\cal L} = \E_{q_\bphi(\z) q_\bphi(\bzeta | \z)} \left[\partial_\phi \log q_\bphi(\z)  \left( f(\z) - f(\bzeta) \right) \right] + \partial_\phi \E_{q_\bphi(\bzeta)} \left[ f(\bzeta)  \right] - \partial_\phi\E_{q_{{\bf sg}(\bphi)}(\z) q_\bphi(\bzeta | \z)} \left[f(\bzeta) \right], \nonumber
\ee
where ${\bf sg}(\bphi)$ denotes ``stop gradient'', and distributions $q_\bphi(\bzeta)$ and $ q_\bphi(\bzeta | \z)$ are reparameterizable. Using explicit reparameterizations for $\bzeta(\brho, q_\bphi)$ and $\bzeta({\bf u}, q_\bphi | \z)$ with ${\bf u} \in \U[0,1]$, we can rewrite the gradient as
\be\label{eq:grad_rebar1}
\partial_\phi {\cal L} = \E_{q_\bphi(\z) q_\bphi(\bzeta | \z)} \left[\partial_\phi \log q_\bphi(\z)  \left( f(\z) - f(\bzeta) \right) \right] + \E_{\brho} \left[ \partial_\bzeta f(\bzeta) \partial_\bphi \bzeta(\brho, q_\bphi) \right] - \E_{q_\bphi(\z),{\bf u}} \left[ \partial_\bzeta f(\bzeta) \partial_\bphi \bzeta({\bf u}, q_\bphi|\z)\right].
\ee
Focusing on binary variables, the reparameterization for the conditional distribution  has the form $\bzeta({\bf u}, q_\bphi|\z) = \bzeta(\tilde \brho, q_\bphi)$, where $\tilde \brho = 1 -q_\bphi  + {\bf u} \left(\z q_\bphi - (1-\z) (1-q_\bphi)\right)$. In order to minimize the variance, it is preferable to tie the parameters $\brho$ and $(\z, {\bf u})$. This can be achieved by setting $\z = \Theta(\brho-1+q_\bphi)$, and choosing 
\be
u = 
\begin{cases}
\frac{\brho - 1 + q_\bphi}{q_\bphi} & {\rm if} \: z=1, \\
-\frac{\brho - 1 + q_\bphi}{1-q_\bphi} & {\rm if} \: z=0.
\end{cases}
\ee
This is equivalent to sampling $(\z, \bzeta) \sim q_{\bphi}(\z) q_\bphi(\bzeta | \z)$ as $\z = \Theta(\brho-1+q_\bphi),  \bzeta = \bzeta(\tilde \brho, q_\bphi)$ where
\ba
&& \tilde \brho = 1 -q_\bphi  + {\bf sg}(\brho - 1 + q_\bphi) \left(\z \frac{q_\bphi}{{\bf sg}(q_\bphi)} + (1-\z) \frac{1-q_\bphi}{1-{\bf sg}(q_\bphi)} \right). \nonumber
\ea
This way $\tilde \brho = \brho$ by value but $\tilde \brho$ does depend on $\bphi$  such that  $\bzeta(\tilde \brho,q_\bphi)$ has the correct gradient w.r.t. $\bphi$ that follows from the conditional distribution $q_\bphi(\bzeta | \z)$. We can rewrite (\ref{eq:grad_rebar1}) as
\be\label{eq:grad_rebar2}
\partial_\phi {\cal L} = \E_{q_\bphi(\z) q_\bphi(\bzeta | \z)} \left[\partial_\phi \log q_\bphi(\z)  \left( f(\z) - f(\bzeta) \right) \right] + \E_{\brho} \left[ \partial_\bzeta f(\bzeta) \left( \partial_\bphi \bzeta(\brho, q_\bphi)  - \partial_\bphi \bzeta(\tilde \brho, q_\bphi)\right) \right].
\ee
Since the explicit dependence of $\bzeta(\brho, q_\bphi)$ and $\bzeta(\tilde \brho, q_\bphi)$ on $q_\bphi$ is the same, we have $\partial_\bphi \bzeta(\brho, q_\bphi) - \partial_\bphi  \bzeta(\tilde \brho, q_\bphi) = -  \partial_\bphi  \bzeta(\tilde \brho, {\bf sg}(q_\bphi))$
and we arrive at our final expression for the REBAR gradient:
\be\label{eq:rebar}
\partial_\phi {\cal L} = \E_{\brho,\z = \Theta(\brho-1+q_\bphi) } \left[\partial_\phi \log q_\bphi(\z)  \left( f(\z) - f(\bzeta(\brho, q_\bphi)) \right) - \partial_\bphi f(\bzeta(\tilde \brho, {\bf sg}(q_\bphi))) \right].
\ee
The advantage of Eq.~(\ref{eq:rebar}) is two-fold: first, it allows for a simple implementation valid for any continuous relaxation and uses only two function evaluations. Second, it gives a suggestive relation to the ICR estimator: indeed using $\partial_\phi \tilde \brho = \partial_\bphi q_\bphi \left( -1  + \z \frac{\brho - 1 + q_\bphi}{q_\bphi} - (1-\z) \frac{\brho - 1 + q_\bphi}{1-q_\bphi} \right)$ we can write REBAR gradient as
\ba\label{eq:rebar_icr}
&& \partial_\phi {\cal L}_{\rm REBAR} = \partial_\phi {\cal L}_{\rm ICR} + R_1 +R_2 \nn
&& \partial_\phi {\cal L}_{\rm ICR}  = \E_{\brho} \left[\partial_\bphi q_\bphi  \partial_\brho \bzeta \partial_\bzeta f(\bzeta) \right] \nn
&& R_1 = \E_{\brho,\z = \Theta(\brho-1+q_\bphi) } \left[  - \partial_\bphi q_\bphi  \partial_\brho \bzeta \partial_\bzeta f(\bzeta) \left( \z \frac{\brho - 1 + q_\bphi}{q_\bphi} - (1-\z) \frac{\brho - 1 + q_\bphi}{1-q_\bphi} \right)
\right]\nn
&& R_2 = \E_{\brho,\z = \Theta(\brho-1+q_\bphi) } \left[ \partial_\phi \log q_\bphi(\z)  \left( f(\z) - f(\bzeta(\brho, q_\bphi)) \right)
\right].
\ea
Here $R_{1,2}$ are correction terms. The term $R_2$ can have high variance due to the fact that ${\rm var} (\partial_{q_\phi} \log q_\bphi(\z)) \sim \frac{1}{q_\phi (1-q_\phi)}$. On the other hand $R_1$ has mean and variance comparable to $\partial_\phi {\cal L}_{\rm ICR}$ because ${\rm mean} (z \frac{\rho - 1 + q_\bphi}{q_\bphi} - (1-z) \frac{\rho - 1 + q_\bphi}{1-q_\bphi} )= \frac{1}{2}$ and ${\rm var} (z \frac{\rho - 1 + q_\bphi}{q_\bphi} - (1-z) \frac{\rho - 1 + q_\bphi}{1-q_\bphi} )= \frac{1}{3}$. So inclusion of $R_1$ roughly lowers the magntidue of the gradient ICR gradient. In summary REBAR estimator can be viewed as ICR plus corrections that remove the bias of ICR estimator. In practice the correction term $R_2$ can increase the variance of REBAR estimator compared to ICR and slow down the training. \cite{tucker2017rebar} learn $\beta$ that minimizes the variance of the gradient (\ref{eq:rebar}) by trading off the variance of the two terms $\partial_\phi \log q_\bphi(\z)  \left( f(\z) - f(\bzeta(\brho, q_\bphi)) \right)$ and $\partial_\bphi f(\bzeta(\tilde \brho, {\bf sg}(q_\bphi)))$.

\subsection{RELAX}\label{subsec:relax}

In this section we discuss the RELAX estimator \cite{grathwohl2017backpropagation} and propose an alternative version of it. The idea behind RELAX estimator is to enhance the control variate $f(\bzeta)$ used in (\ref{eq:grad_rebar}) by adding to it an additional learnable function $r_\psi(\bzeta)$ parameterized by neural networks:
\be
f(\bzeta) \to c_\psi(\bzeta) \equiv  f(\bzeta) + r_\psi(\bzeta).
\ee
Parameters $\psi$ and $\beta$ are learned in order to minimizing the variance of the gradient  
\be\label{eq:grad_rho}
{\rm grad}_{\bphi}(\brho) = \partial_\phi \log q_\bphi(\z)  \left( f(\z) - c_\psi(\bzeta(\brho, q_\bphi)) \right) - \partial_\bphi c_\psi(\bzeta(\tilde \brho, {\bf sg}(q_\bphi))).
\ee
Since $\partial_{\psi, \beta} \E_\brho \left[ {\rm grad}_{\bphi}(\brho)\right]=0$ this corresponds to a penalty term \be\label{eq:penalty_relax}
O_{\psi, \beta}^{\rm RELAX} = \E_\brho \left[ \left({\rm grad}_{\bphi}(\brho) \right)^2\right],
\ee
added to the objective. As noted in \cite{grathwohl2017backpropagation} the new control variate $c_\psi(\bzeta)$ can be a better approximation to $f(\z)$, as a function of $\brho$, thereby reducing the first term in (\ref{eq:grad_rho}) and reducing the variance of the full gradient ${\rm grad}_{\bphi}(\brho)$. 

In general the gradient (\ref{eq:grad_rho}) has two major sources of variance: $ \partial_\phi \log q_\bphi(\z)$ in the first term and $\p_{\tilde \brho} \bzeta(\tilde \brho,q_\bphi)$ in the second term. As discussed in Section~\ref{sec:CREstimators} the variance of $\p_{\tilde \brho} \bzeta(\tilde \brho,q_\bphi)$ is controlled by relaxation parameter $\beta$, while the variance of the second term is controlled by the magnitude of the coefficient in front of $ \partial_\phi \log q_\bphi(\z)$. \cite{grathwohl2017backpropagation} mimize the penalty (\ref{eq:penalty_relax}) by adjusting both $\psi$ and $\beta$, thus finding a trade-off between the contribution of the terms to the variance. We pursue a slightly different approach: we fix $\beta$ and try to minimize the variance of the first term only in (\ref{eq:grad_rho}), which can be done by minimizing 
\be\label{eq:penalty_ours}
O_{\psi}^{\rm RELAX+} = \E_\brho \left[ \left| f(\z) - c_\psi(\bzeta(\brho, q_\bphi) \right|\right] = \E_\brho \left[ \left| f(\z) - f(\bzeta) - r_\psi(\bzeta) \right|_{\bzeta=\bzeta(\brho, q_\bphi)}\right].
\ee
Furthermore, since even with the penalty (\ref{eq:penalty_ours}) the variance of the gradient (\ref{eq:grad_rho}) can be dominated by the first term, we can introduce a biased gradient estimate by adding a hyperparameter $\gamma \in [0, 1]$ in front of the first term. By adjusting $\gamma$ it is possible to trade variance for bias in the gradient ${\rm grad}_{\bphi}(\brho)$. In summary our proposed method, that we will refer to as RELAX+, is given by
\ba
&& {\rm grad}_{\bphi}(\brho) = \gamma \partial_\phi \log q_\bphi(\z)  \left( f(\z) - c_\psi(\bzeta(\brho, q_\bphi)) \right) - \partial_\bphi c_\psi(\bzeta(\tilde \brho, {\bf sg}(q_\bphi))) \nn
&& {\rm grad}_{\psi}(\brho) = \p_\psi  \left| f(\z) - f(\bzeta) - r_\psi(\bzeta) \right|_{\bzeta=\bzeta(\brho, q_\bphi)}.
\ea

This formulation makes it clear that the optimal function $r_\psi(\bzeta)$ would satisfy $r_\psi(\z)=0$. We can use this information when constructing parametric form of $r_\psi(\bzeta)$. In our experiments we found that parameterization  $r_\psi(\bzeta) = \sum_{i=1}^M \bzeta_i (1 - \bzeta_i) g^i_\psi(\bzeta)$, with a neural network $g^i_\psi(\bzeta)$, worked noticeably better then directly parameterizing $r_\psi(\bzeta)$. It is interesting to note that RELAX+ at $\gamma=0$ is very similar to the continuous relaxation  with the objective $f(\bzeta)$ replaced by $f(\bzeta) + r_\psi(\bzeta)$. In terms of computational complexity we find RELAX+ to be about $40\%$ cheaper then RELAX.

\section{Experiments}
\label{sec:exp}

In this section we compare the FD, SF and CR estimators and their improved variants on a number of examples. We start with one-variable toy examples that illustrate the bias of GSM estimator and then move to the training of variational auto-encoders. Finally, we apply the estimators to the discrete maximum clique optimization problem analyzed in \cite{patish2018cakewalk}.

\subsection{Toy example}\label{subsec:toy}

\begin{figure}[t]
    \centering
    \vspace{-1.5cm}
 \subfloat[Relaxed function]{ \includegraphics[scale=0.29]{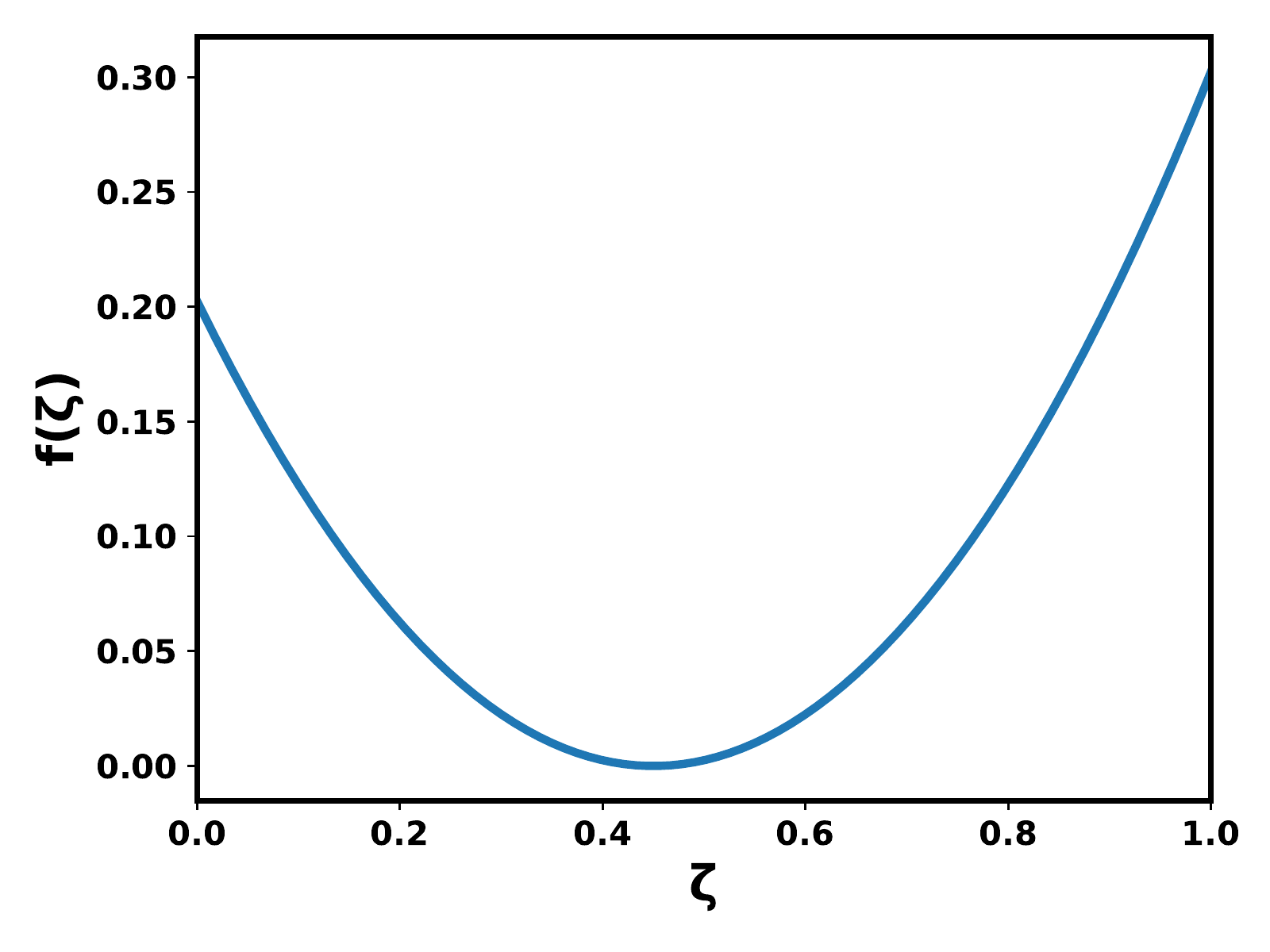}} 
  \subfloat[Probability $q(z=1)$]{ \includegraphics[scale=0.29]{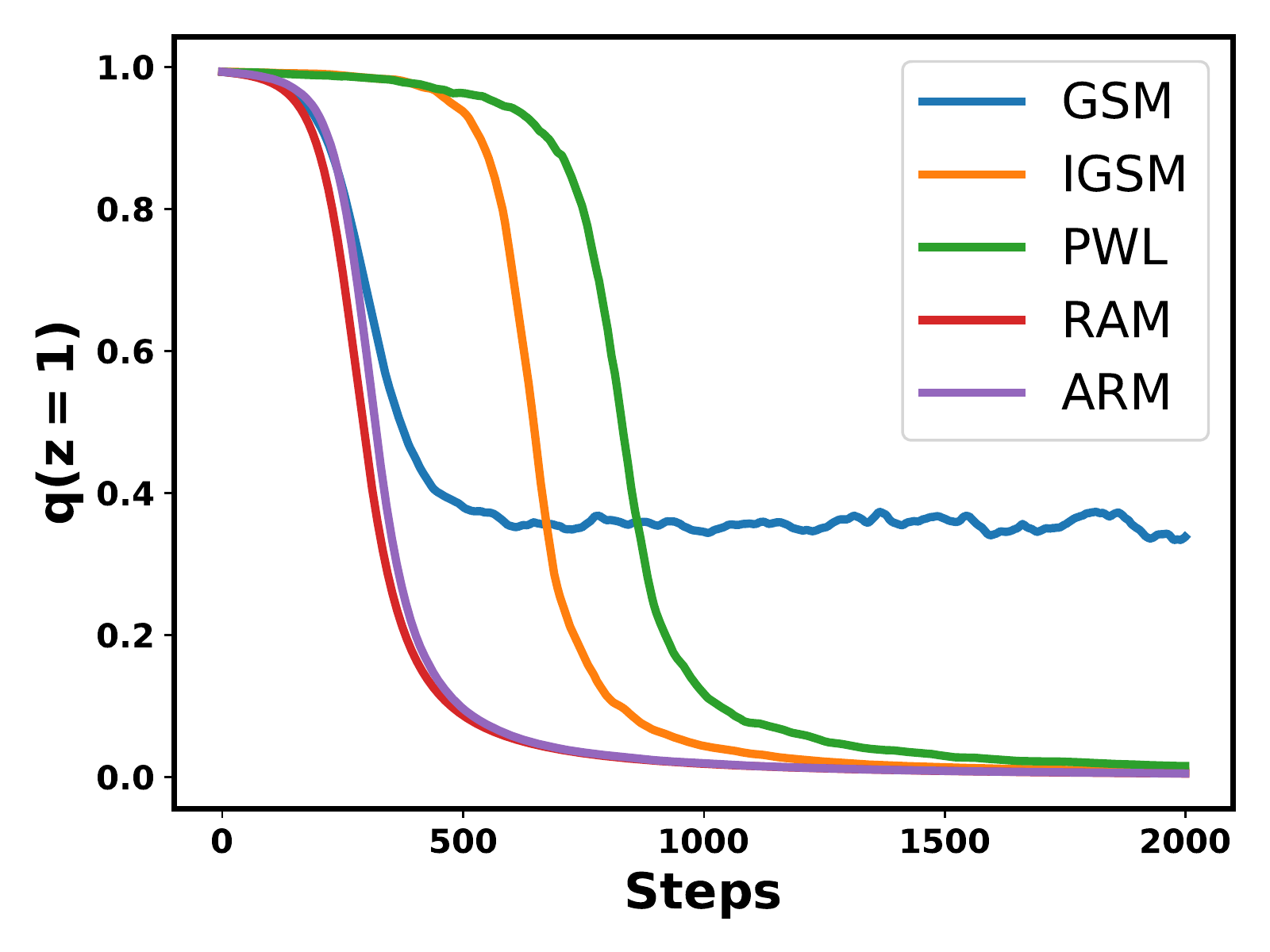}}
    \subfloat[Gradient $\p_l \mathcal{L}$]{ \includegraphics[scale=0.29]{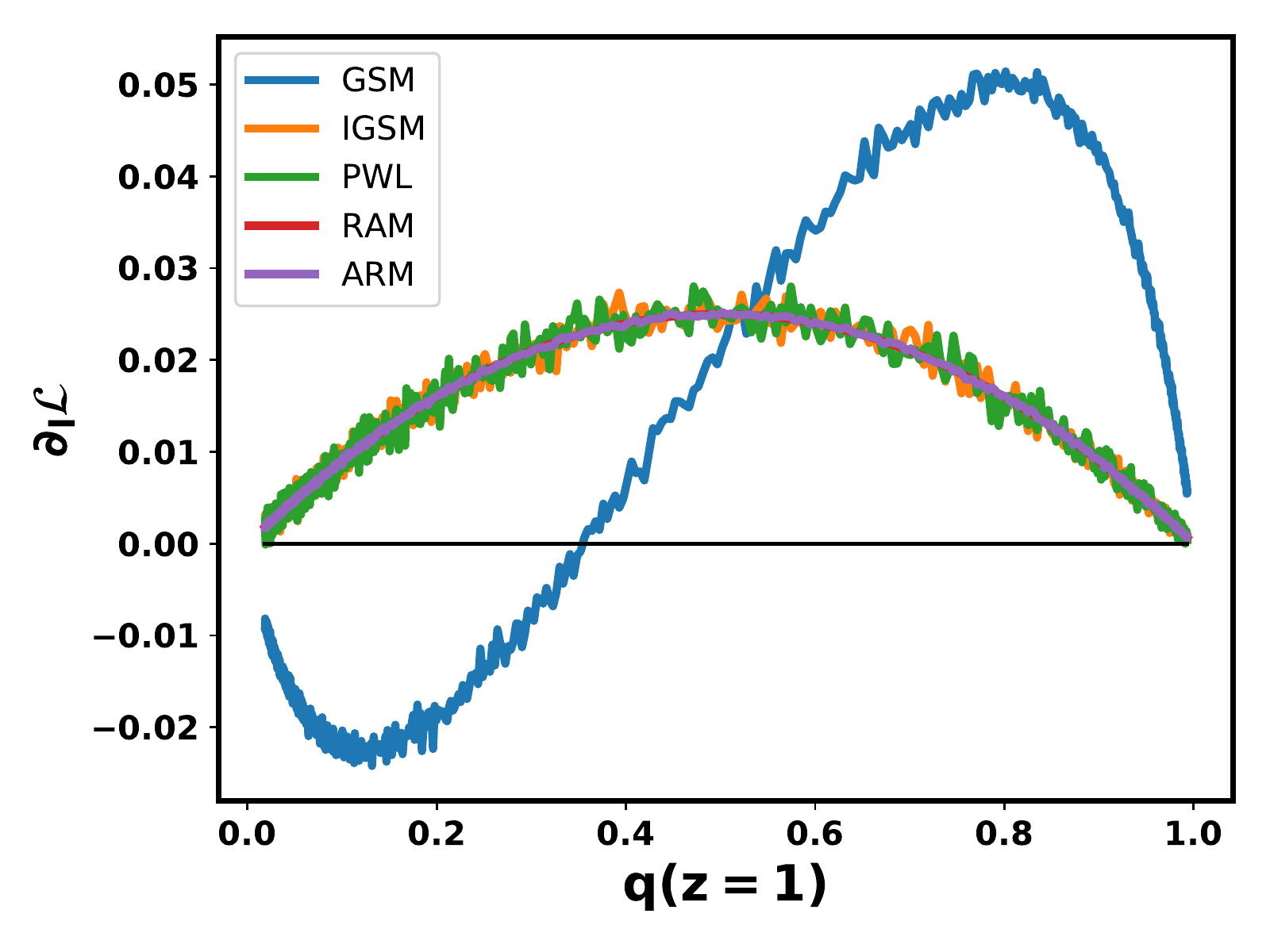}}
\caption{Convex, single  binary-variable toy example: (a) The relaxed objective$f(\zeta) = (\zeta - 0.45)^2$. (b) The probability $q_\phi(z=1)$ during optimization. (c) The gradient $\p_l \mathcal{L}$ of all estimators; the bias of GSM prevents proper minimization.} \label{fig:binary_toy_convex}
\end{figure}

We begin with an illustrative single-variable example (\cite{tucker2017rebar}) with objective ${\cal L} = \sum_z q_\phi(z) f(z) $ where $f(z) = (z - 0.45)^2$. The relaxed convex function $f(\zeta)$ depicted in Fig.~\ref{fig:binary_toy_convex}(a) has two local maxima, one of which is the minimum over the discrete domain. We compare five gradient methods: RAM, Eq.~(\ref{eq:grad_1}); ARM, Eq.~(\ref{eq:grad_arm_bin}); PWL, Eq.~(\ref{eq:pwl_relax}); GSM, Eq.~(\ref{eq:gsm_relax}); and improved Gumbel-Softmax (IGSM), Eq.~(\ref{eq:qrho_trick}) (see Appendix \ref{app:exp_det} for experimental details).  Fig.~\ref{fig:binary_toy_convex}(b) shows the evolution of $q_\phi(z=1)$ during training and it confirms the bias associated with GSM. To quantify this bias, we plot the value of the gradient of all estimators for different values of $q_\phi(z=1)$ in Fig.~\ref{fig:binary_toy_convex}(c). We observe that the GSM gradient has the wrong sign for a large interval in $q_\phi(z=1)$ which prevents GSM from converging to the true minimum.

To understand the nature of the bias that GSM introduces we plot the derivatives $\p_q \zeta$ and $\p_\rho \zeta$ corresponding to GSM and IGSM respectively in Fig.~\ref{fig:cr}(c). We see that $\p_q \zeta$ is biased towards the value of $z = {\rm round}(\zeta)$ that has the highest probability. This means that
GSM in Eq.~(\ref{eq:grad_relaxed_fact}) will oversample the derivative $\p_{\zeta} f(\zeta)$ from the most probable mode. In example Fig.~\ref{fig:binary_toy_convex}(a) this results in oversampling the derivative from $z=0$ mode which creates a gradient that pushes optimization away from the true minimum $z=0$. The bias is reduced as $\beta$ is increased.

In Appendix~\ref{app:binary_toy_concave} we consider an example for a concave function over a binary variable and observe similar effects. We also consider both concave and convex functions over a categorical variable in Appendix~\ref{app:categorical_toy_concave}. In all scenarios, the bias of GSM prevents its convergence to the correct minimum.

\subsection{Discrete Variational Autoencoders}\label{subsec:DVAE}

Next, we test the estimators by training variational autoencoders \cite{kingmaICLR15} with discrete priors. The objective is the negative expectation lower bound on the log-likelihood (NELBO):
\be\label{eq:elbo}
 {\cal L}[\bphi, \btheta] = \sum_{\x \sim {\rm data}} \E_{\z \sim q_\bphi(\z|\x)} \left[ f_{\btheta, \bphi}(\z, \x) \right], \: \text{where} \quad  f_{\btheta, \bphi}(\z, \x) = -\log \frac{p_\btheta(\z) p_\btheta(\x|\z)}{q_\bphi(\z|\x)}.
\ee
where $p_\btheta(\z)$ is the prior, $p_\btheta(\x|\z)$ is the decoder, and $q_\bphi(\z|\x)$ is the approximating posterior. ${\cal L}$ is minimized with respect to the $\btheta$ parameters of the generative model and the $\bphi$ parameters of the approximate posterior. The latter minimization corresponds to Eq.~(\ref{eq:object_gen}) and we can apply the various estimators to propagate $\bphi$-derivatives through discrete samples $\z$. For CR estimators we replace $\z \to \bzeta$ to compute
\be\label{eq:phi_deriv}
 \p_\bphi  {\cal L}[\bphi, \btheta] \approx \sum_{\x \sim {\rm data}} \E_{\rho} \left[\p_\bphi f_{\btheta, \bphi}(\bzeta, \x) \right],
\ee
where the expectation is evaluated with a single relaxed sample per data point $\x$. The $\btheta$-derivative can be calculated
directly from Eq.~(\ref{eq:elbo}) using the discrete $\z$ (not the relaxed $\bzeta$) variables.
\iffalse
On the other hand $\btheta$ derivative is given by 
\be\label{eq:theta_deriv_discr}
\p_\theta L_{\bphi, \btheta} = -\!\sum_{x \sim {\rm data}} \E_{z \sim q_\bphi}[ \p_\btheta \log \frac{p_\btheta(z) p_\btheta(x|z)}{q_\bphi(z|x)}]
\ee
%
where expectation is evaluated with a single discrete sample per data point $x$. 
\fi
Thus, the $\bphi$ and $\btheta$ derivatives are evaluated separately requiring two passes through the computation graph with either relaxed or discrete samples. \cite{jang2016categorical} evaluate both derivatives in one pass (using $\bzeta$) thereby introducing bias in the $\btheta$ derivatives. We refer to these two possibilities as one-pass and two-pass training and compare their performance.
\begin{figure}[t]
\vspace{-1cm}
    \centering
  \subfloat[NELBO]{ \includegraphics[scale=0.25]{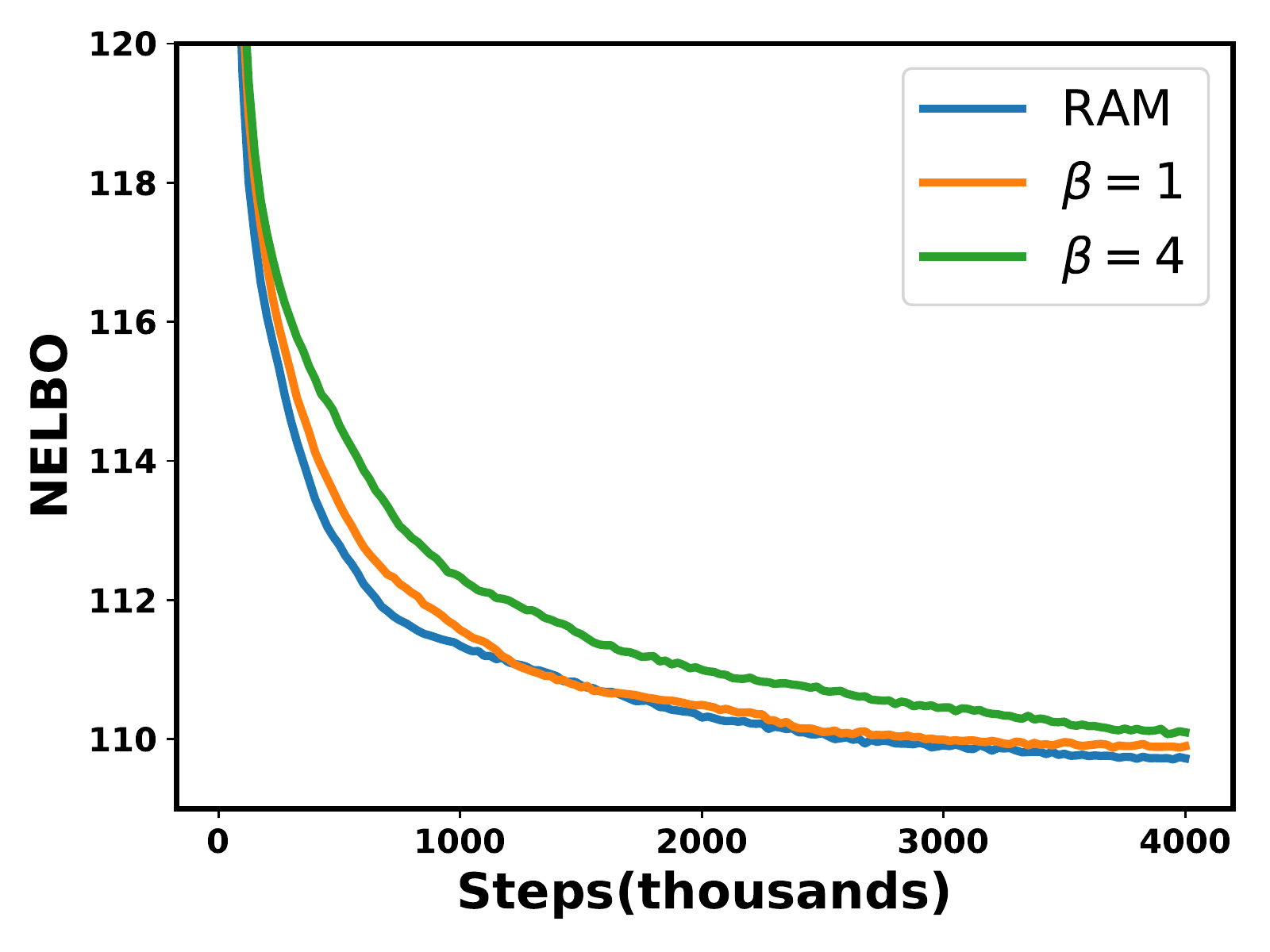}}
    \subfloat[Evaluations per sample]{ \includegraphics[scale=0.25]{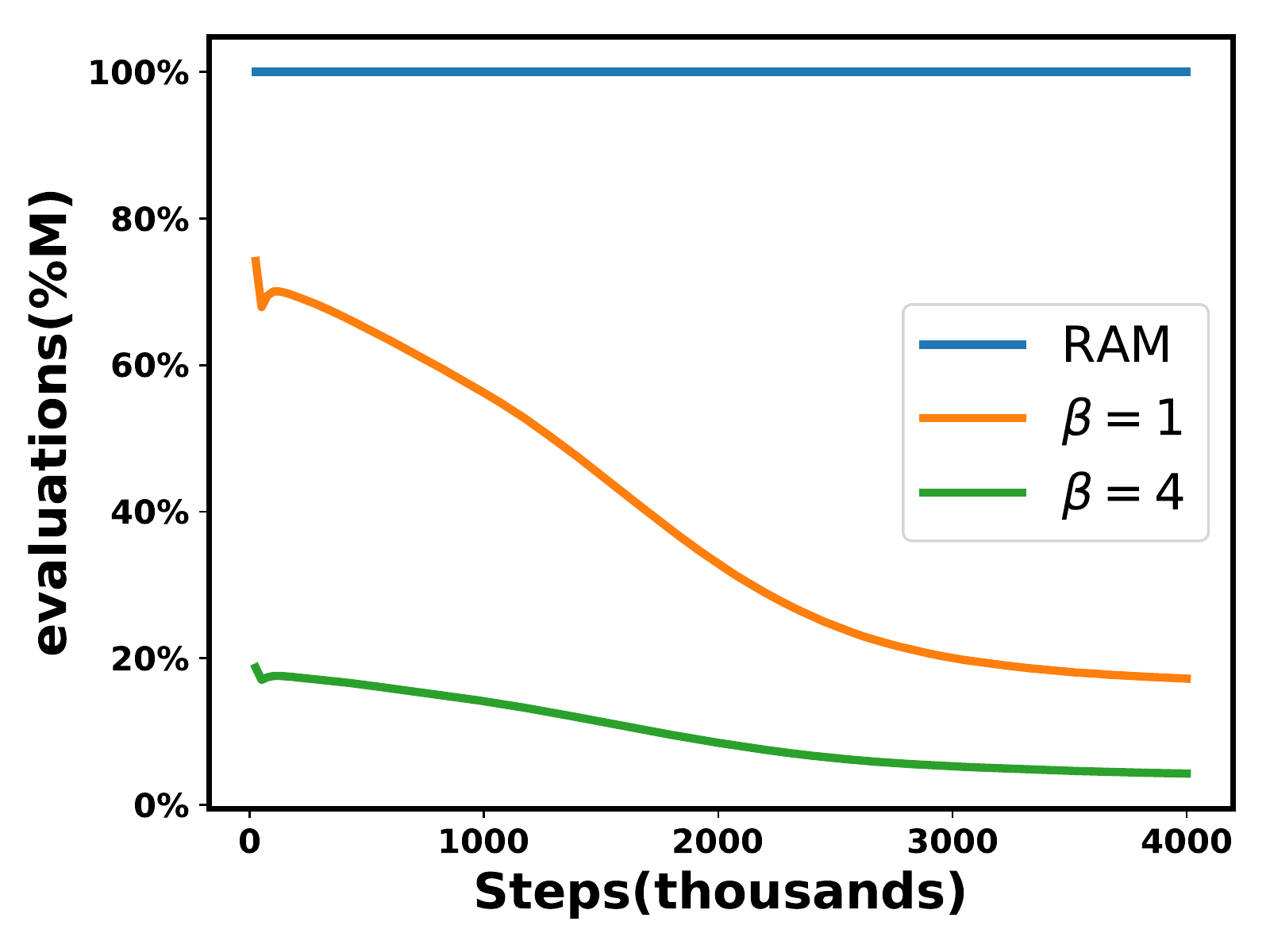}}
\caption{NELBO of the RAM and sampled RAM estimators on MNIST trained using the $200{\rm H}-784{\rm V}$ architecture having a linear decoder: (a) plots the decrease in training NELBO for RAM and two variants of sampled RAM. (b) shows the computational savings of sampled RAM.}
    \label{fig:bin_vae_fd_beta2}
\end{figure}

Following \cite{maddison2016concrete, jang2016categorical, tucker2017rebar}, we consider four architectures with binary variables denoted by  $200{\rm H}-784{\rm V}$, $200{\rm H}-200{\rm H}-784{\rm V}$, $200{\rm H}\sim784{\rm V}$, and $200{\rm H}\sim200{\rm H}\sim784{\rm V}$ (see Appendix \ref{app:exp_det} for details). First, we compare the sampled RAM estimator of Eq.~(\ref{eq:grad_fact_sampled}) on $200{\rm H}-784{\rm V}$ model for different values of $\beta$, where the probability of updating variable $z_i$ is $p_i = 4 q_i (1-q_i)/\beta$. Fig.~\ref{fig:bin_vae_fd_beta2}(a) shows NELBO on the training set. As expected, increased $\beta$ leads to increased gradient variance which slows training. Fig.~\ref{fig:bin_vae_fd_beta2}(b) shows the average number of function evaluations performed in Eq.~(\ref{eq:grad_fact_sampled}). Many units become deterministic early in training leading to significant computational savings with the sampled RAM method.

\begin{figure}[t]	
%\vspace{-1cm}
    \centering
  \subfloat[train NELBO]{ \includegraphics[scale=0.25]{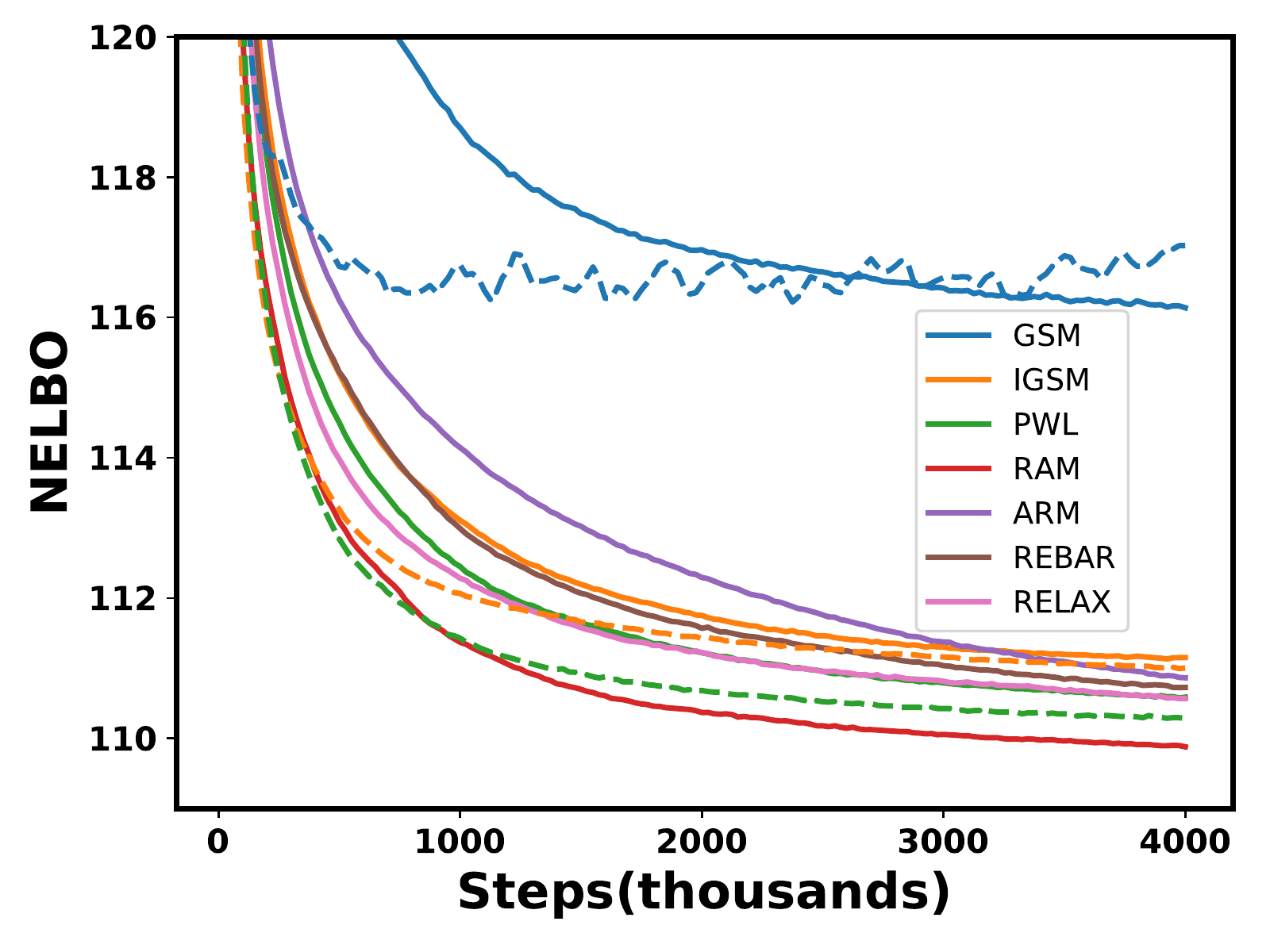}}
    \subfloat[Entropy of $q_\phi(z)$]{ \includegraphics[scale=0.25]{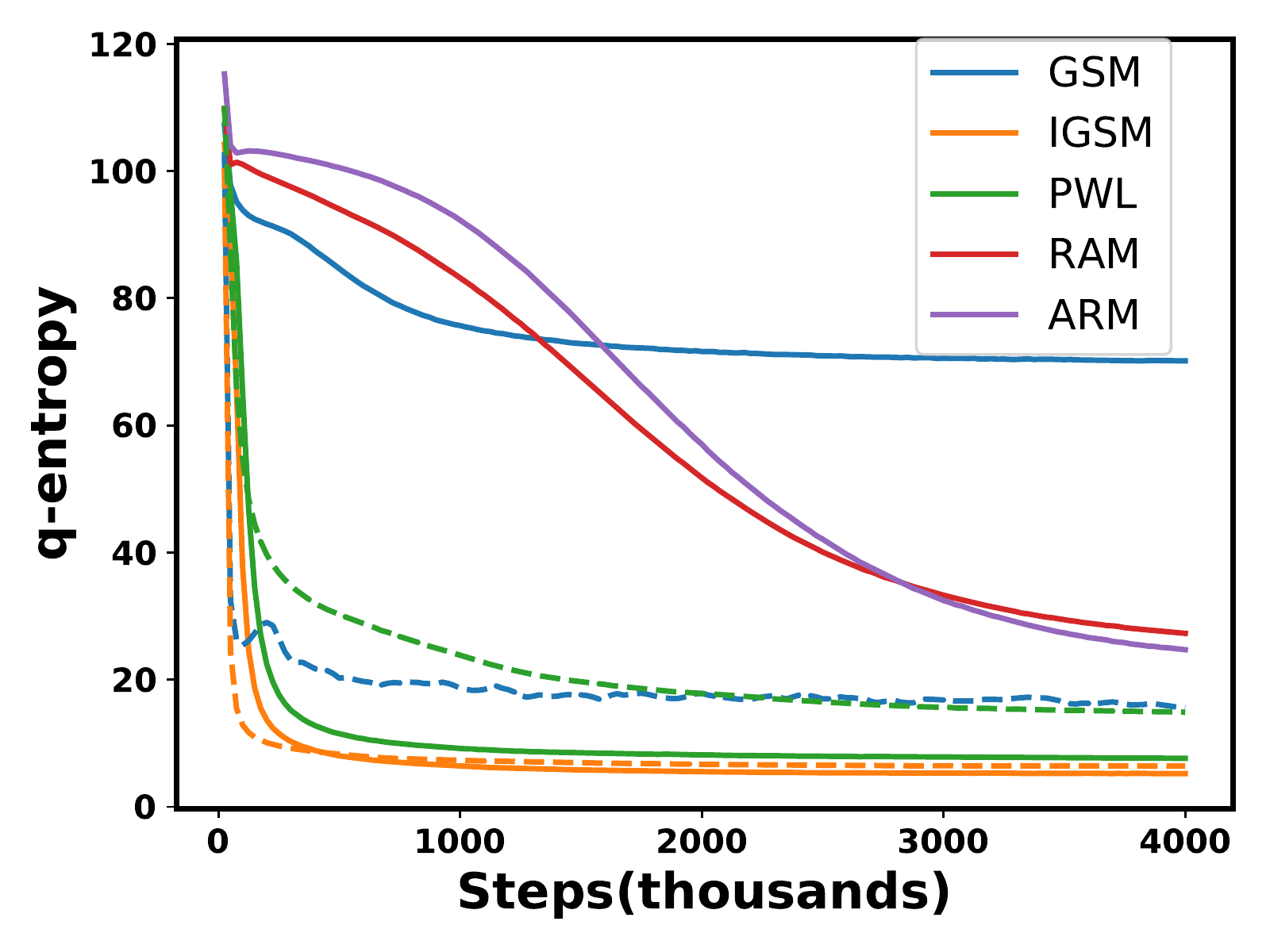}}
 \subfloat[Final train NELBO]{ \includegraphics[scale=0.25]{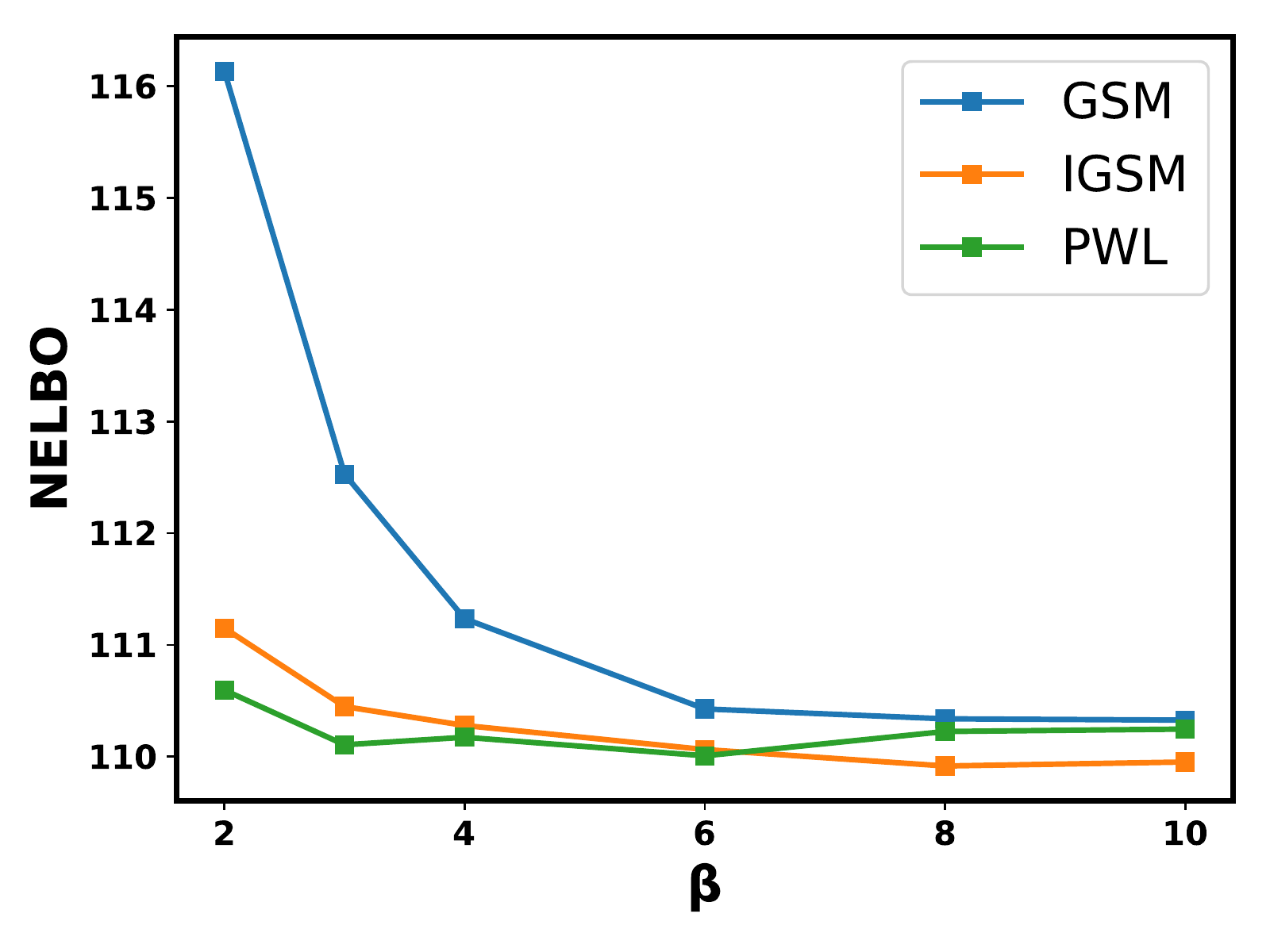}}
\caption{MNIST training on the linear architecture $200{\rm H}-784{\rm V}$: (a) compares training NELBO of CR estimators (all estimators use $\beta=2$). Solid/dashed lines correspond to one/two-pass training. (b) shows the entropy of the learned posterior approximations; the GSM bias induces much higher entropy. (c) dependence of the final trained NELBO on $\beta$; higher $\beta$ corresponds to lower bias.}
    \label{fig:bin_vae_cr_beta2}
\end{figure}

In Fig.~\ref{fig:bin_vae_cr_beta2}(a), we include several CR estimators on the same architecture and observe that the GSM estimator performs significantly worse than other estimators. With linear decoders, the objective function Eq.~(\ref{eq:elbo}) is convex in $z_i$. Thus, similar to the example in Fig.~\ref{fig:binary_toy_convex}, GSM learns a distribution with higher entropy leading to poorer performance. The entropy of all estimators during training is shown in Fig.~\ref{fig:bin_vae_cr_beta2}(b). We note that two-pass training performs better then one-pass training for all estimators. Although two-pass training requires twice the computation in the worst case, this overhead is negligible for these models due to GPU parallelization. We observe that ARM performs poorly confirming that its high variance impedes training. The PWL estimator with two-pass training performs on par with RAM, while being more computationally efficient. The REBAR estimator (\ref{eq:rebar}), using the settings of \cite{tucker2017rebar} and training $\beta$ to minimize the variance of the gradients, performs worse then PWL. The RELAX estimator of \cite{grathwohl2017backpropagation} performs on par with PWL by using a better control variate for the gradient estimation then REBAR. In the above experiments we used $\beta=2$ for CR estimators, similar to \cite{jang2016categorical, maddison2016concrete}. To  understand the dependence on $\beta$, we plot final train NELBO in Fig.~\ref{fig:bin_vae_cr_beta2}(c). We find that improved CR estimators are less sensitive to the choice of $\beta$.

\begin{figure}[t]	
%\vspace{-1cm}
    \centering
  \subfloat[train NELBO]{ \includegraphics[scale=0.25]{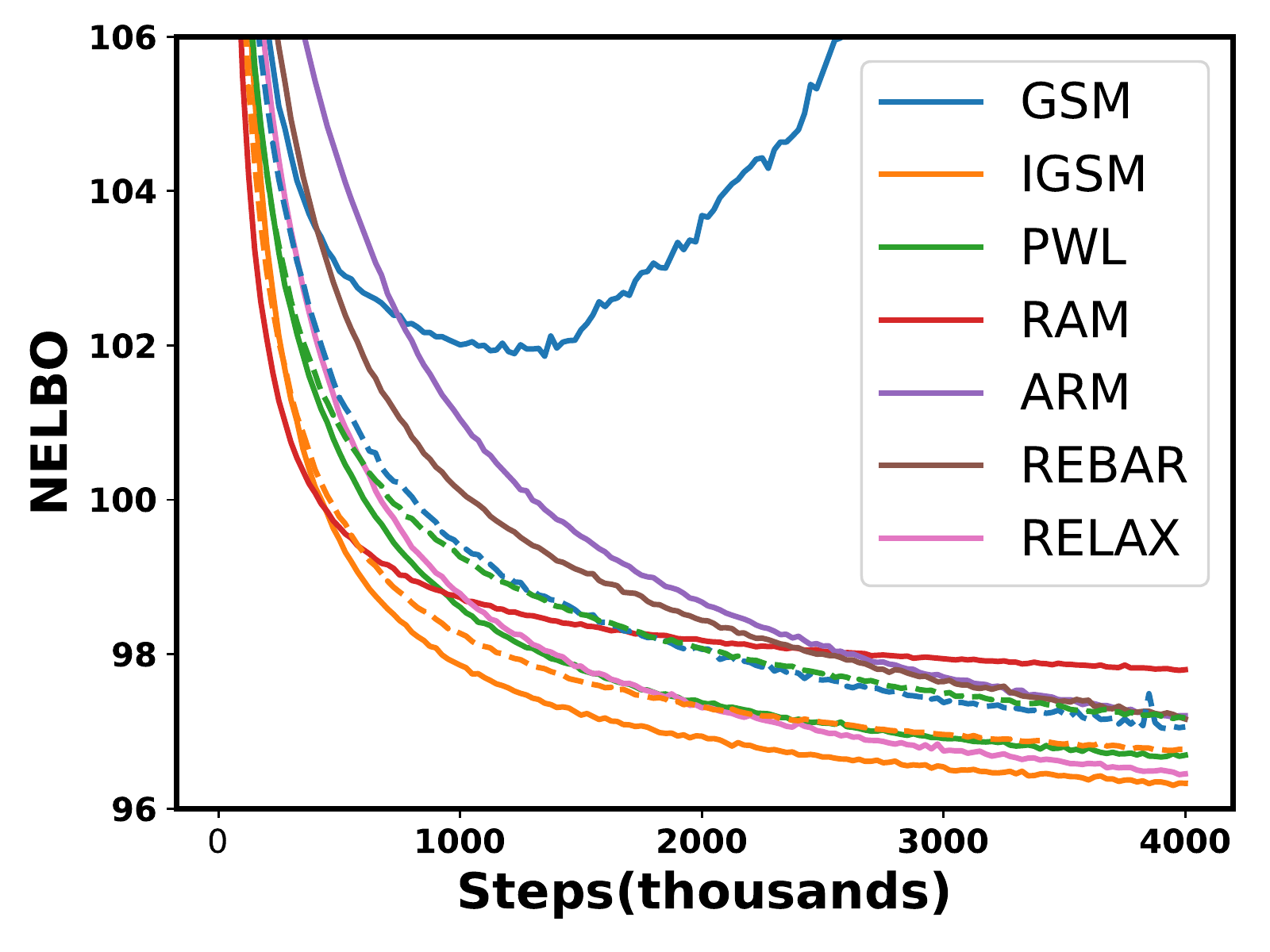}}
 \subfloat[Final train NELBO]{ \includegraphics[scale=0.25]{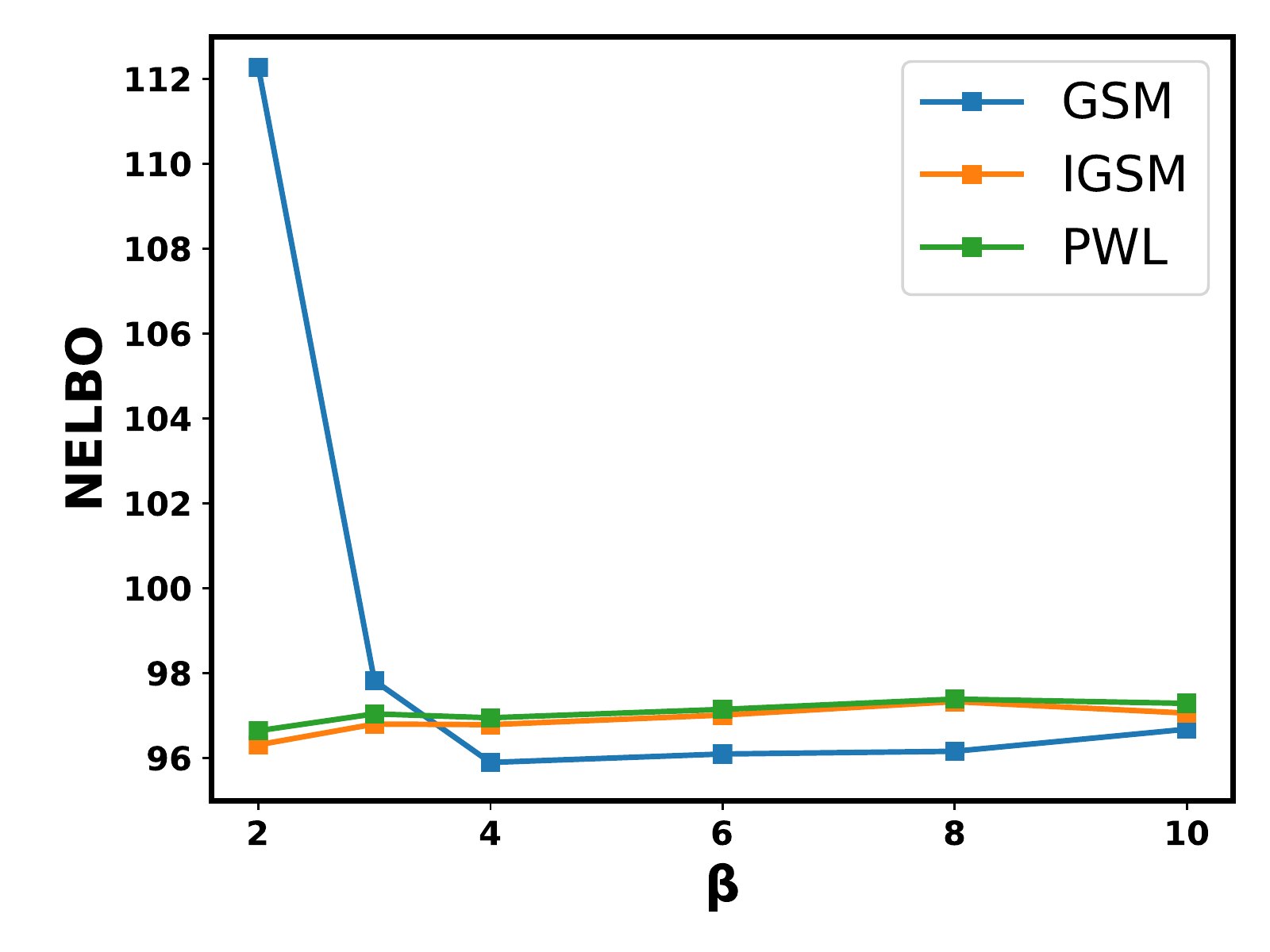}}
 \subfloat[Final train NELBO, KL anneal 10\% of training]{ \includegraphics[scale=0.25]{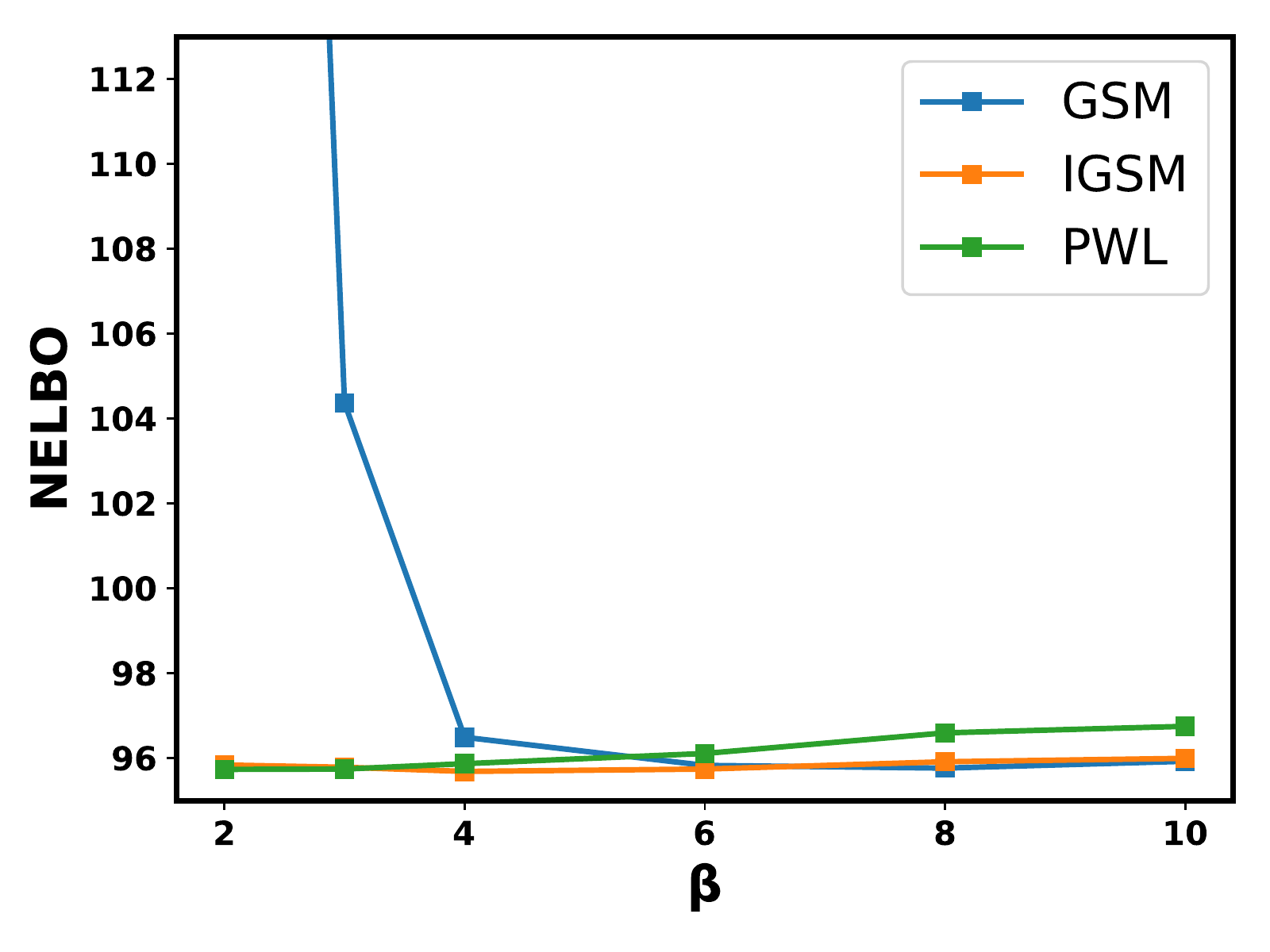}}
\caption{MNIST training on the non-linear architecture $200{\rm H}\sim784{\rm V}$: (a) compares training NELBO of CR estimators (all estimators use $\beta=2$). Solid/dashed lines correspond to one/two-pass training. (b) Final training NELBO for different $\beta$, (c) Final training NELBO using KL annealing; explicit annealing erases the gains of GSM.}
    \label{fig:bin_vae_cr_nonlin_beta2}
\end{figure}

In Fig.~\ref{fig:bin_vae_cr_nonlin_beta2}, we repeat the experiments for the non-linear architecture $200{\rm H}\sim784{\rm V}$. The one-pass GSM estimator exhibits instability which is remedied by two-pass training. However, two-pass GSM still performs worse than IGSM and PWL due to its bias. Interestingly, one-pass training works better for IGSM/PWL. We observe this repeatedly in the nonlinear models. Unlike the linear case, the RAM estimator converges faster initially but later in training is outperformed by the higher-variance IGSM, PWL and ARM. It is likely that additional noise prevents the latent units from being turned off early in training which is known to cause poor performance in VAEs. As in the linear case we plot the dependence of the final train NELBO on $\beta$ in Fig.~\ref{fig:bin_vae_cr_nonlin_beta2}(b). The GSM estimator outperforms IGSM and PWL for $\beta \ge 4$ because its bias favors higher entropy approximating posteriors which inhibit latent units from turning off early in training. A well known resolution for inactive latent units is KL-annealing \cite{bowman2016generating}. Fig.~\ref{fig:bin_vae_cr_nonlin_beta2}(c) shows that KL annealing indeed improves IGSM and PWL by inhibiting over-pruning of latent units\footnote{In experiments not reported here we observed that the entropy regularizing benefits of the GSM bias can be achieved with explicit entropy regularization of the objective.}. As with the linear case, the IGSM and PWL estimators are more stable under variations in $\beta$.   Here, REBAR underperforms the ICR estimators likely due to its higher variance. At the same time RELAX estimator is able to successfully reduce the variance and performs on par with ICR.

\begin{figure}
\vspace{-0.5cm}
\centering
  \subfloat[$200{\rm H}-784{\rm V}$]{\includegraphics[scale=0.25]{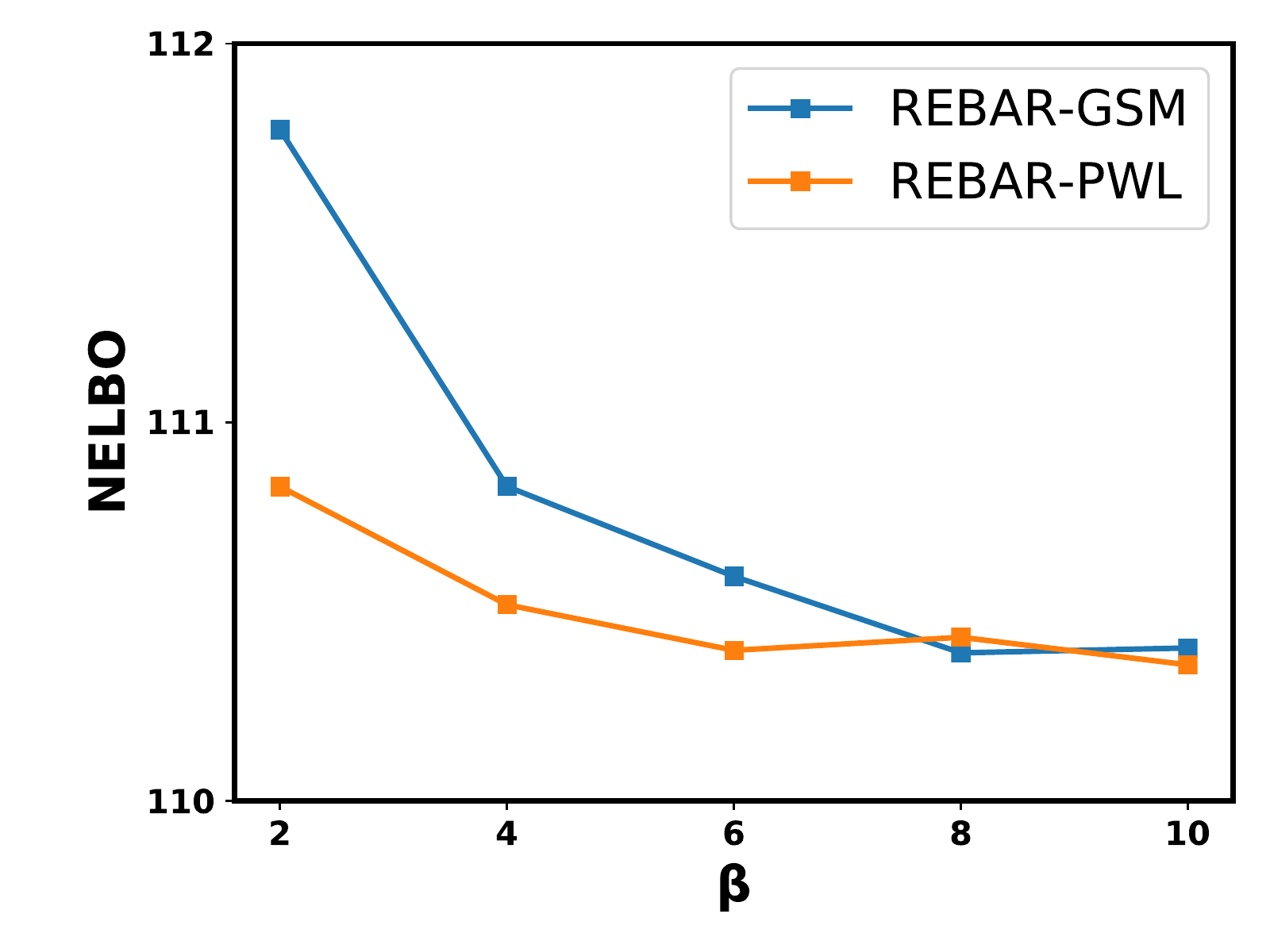}}
  \subfloat[$200{\rm H}\sim784{\rm V}$]{\includegraphics[scale=0.25]{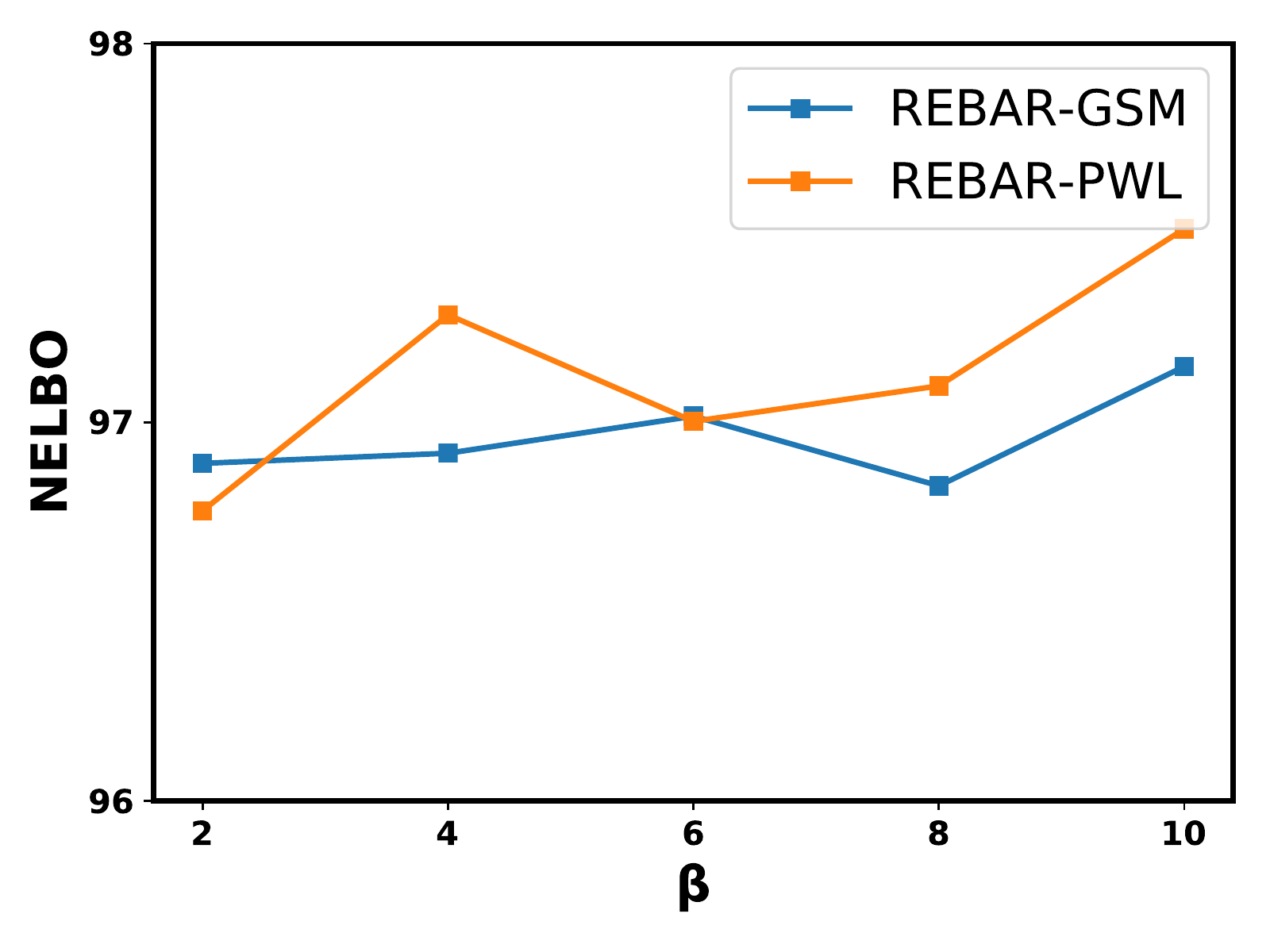}}
  \caption{MNIST training with REBAR using GSM and PWL relaxations, dependence of the final trained NELBO on $\beta$: (a) the linear architecture. (b) the non-linear architecture.}
 \label{fig:bin_vae_rebar_betas}
\end{figure}

We perform the comparison of GSM and PWL continuous relaxations used by REBAR estimator. Fig.~\ref{fig:bin_vae_rebar_betas} shows the dependence of the final NELBO on $\beta$. In the case of the linear model $200{\rm H}-784{\rm V}$ we find that PWL relaxation is slightly more stable while for the non-linear model the two relaxations perform similarly.

\begin{figure}
\vspace{-0.5cm}
\centering
  \subfloat[$200{\rm H}-784{\rm V}$]{\includegraphics[scale=0.25]{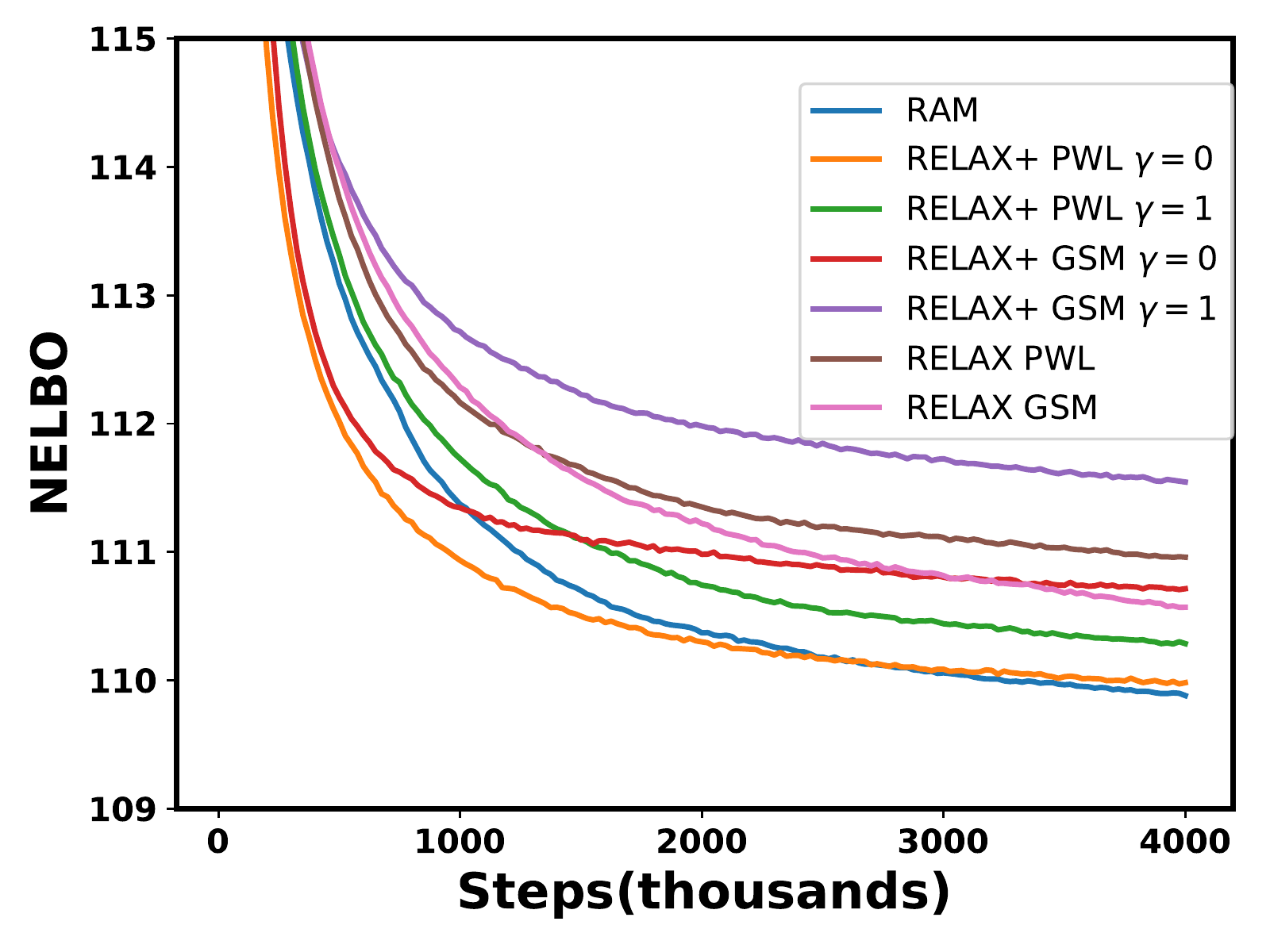}}
  \subfloat[$200{\rm H}\sim784{\rm V}$]{\includegraphics[scale=0.25]{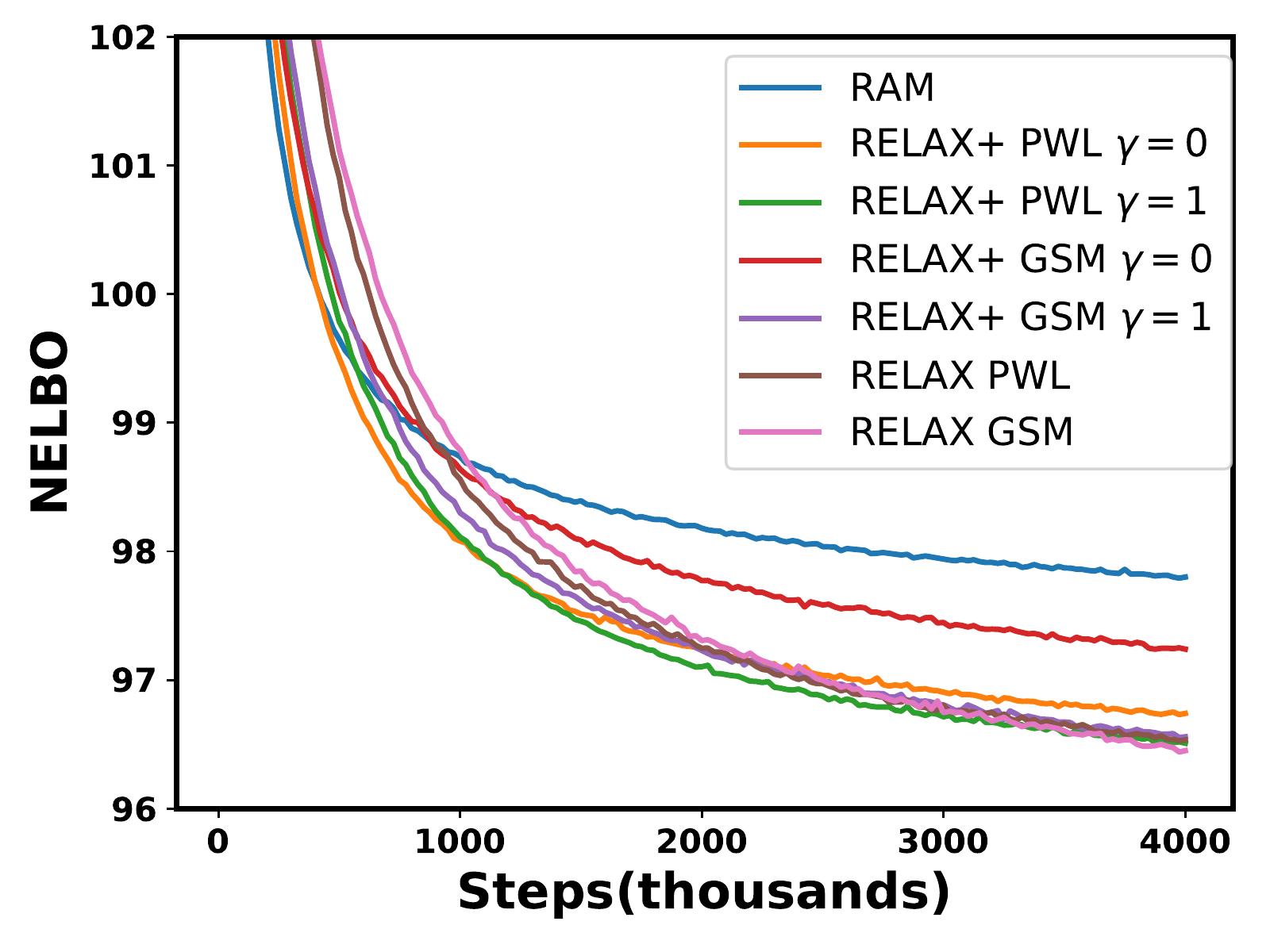}}
  \caption{MNIST training with RELAX using GSM and PWL relaxations: (a) the linear architecture. (b) the non-linear architecture.}
 \label{fig:bin_vae_relax_betas}
\end{figure}

We also compare RELAX and RELAX+ estimators. We use 1 fully connected layer with 200 hidden units, $\tanh$ nonlinearity and residual connection to produce $g^i_\psi(\bzeta)$. We used fixed $\beta=2$ for RELAX+ and trainable $\beta$ for RELAX.  Fig.~\ref{fig:bin_vae_relax_betas} shows the results on MNIST. We compare RELAX with PWL and GSM relaxations, RELAX+ with the same relaxations and with $\gamma=0,1$. We see that PWL relxations performs better for RELAX+ but worse for RELAX. In the case of the linear model RELAX+ with noise set to zero $\gamma=0$ performs significantly better then original RELAX, and almost on par with RAM. In the nonlinear case RELAX+ with $\gamma=1$ performs better then $\gamma=0$ indicating that additional noise in the gradients helps optimization.

\begin{figure}
\vspace{-0.5cm}
\centering
  \subfloat[$200{\rm H}-784{\rm V}$]{\includegraphics[scale=0.25]{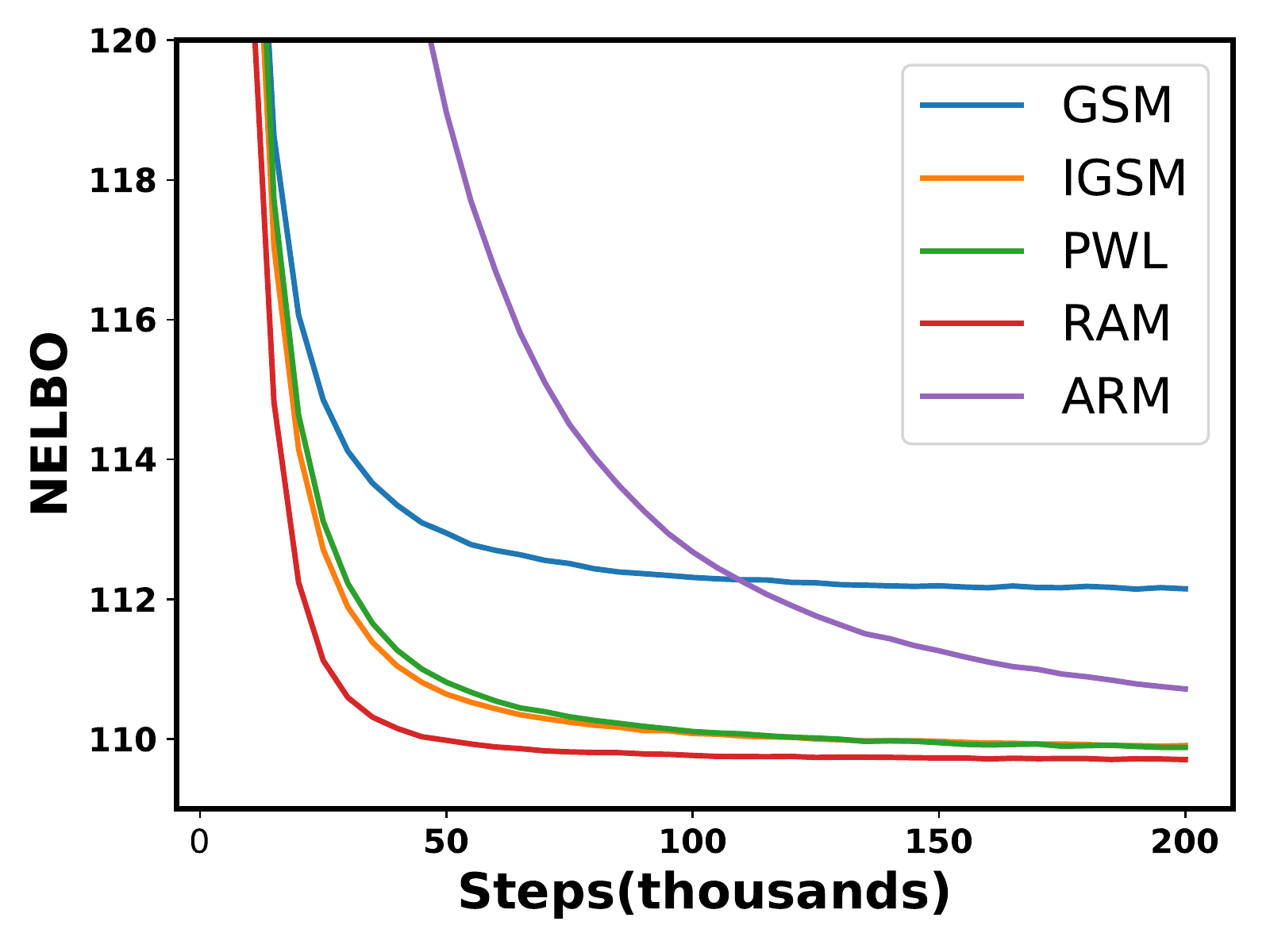}}
  \subfloat[$200{\rm H}\sim784{\rm V}$]{\includegraphics[scale=0.25]{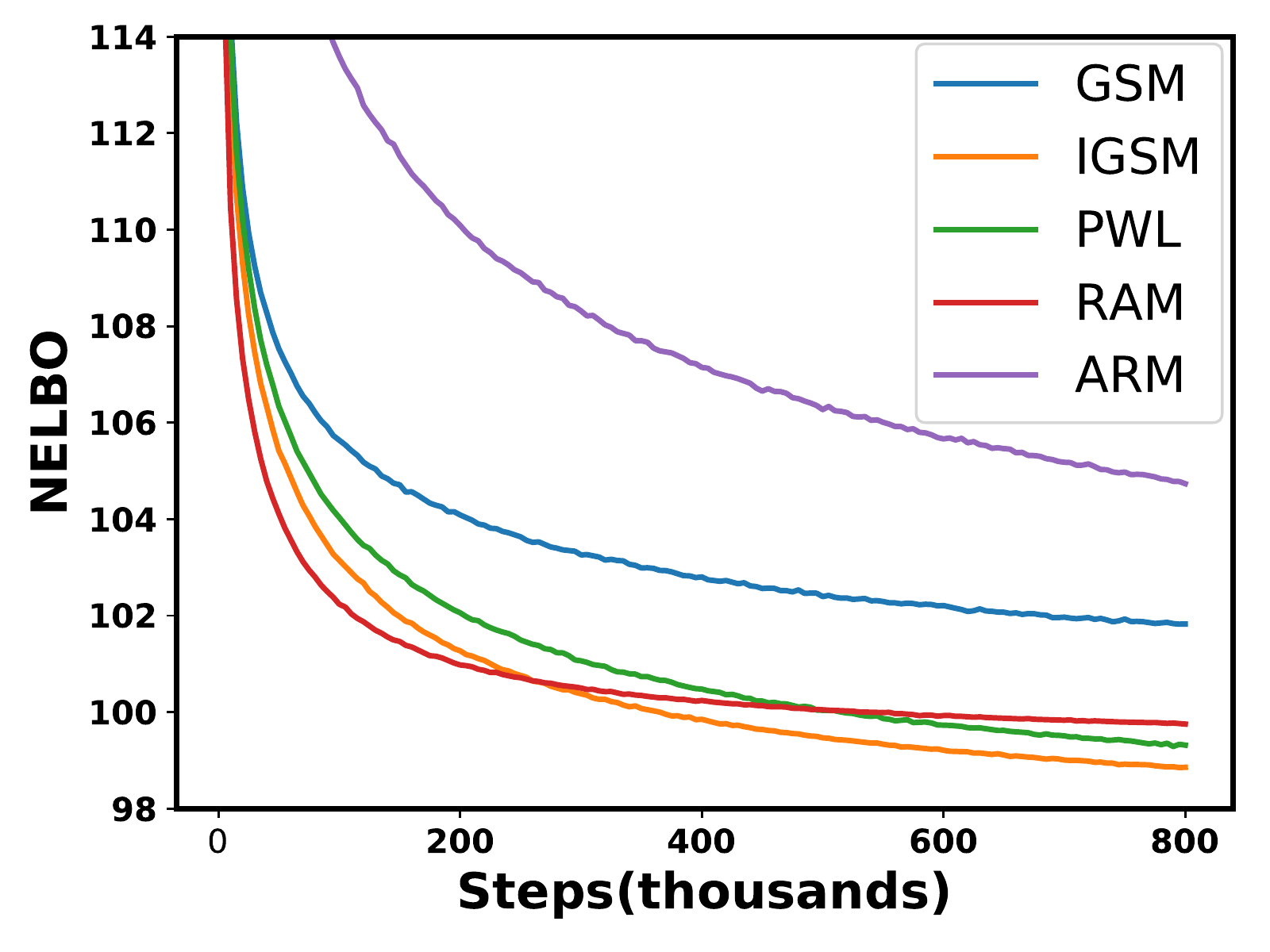}}
  \caption{Training the with fixed pre-trained decode on MNIST: (a) the linear architecture. (b) the non-linear architecture.}
 \label{fig:bin_vae_enc_beta2}
\end{figure}

During training of VAEs, the encoder and decoder models interact in complex ways. To eliminate these effects to more directly identify the impact of better gradients, we use a fixed pre-trained decoder, 
and minimize Eq.~(\ref{eq:elbo}) with respect to encoder parameters $\bphi$ only. As shown in Fig.~\ref{fig:bin_vae_enc_beta2}, the conclusions from joint training remain unchanged: GSM underperforms due to its bias. Interestingly however, RAM still overfits early in training for the $200{\rm H}\sim784{\rm V}$ model, showing that better gradients can negatively affect optimization. 

%As in \cite{jang2016categorical, maddison2016concrete} these experiments use $\beta=2$ which, as %we have seen, results in a bias for GSM towards higher entropy. In Appendix~\ref{app:plots} we %consider up $\beta$ up to 10 where the GSM bias disappears. We observe that the entropy %regularizing benefits of the GSM bias can also be achieved with explicit entropy regularization %in the objective. 

Additional results for other VAE models and for categorical variables on both MNIST and Omniglot are presented in Appendix \ref{app:vaes} with similar conclusions. 

\subsection{Optimization}

Lastly, we apply gradient estimators for discrete optimization. Following \cite{patish2018cakewalk} we study the NP-hard task of finding a maximal clique in a graph. To model this problem, the binary variable $z_i=1/0$ indicates the presence/absence of vertex $i$ in a maximal clique. If $A_{i,j}$ is the adjacency matrix of the graph and $d = \sum_i z_i$ is the size of the clique, then the objective function considered in \cite{patish2018cakewalk} is
\be
f(z) = -\frac{\sum_{ij} z_i A_{ij} z_j}{d(d-1 + \kappa)},
\ee
where $\kappa \in [0,1]$ is a hyperparameter. \cite{patish2018cakewalk} use factorial $q_\bphi(z)$ to  minimize ${\cal L}[\bphi] = \sum_\z q_\bphi(\z) f(\z)$. At the end of optimization $q_\bphi(z)$ typically collapses to a single state corresponding to a clique. Most often this clique is a local minimum and not the maximal clique. 

\begin{figure}
\vspace{-0.5cm}
\centering
  \subfloat[$\kappa=0.1$]{\includegraphics[scale=0.25]{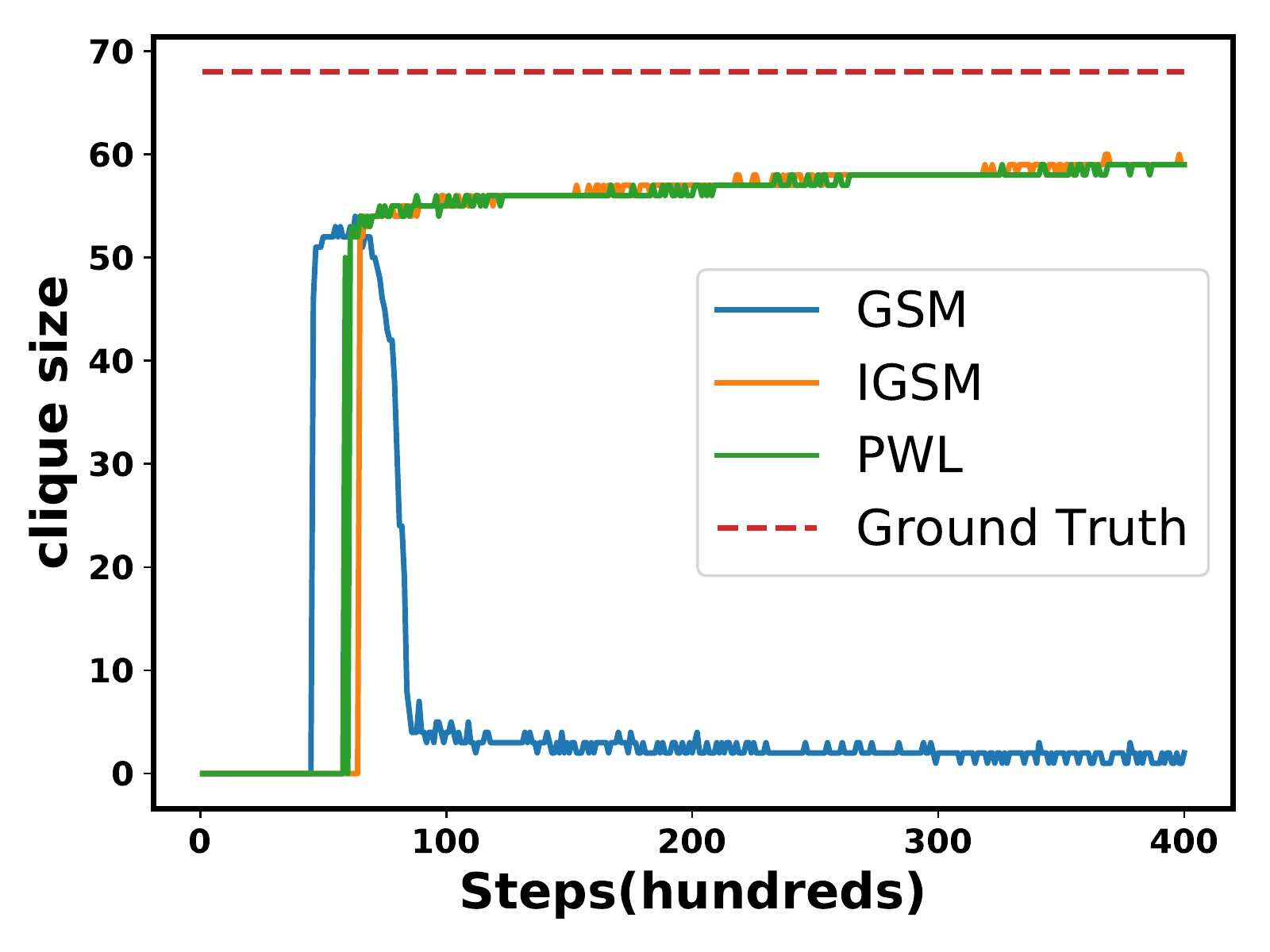}}
  \subfloat[$\kappa=0.9$]{\includegraphics[scale=0.25]{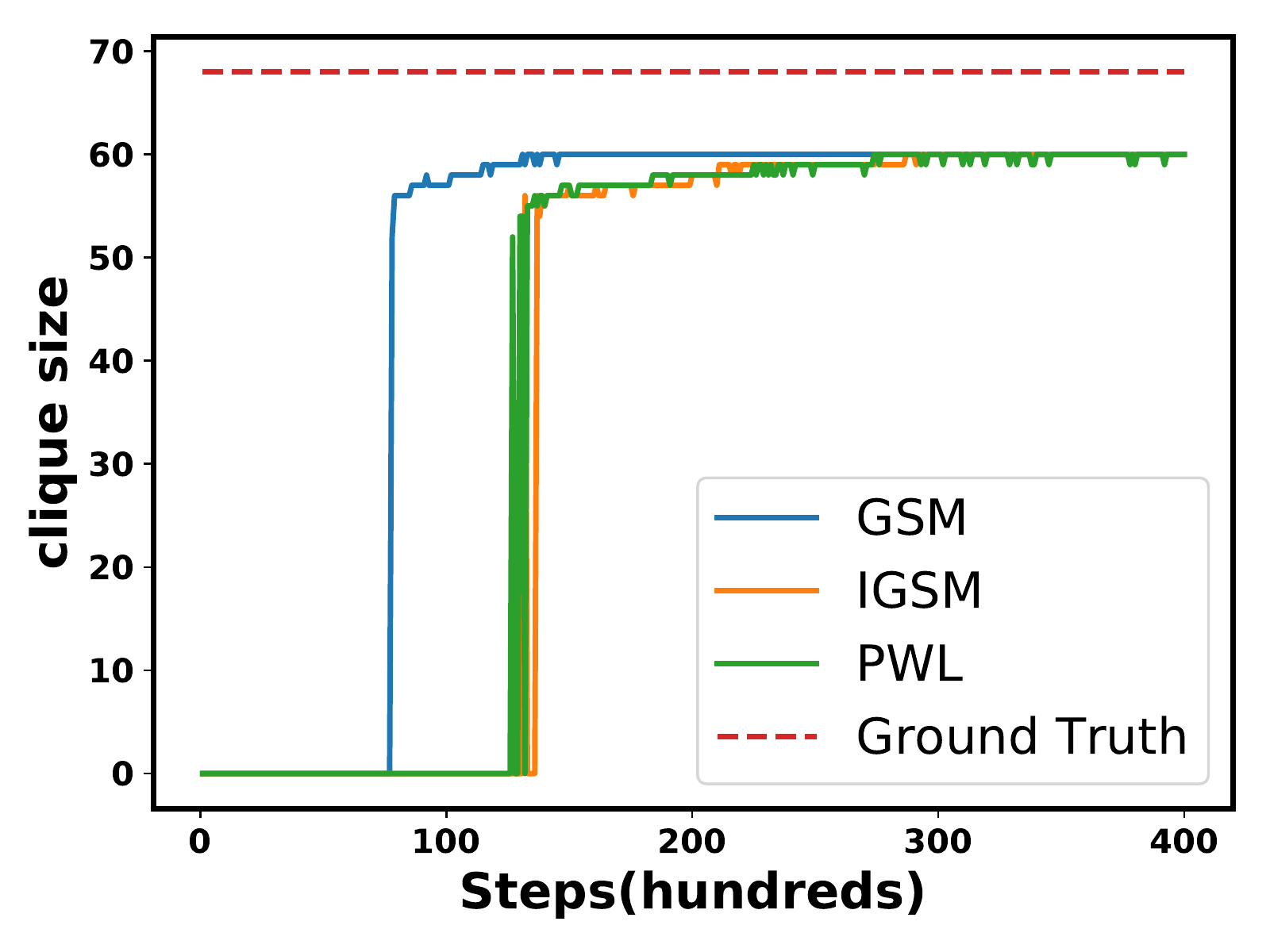}}
  \subfloat[Final max clique size]{\includegraphics[scale=0.25]{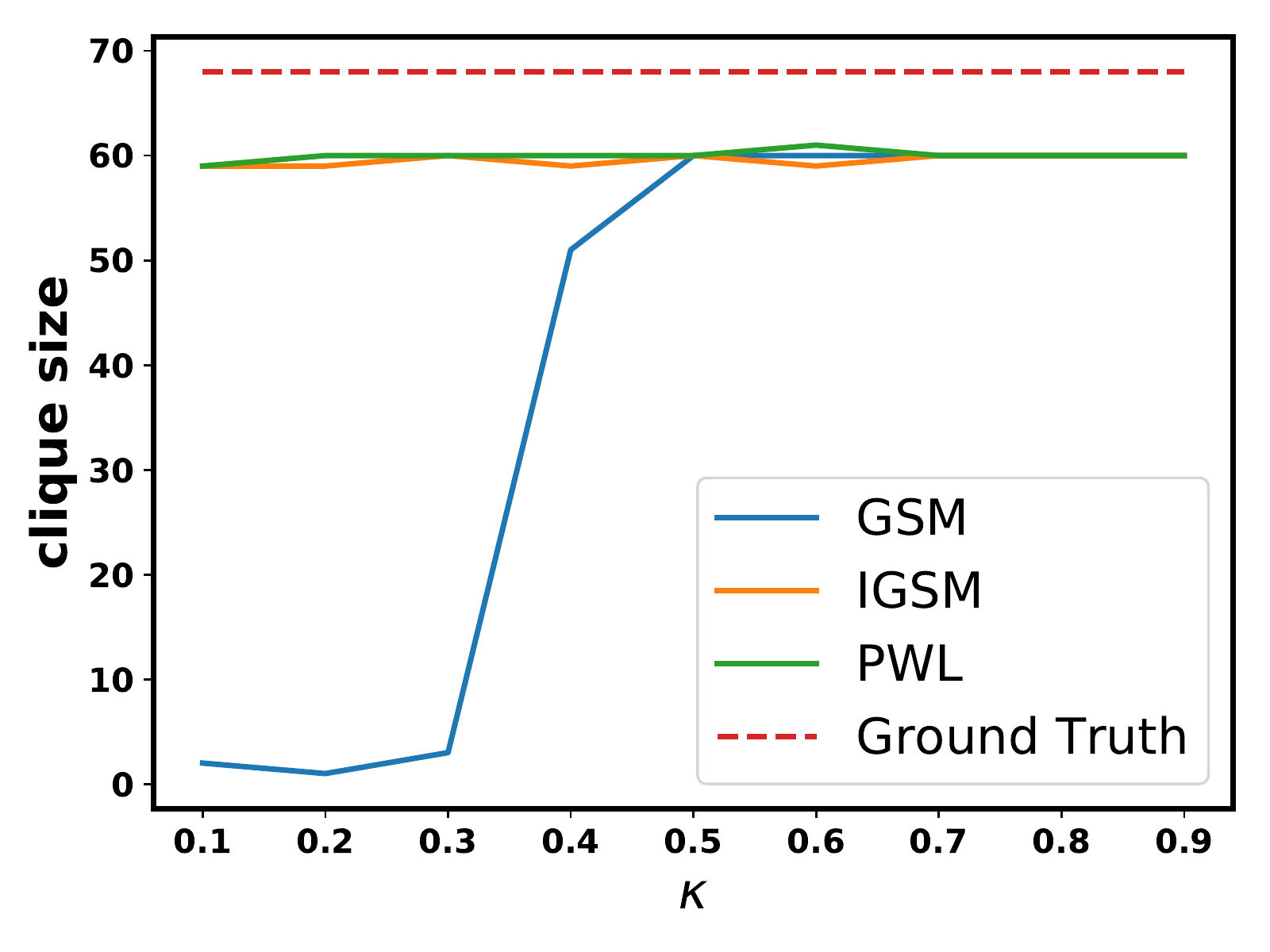}}
  \caption{Finding maximal cliques in the $C1000.9$ graph: (a) and (b) show the course of optimization for two settings of $\kappa$ and different gradient estimators. (c) plots the final maximal clique size across $\kappa$ for the estimators.}
 \label{fig:maxclique_beta2}
\end{figure}

We illustrate the performance of CR estimators on a particular graph (1000 nodes, ~450000 edges), labeled $C1000.9$ from the DIMACS data set \cite{johnson1996cliques}.  We minimize $f(\z)$ using Adam with default settings, learning rate $0.01$ for $40000$ iterations. We perform $1000$ minimizations in parallel and choose the best clique found at each iteration. Fig.~\ref{fig:maxclique_beta2}(a),(b) show the size of clique found by each of the estimators for two values of the hyperparameter $\kappa=0.1, 0.9$. At $\kappa=0.1$ the bias of GSM causes trapping in the wrong minima but ICR converges to a good local minimum. In contrast, at $\kappa=0.9$ the GSM bias accelerates convergence to a good local minimum. Fig.~\ref{fig:maxclique_beta2}(c) shows the dependence of the final clique size on $\kappa$; unbiased estimators provide more stable results.

\section{Conclusion}

We have reviewed several finite-difference (FD), continuous relaxation (CR) and Score Function (SF) estimators. FD estimators like RAM  \cite{tokui2017evaluating}, ARM \cite{yin2018arm} and ARGMAX \cite{lorberbom2018direct} often require multiple function evaluation to estimate the derivative. On the other hand, CR estimators, like Gumbel-Softmax(GSM) \cite{jang2016categorical},\cite{maddison2016concrete}, require a single pass through the objective function and can be computed efficiently. SF estimators, like \cite{tucker2017rebar}, \cite{grathwohl2017backpropagation}, give unbiased but generally higher variance gradients. We compared FD estimators theoretically and empirically and showed that ARGMAX is strictly suboptimal to RAM, while ARM has significantly higher variance. We then proposed a less expensive version of RAM estimator that requires order of magnitude fewer function evaluations with a controllable decrease in performance. CR methods provide lower variance but biased gradients. We analyzed the nature of the bias introduced by CR estimators and proposed a way to reduce it. This gives rise to improved CR (ICR) estimators, like improved GSM and piece-wise linear. These ICR estimators are unbiased for a single variable and less biased for many variables. The advantage of ICR estimators were explored in variational inference and optimization tasks. Finally, we derived a connection between REBAR and ICR estimators and proposed a lower variance version of RELAX estimator. Experiments on VAE training and discrete optimization confirm theoretical predictions and illustrate the advantages of lower-bias estimators. 

%\subsubsection*{Acknowledgments}
%Use unnumbered third level headings for the acknowledgments. All
%acknowledgments, including those to funding agencies, go at the end of the paper.

\bibliography{references}

\begin{thebibliography}{26}
\providecommand{\natexlab}[1]{#1}
\providecommand{\url}[1]{\texttt{#1}}
\expandafter\ifx\csname urlstyle\endcsname\relax
  \providecommand{\doi}[1]{doi: #1}\else
  \providecommand{\doi}{doi: \begingroup \urlstyle{rm}\Url}\fi

\bibitem[Bengio et~al.(2013)Bengio, L{\'e}onard, and
  Courville]{bengio2013estimating}
Yoshua Bengio, Nicholas L{\'e}onard, and Aaron Courville.
\newblock Estimating or propagating gradients through stochastic neurons for
  conditional computation.
\newblock \emph{arXiv preprint arXiv:1308.3432}, 2013.

\bibitem[Bowman et~al.(2016)Bowman, Vilnis, Vinyals, Dai, Jozefowicz, and
  Bengio]{bowman2016generating}
Samuel~R Bowman, Luke Vilnis, Oriol Vinyals, Andrew Dai, Rafal Jozefowicz, and
  Samy Bengio.
\newblock Generating sentences from a continuous space.
\newblock In \emph{Proceedings of The 20th SIGNLL Conference on Computational
  Natural Language Learning}, pp.\  10--21, 2016.

\bibitem[Glynn(1990)]{glynn1990likelihood}
Peter~W Glynn.
\newblock Likelihood ratio gradient estimation for stochastic systems.
\newblock \emph{Communications of the ACM}, 33\penalty0 (10):\penalty0 75--84,
  1990.

\bibitem[Grathwohl et~al.(2017)Grathwohl, Choi, Wu, Roeder, and
  Duvenaud]{grathwohl2017backpropagation}
Will Grathwohl, Dami Choi, Yuhuai Wu, Geoff Roeder, and David Duvenaud.
\newblock Backpropagation through the void: Optimizing control variates for
  black-box gradient estimation.
\newblock \emph{arXiv preprint arXiv:1711.00123}, 2017.

\bibitem[Gregor et~al.(2013)Gregor, Danihelka, Mnih, Blundell, and
  Wierstra]{gregorICML14}
Karol Gregor, Ivo Danihelka, Andriy Mnih, Charles Blundell, and Daan Wierstra.
\newblock Deep autoregressive networks.
\newblock \emph{arXiv preprint arXiv:1310.8499}, 2013.

\bibitem[Gu et~al.(2015)Gu, Levine, Sutskever, and Mnih]{gu2015muprop}
Shixiang Gu, Sergey Levine, Ilya Sutskever, and Andriy Mnih.
\newblock Muprop: Unbiased backpropagation for stochastic neural networks.
\newblock \emph{arXiv preprint arXiv:1511.05176}, 2015.

\bibitem[Jang et~al.(2016)Jang, Gu, and Poole]{jang2016categorical}
Eric Jang, Shixiang Gu, and Ben Poole.
\newblock Categorical reparameterization with {G}umbel-{S}oftmax.
\newblock \emph{arXiv preprint arXiv:1611.01144}, 2016.

\bibitem[Johnson \& Trick(1996)Johnson and Trick]{johnson1996cliques}
David~S Johnson and Michael~A Trick.
\newblock \emph{Cliques, coloring, and satisfiability: second DIMACS
  implementation challenge, October 11-13, 1993}, volume~26.
\newblock American Mathematical Soc., 1996.

\bibitem[Kingma \& Ba(2014)Kingma and Ba]{kingma2014adam}
Diederik Kingma and Jimmy Ba.
\newblock Adam: A method for stochastic optimization.
\newblock \emph{arXiv preprint arXiv:1412.6980}, 2014.

\bibitem[Kingma \& Welling(2013)Kingma and Welling]{kingmaICLR15}
Diederik~P Kingma and Max Welling.
\newblock Auto-encoding variational bayes.
\newblock \emph{arXiv preprint arXiv:1312.6114}, 2013.

\bibitem[Lorberbom et~al.(2018)Lorberbom, Gane, Jaakkola, and
  Hazan]{lorberbom2018direct}
Guy Lorberbom, Andreea Gane, Tommi Jaakkola, and Tamir Hazan.
\newblock Direct optimization through $\arg \max$ for discrete variational
  auto-encoder.
\newblock \emph{arXiv preprint arXiv:1806.02867}, 2018.

\bibitem[Maddison et~al.(2016)Maddison, Mnih, and Teh]{maddison2016concrete}
Chris~J Maddison, Andriy Mnih, and Yee~Whye Teh.
\newblock The concrete distribution: A continuous relaxation of discrete random
  variables.
\newblock \emph{arXiv preprint arXiv:1611.00712}, 2016.

\bibitem[Mnih \& Gregor(2014)Mnih and Gregor]{mnih2014neural}
Andriy Mnih and Karol Gregor.
\newblock Neural variational inference and learning in belief networks.
\newblock \emph{arXiv preprint arXiv:1402.0030}, 2014.

\bibitem[Mnih \& Rezende(2016)Mnih and Rezende]{mnih2016variational}
Andriy Mnih and Danilo Rezende.
\newblock Variational inference for {M}onte {C}arlo objectives.
\newblock In \emph{International Conference on Machine Learning}, pp.\
  2188--2196, 2016.

\bibitem[Patish \& Ullman(2018)Patish and Ullman]{patish2018cakewalk}
Uri Patish and Shimon Ullman.
\newblock Cakewalk sampling.
\newblock \emph{arXiv preprint arXiv:1802.09030}, 2018.

\bibitem[Raiko et~al.(2014)Raiko, Berglund, Alain, and
  Dinh]{raiko2014techniques}
Tapani Raiko, Mathias Berglund, Guillaume Alain, and Laurent Dinh.
\newblock Techniques for learning binary stochastic feedforward neural
  networks.
\newblock \emph{arXiv preprint arXiv:1406.2989}, 2014.

\bibitem[Rezende et~al.(2014)Rezende, Mohamed, and
  Wierstra]{rezende2014stochastic}
Danilo~Jimenez Rezende, Shakir Mohamed, and Daan Wierstra.
\newblock Stochastic backpropagation and approximate inference in deep
  generative models.
\newblock In \emph{International Conference on Machine Learning}, pp.\
  1278--1286, 2014.

\bibitem[Rolfe(2016)]{rolfe2016discrete}
Jason~Tyler Rolfe.
\newblock Discrete variational autoencoders.
\newblock \emph{arXiv preprint arXiv:1609.02200}, 2016.

\bibitem[Schulman et~al.(2015)Schulman, Heess, Weber, and
  Abbeel]{schulman2015gradient}
John Schulman, Nicolas Heess, Theophane Weber, and Pieter Abbeel.
\newblock Gradient estimation using stochastic computation graphs.
\newblock In \emph{Advances in Neural Information Processing Systems}, pp.\
  3528--3536, 2015.

\bibitem[Titsias \& L{\'a}zaro-Gredilla(2015)Titsias and
  L{\'a}zaro-Gredilla]{aueb2015local}
Michalis Titsias and Miguel L{\'a}zaro-Gredilla.
\newblock Local expectation gradients for black box variational inference.
\newblock In \emph{Advances in neural information processing systems}, pp.\
  2638--2646, 2015.

\bibitem[Tokui \& Sato(2017)Tokui and Sato]{tokui2017evaluating}
Seiya Tokui and Issei Sato.
\newblock Evaluating the variance of likelihood-ratio gradient estimators.
\newblock In \emph{International Conference on Machine Learning}, pp.\
  3414--3423, 2017.

\bibitem[Tucker et~al.(2017)Tucker, Mnih, Maddison, Lawson, and
  Sohl-Dickstein]{tucker2017rebar}
George Tucker, Andriy Mnih, Chris~J Maddison, John Lawson, and Jascha
  Sohl-Dickstein.
\newblock Rebar: Low-variance, unbiased gradient estimates for discrete latent
  variable models.
\newblock In \emph{Advances in Neural Information Processing Systems}, pp.\
  2624--2633, 2017.

\bibitem[Vahdat et~al.(2018{\natexlab{a}})Vahdat, Andriyash, and
  Macready]{vahdat2018dvae}
Arash Vahdat, Evgeny Andriyash, and William~G Macready.
\newblock {DVAE}\#: Discrete variational autoencoders with relaxed {B}oltzmann
  priors.
\newblock In \emph{Neural Information Processing Systems (NIPS)},
  2018{\natexlab{a}}.

\bibitem[Vahdat et~al.(2018{\natexlab{b}})Vahdat, Macready, Bian, Khoshaman,
  and Andriyash]{VahdatMBKA18}
Arash Vahdat, William~G. Macready, Zhengbing Bian, Amir Khoshaman, and Evgeny
  Andriyash.
\newblock {DVAE}++: Discrete variational autoencoders with overlapping
  transformations.
\newblock In \emph{International Conference on Machine Learning (ICML)},
  2018{\natexlab{b}}.

\bibitem[Williams(1992)]{williams1992simple}
Ronald~J Williams.
\newblock Simple statistical gradient-following algorithms for connectionist
  reinforcement learning.
\newblock In \emph{Reinforcement Learning}, pp.\  5--32. Springer, 1992.

\bibitem[Yin \& Zhou(2018)Yin and Zhou]{yin2018arm}
Mingzhang Yin and Mingyuan Zhou.
\newblock Arm: Augment-reinforce-merge gradient for discrete latent variable
  models.
\newblock \emph{arXiv preprint arXiv:1807.11143}, 2018.

\end{thebibliography}
\bibliographystyle{iclr2019_conference}

\appendix

\section{Improved CR estimator for Bayesian network}\label{app:icr_bayes}

In this appendix we derive the improved CR estimator for a hierarchical $q_\bphi(\z) = \prod_i q_{\bphi,i}(z_i | \z_{<i})$. For simplicity, we do this for two variables $q_\bphi(z_1,z_2) = q_{\bphi,2}(z_2 | z_1) q_{\bphi,1}(z_1)$ with the gradient given in Eq.~(\ref{eq:grad_bayes}):
\begin{align*}
\p_{\bphi} {\cal L} =& \, \p_{\bphi} q_{\bphi,1}  \left[ \sum_{z_2} q_{\bphi,2}(z_2|1) f(1,z_2) -  \sum_{z_2} q_{\bphi,2}(z_2|0) f(0,z_2) \right]  \\  & \, + \sum_{z_1} q_{\bphi,1}(z_1) \p_{\bphi} q_{\bphi,2}(1|z_1)  \left[f(z_1,1) -f(z_1,0) \right]
\end{align*}
where $q_{\bphi,1}=q_{\bphi,1}(z_1=1)$. We denote the two contributions to $\partial_\bphi \mathcal{L}$ by $\p_{\bphi} {\cal L}^{(1,2)}$ and determine them separately. Using the integral trick (\ref{eq:grad_1_int}) the second contribution can be written as 
\be
\p_{\bphi} {\cal L}^{(2)} =  \sum_{z_1} q_{\bphi,1}(z_1) \p_{\bphi} q_{\bphi,2}(1|z_1)  \int d \rho_1 \frac{\p \zeta_2}{\p \rho_2} \p_{\zeta_2}  f(z_1,\zeta_2)  \approx \E_{\brho} \left[  \p_{\bphi} q_{\bphi,2}(1|\zeta_1)  \frac{\p \zeta_2}{\p \rho_2} \p_{\zeta_2}  f(\zeta_1,\zeta_2) \right]
\ee
where $\zeta_2 = g(\rho_2,  q_2(1| \zeta_1))$ and $\zeta_1 = g(\rho_1,  q_1)$. The first term can be handled similarly:
\be
\p_{\bphi} {\cal L}^{(1)} = \p_{\bphi} q_{\bphi,1}  \int d \rho_1 \frac{\p \zeta_1}{\p \rho_1} \p_{\zeta_1} \left(\sum_{z_2}  q_{\bphi,2}(z_2|\zeta_1) f(\zeta_1, z_2) \right) \approx \E_{\brho} \left[  \p_{\phi} q_{\bphi,1}  \frac{\p \zeta_1}{\p \rho_1} \p_{\zeta_1} f(\zeta_1, \zeta_2) \right]
\ee
Combining these contributions we arrive at an expression very similar to the reparameterization trick with the replacement $ \p_{q_i} \zeta_i \to \p_{\rho_i} \zeta_i$:
\be
\p_{\bphi} {\cal L} \approx \p_\bphi \E_{\brho} \left[ f(\zeta_1, \zeta_2) \right]_{\p_{q_i} \zeta_i \to \p_{\rho_i} \zeta_i}
\ee

\section{Categorical PWL estimator}\label{app:pwl_cat}

Here, we give derive the PWL estimator for categorical variables. The derivative for a single categorical variable $y = (y^0,...y^{A-1})$ with $y^a \in \{0,1\}$ and $\sum_a y^a = 1$ is given in Eq.~(\ref{eq:grad_cat}) as
\be\label{eq:grad_cat_app}
\p_{\bphi} {\cal L} =\p_{\bphi}  \sum_{y} q_\bphi(y) f(y) = \sum_a  \p_{\bphi} l^a_\bphi \sum_b q^a q^b (f^a - f^b),
\ee
We again apply the integral trick (\ref{eq:grad_1_int}) to represent the difference $f^a-f^b$. To do that we relax the variable $y$ so that it interpolates between $(y^a,y^b) = (1,0)$ and $(y^a,y^b)=(0,1)$ as $y \to y^{a,b} = \bigl\{y^a = \left[0.5+\alpha^{a,b}(\rho^{a,b} - q^b/(q^a + q^b))\right]_0^1,y^b=1 - y^a, y^{c \ne a,b}=0 \bigr\}$ where  $\rho^{a,b} \in \U[0,1]$ and $\alpha^{a,b}$ is the slope. The relaxed objective then takes the form
\be\label{eq:grad_1hot_int}
{\cal L} = \int_0^1 \prod_{a<b} d \rho^{a,b} \sum_{a<b} w^{a,b} f(y^{a,b}) ,
\ee
where $w^{a,b}$ are weights to be determined. The gradient of this relaxed objective with respect to the logit  $l^a$ is
\ba
 \p_{l^a} {\cal L} = \E_{\rho} \left[ \sum_{b\ne a} w^{a,b} \p_{\rho^{a,b}}f(y^{a,b}) \frac{q^a q^b}{(q^a + q^b)^2} \right] = \sum_{b\ne a} w^{a,b} \frac{q^a q^b}{(q^a + q^b)^2} (f^a - f^b).
\ea
Comparison with Eq.~(\ref{eq:grad_cat_app}) gives $w^{a,b} = (q^a + q^b)^2$. However, computing the sum over the edges of the simplex is prohibitively expensive, so we choose to replace it with sampling from the set of edges with probability $p^{a,b} = (q_a + q_b)/(\sum_{a<b} q^a + q^b) =  (q^a + q^b)/(A-1)$. The reason for choosing this distribution is that Eq.~(\ref{eq:grad_1hot_int}) must give correct value of the objective (not just the derivative) as the relaxation parameter $\beta \to \infty$, which requires the probability of each state $y^a=1$ to be equal to $q^a$. For relaxed edge $(a,b)$ the probability of $y^a=1$ is equal to $q^a/(q^a + q^b)$, and thus the total probability of $y^a=1$ is:
\be
\sum_{b \ne a} p^{a,b} \frac{q^a}{q^a + q^b}= \sum_{b \ne a} \frac{q^a + q^b}{A-1} \frac{q^a}{q^a + q^b} = q^a.
\ee
Finally, to get the correct weights $w^{a,b}$ we must rescale each term by a factor $\gamma^{a,b} = (A-1)(q^a + q^b)$ so that $w^{a,b} = \gamma^{a,b}  p^{a,b}$. In summary, the relaxed objective has the following form:
\be\label{eq:grad_1hot_int_sample_1}
{\cal L} = \E_{(a,b) \sim p^{a,b}} \left[   \E_{\rho \in \U[0,1]} \left[  f(\tilde y^{a,b}) \right] \right],
\ee
where $\tilde y^{a,b}$ has the same value as $y^{a,b}$ but has the gradient scaled by  $\gamma^{a,b} = (A-1)(q^a + q^b)$.

\section{Experimental details}\label{app:exp_det}
For the toy example of Section \ref{subsec:toy} we optimize ${\cal L}$ with respect to $q_\phi(z=1)$ using Adam (\cite{kingma2014adam}) for 2000 iterations with learning rate $r = 0.01$ using minibatches of size 100 to reduce the variance of gradients. We initialize $q_\phi(z=1)$ to the wrong maximum of the relaxed function by setting $q_\phi(z=1) = \sigma(5)$.

For VAE training experiments, following \cite{maddison2016concrete, jang2016categorical, tucker2017rebar}, we consider four architectures with binary variables denoted by  $200{\rm H}-784{\rm V}$, $200{\rm H}-200{\rm H}-784{\rm V}$, $200{\rm H}\sim784{\rm V}$, and $200{\rm H}\sim200{\rm H}\sim784{\rm V}$. Here $-$ denotes a linear layer, while $\sim$ denotes two layers of 200 hidden units with ${\rm tanh}$ nonlinearity and batch normalization.  In our experiments we use the Adam optimizer with default parameters and a fixed learning rate $0.0003$, run for $4\cdot 10^6$ steps with minibatches of size 100. We repeat each experiment 5 times with random seed and plot the mean.

\section{Additional Experiments}\label{app:plots}
\subsection{Binary Concave Toy Example} \label{app:binary_toy_concave}

\begin{figure}[H]
    \centering
 \subfloat[Concave relaxed function]{ \includegraphics[scale=0.25]{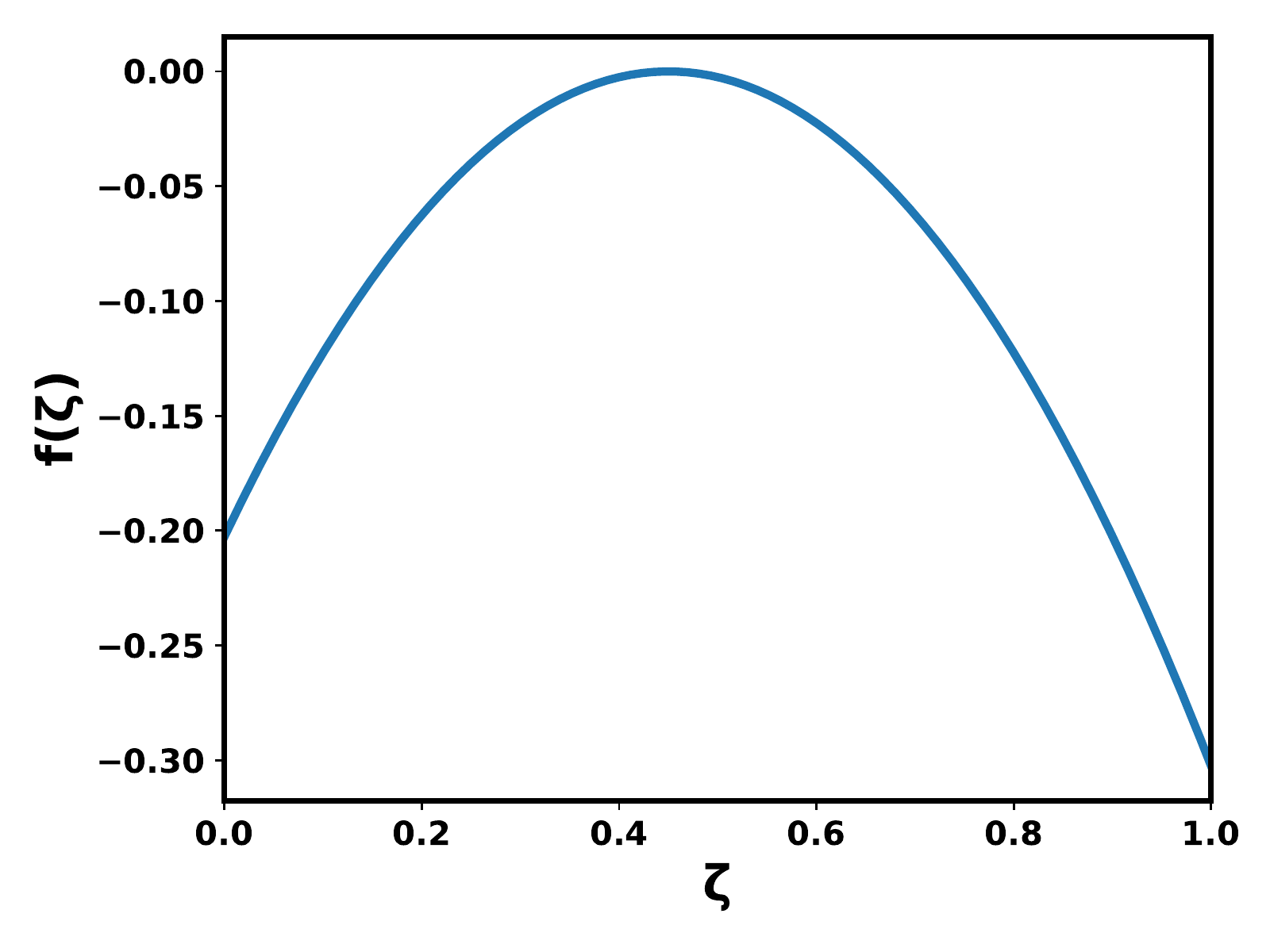}} 
  \subfloat[Probability $q(z=1)$]{ \includegraphics[scale=0.25]{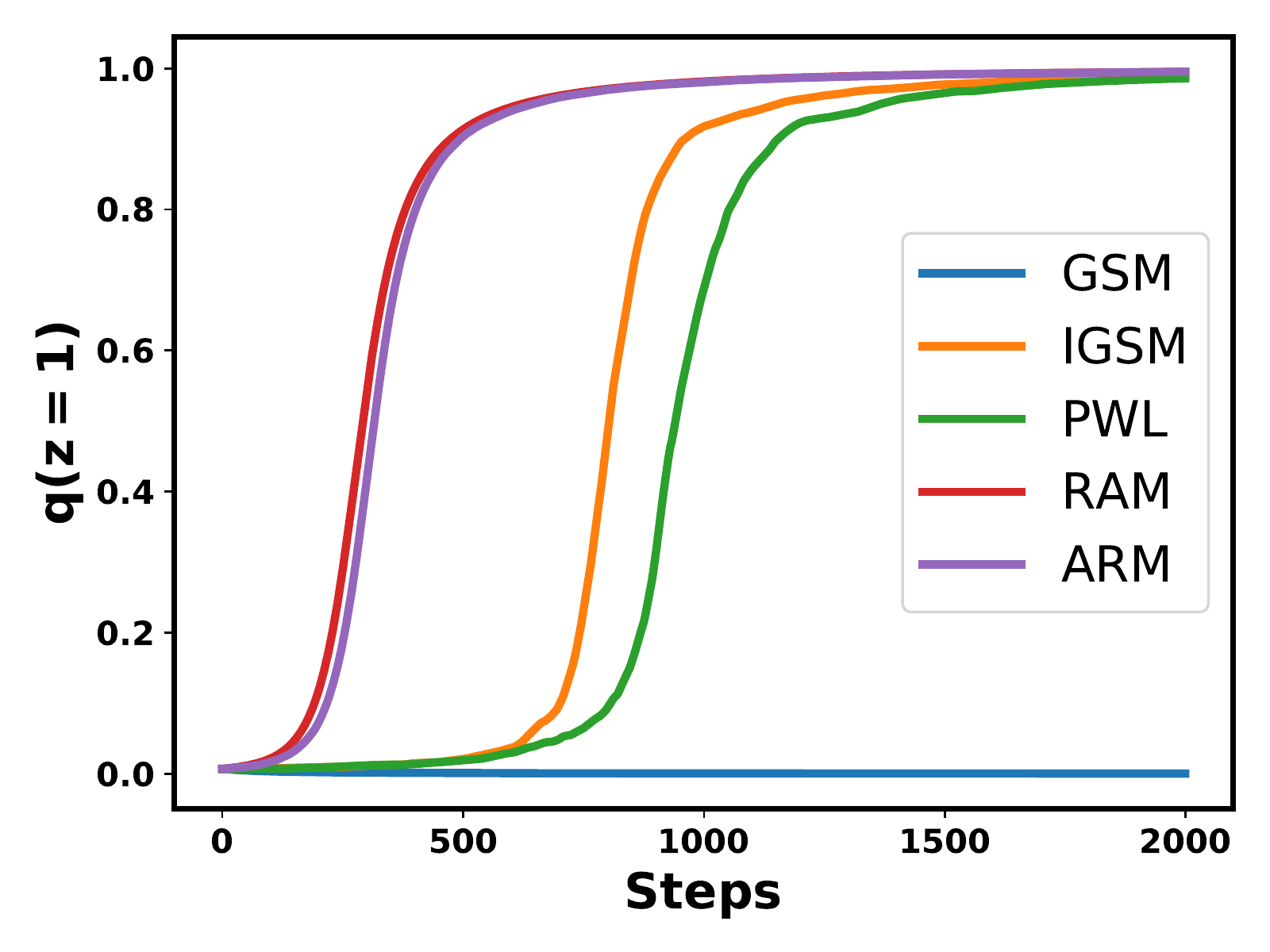}}
\caption{Single binary variable concave toy example. (a) Relaxed function $f(\zeta) = - (\zeta - 0.45)^2$ 
(b) Probability $q_\phi(z=1)$ during optimization }
    \label{fig:binary_toy_concave}
\end{figure}

We evaluate performance on the concave function $f(\zeta) = - (\zeta - 0.45)^2$
shown in Fig.~\ref{fig:binary_toy_concave}(a). 
The training setup is identical to Section \ref{subsec:toy} but with $q_\phi(z=1)$ initialized
to  $\sigma(-5)$. Fig.~\ref{fig:binary_toy_concave}(b) demonstrates that the GSM optimization gets trapped in the wrong minimum due to it's bias towards the dominant mode $\zeta=0$.

\subsection{Categorical Toy Example}  \label{app:categorical_toy_concave}

\begin{figure}[H]
    \centering
  \subfloat[Concave function]{ \includegraphics[scale=0.25]{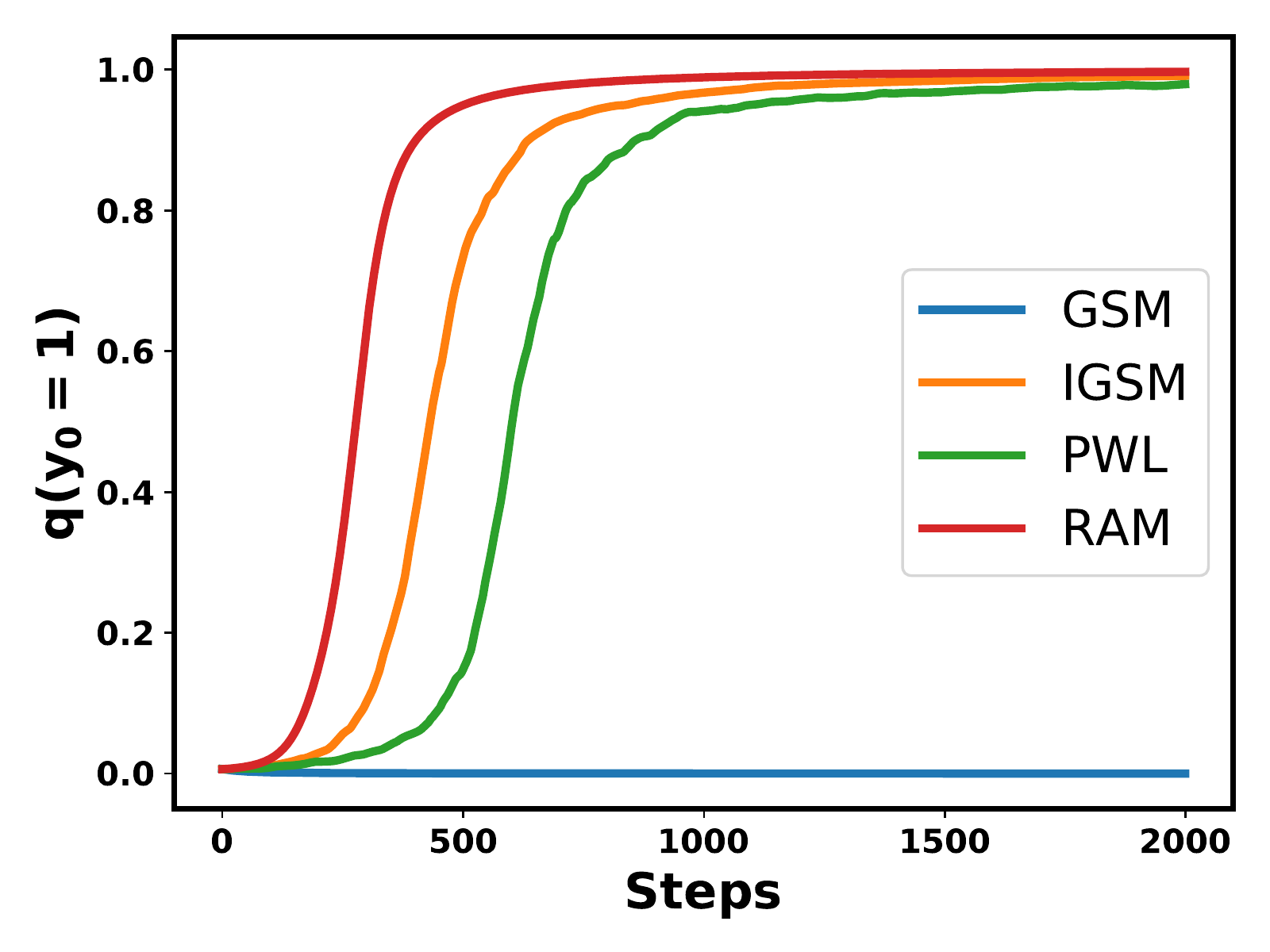}}
  \subfloat[Convex function]{ \includegraphics[scale=0.25]{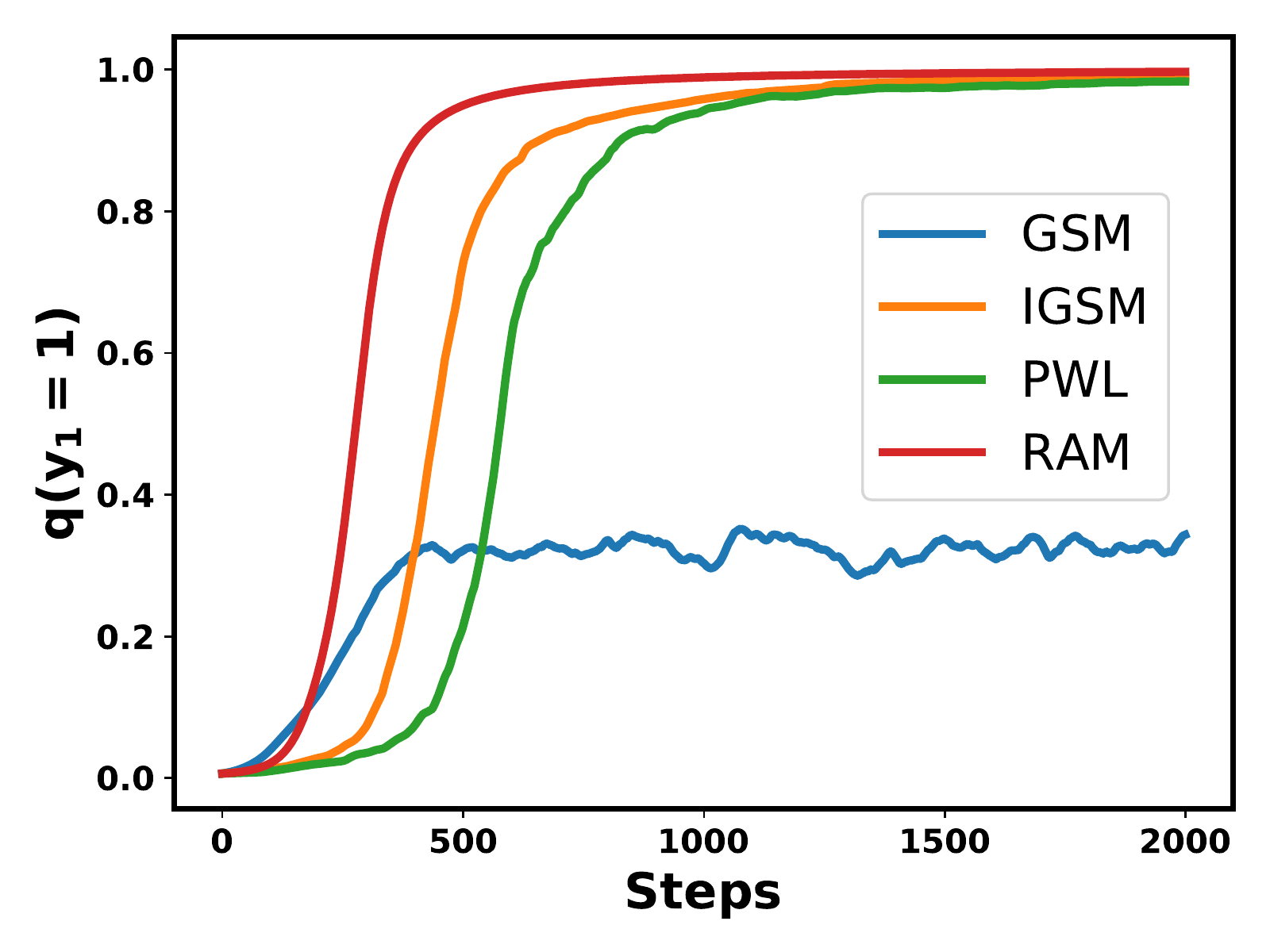}}
\caption{The probablity of the true minimum for a single categorical-variable toy example with 
(a) concave function $f(y) = - \sum_a (g^a - y^a)^2$ and (b) convex function $f(y) = \sum_a (g^a - y^a)^2$}
    \label{fig:cat_toy}
\end{figure}

We consider a categorical example with convex and concave functions of a single categorical variable $y$ having 10 values. We take $f(y) = \pm \sum_a (g^a - y^a)^2$ such that $g^0=0.9, g^1=1.1$, and $g^{i>1}=1$. The convex function has a minimum at $y^1=1$, while the concave function is minimized at $y^0=1$. We compare 4 estimators for minimizing this function: RAM of Eq.~(\ref{eq:grad_cat}),  PWL of Eq.~(\ref{eq:grad_1hot_int_sample}),  GSM ofEq.~(\ref{eq:gsm_1hot}) and IGSM of Eq.~(\ref{eq:gsm_trick_bin}). The probability of the true minimum is shown in Fig.~\ref{fig:cat_toy}. In both the convex and concave cases the GSM estimator exhibits a bias preventing it from finding the minimum. IGSM is less biased than GSM which allows it to find the true minimum. The PWL estimator is unbiased but has higher variance then IGSM which slows down its optimization.

\subsection{Discrete Variational Autoencoders} \label{app:vaes}

\begin{figure}[H]	
\centering
  \subfloat[$200{\rm H}-784{\rm V}$]{\includegraphics[scale=0.25]{bin_vae_1lin_beta2_elbo_train.pdf}}
  \subfloat[$200{\rm H}-200{\rm H}-784{\rm V}$]{\includegraphics[scale=0.25]{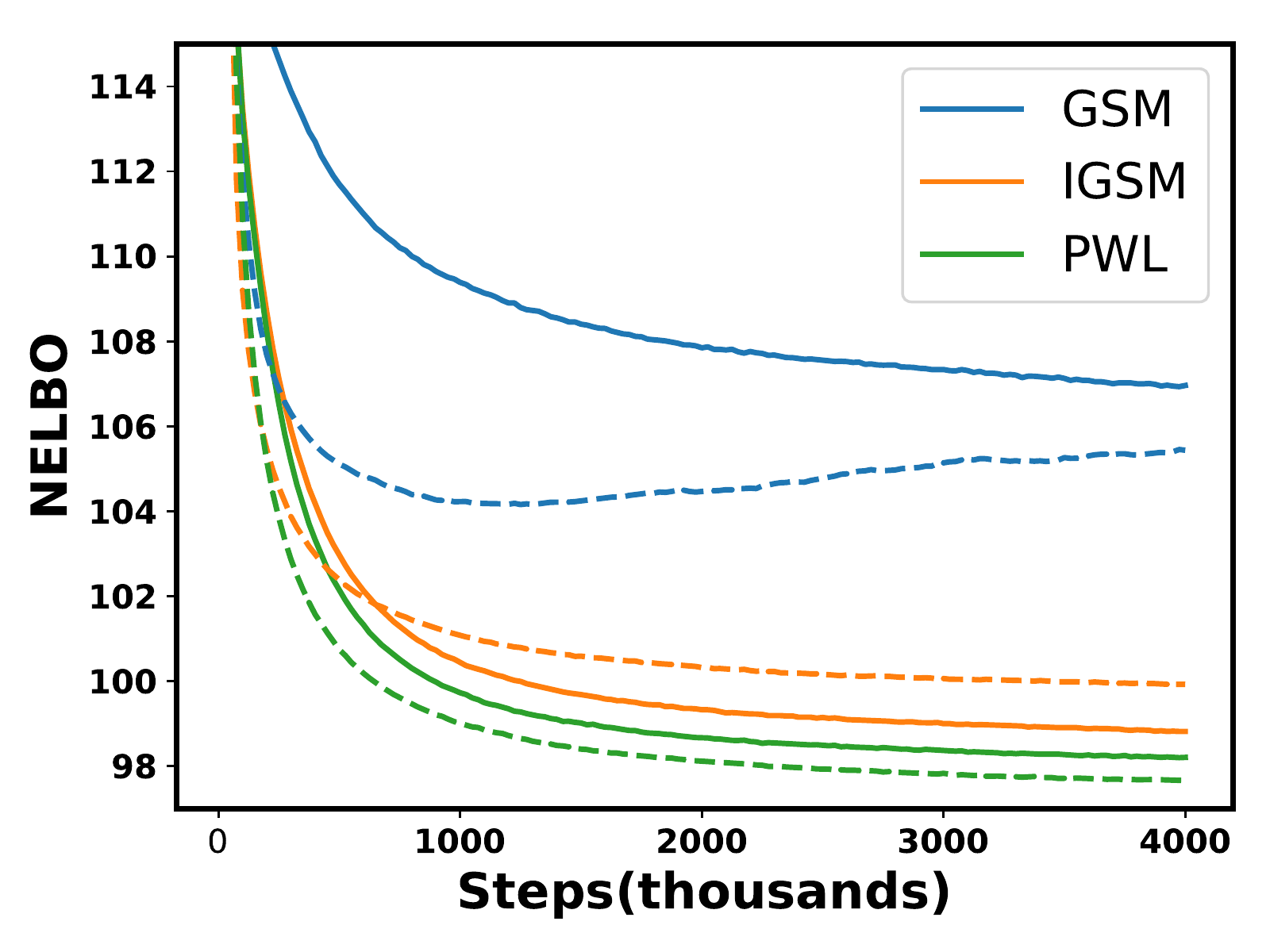}}
  \vspace{-0.4cm}
    \subfloat[$200{\rm H}\sim784{\rm V}$]{\includegraphics[scale=0.25]{bin_vae_1nonlin_beta2_elbo_train.pdf}}
      \subfloat[$200{\rm H}\sim200{\rm H}\sim784{\rm V}$]{\includegraphics[scale=0.25]{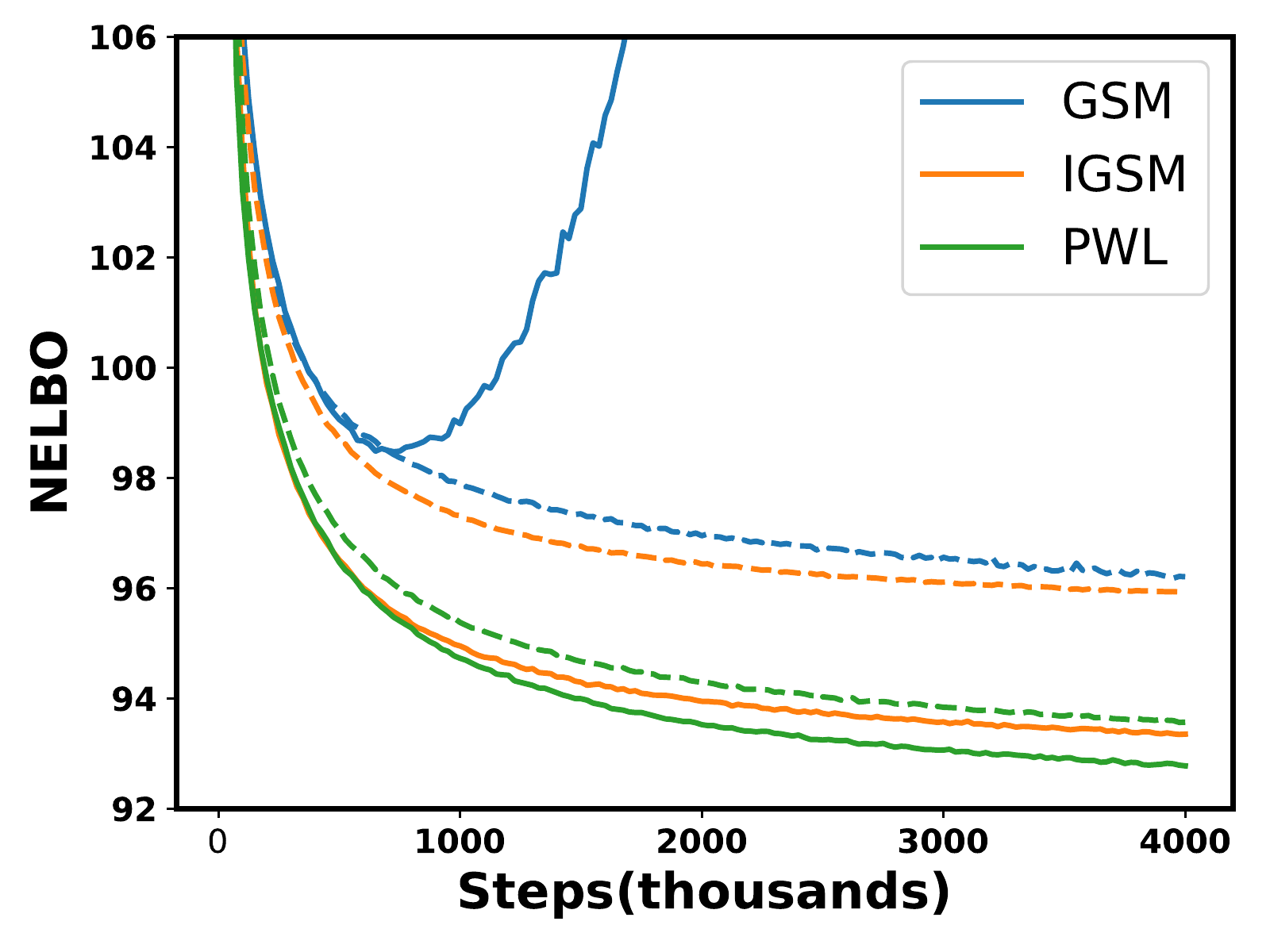}}
      \caption{MNIST training (CR estimators use $\beta=2$). Solid/dashed lines correspond to one/two-pass training.}
    \label{fig:bin_vae_elbo_beta2}
\end{figure}

Fig.~\ref{fig:bin_vae_elbo_beta2} shows the results for all 4 architectures $200{\rm H}-784{\rm V}$, $200{\rm H}-200{\rm H}-784{\rm V}$, $200{\rm H}\sim784{\rm V}$, and $200{\rm H}\sim200{\rm H}\sim784{\rm V}$ at $\beta=2$. Hierarchical models with two layers of latent units in Fig.~\ref{fig:bin_vae_elbo_beta2}(b),(d) exhibit similar trends to the factorial case considered in the main text.The GSM estimator converges to a higher NELBO in the linear case and becomes unstable in the non-linear case.

\begin{figure}[H]	
    \centering 
  \subfloat[]{ \includegraphics[scale=0.25]{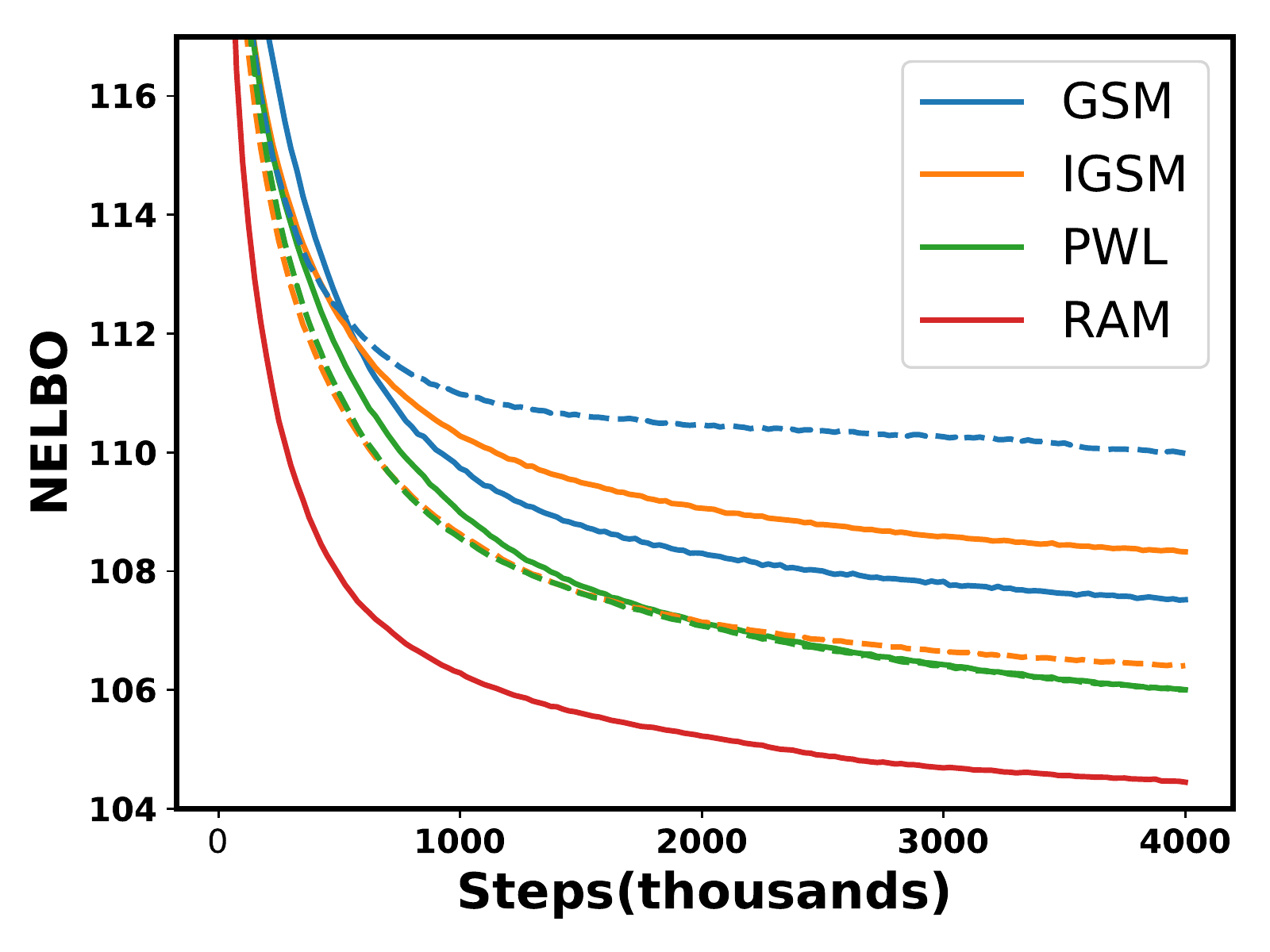}}
\caption{MNIST training on categorical linear architecture $20 \times 10{\rm H}-784{\rm V}$ (CR estimators use $\beta=2$). Solid/dashed lines correspond to one/two-pass training.}
    \label{fig:bin_vae_cat_beta2}
\end{figure}

We also compare the performance of GSM and IGSM estimators in the categorical case $20 \times 10{\rm H}-784{\rm V}$ with latent space being 20 categorical variables with 10 classes Fig.~\ref{fig:bin_vae_cat_beta2}. We see that IGSM outperforms GSM estimator when using 2-pass training. This indirectly confirms that IGSM is less biased. The PWL estimator performs best indicating the advantages ICR estimators.

\begin{figure}[H]	
\centering
  \subfloat[$200{\rm H}-784{\rm V}$]{\includegraphics[scale=0.25]{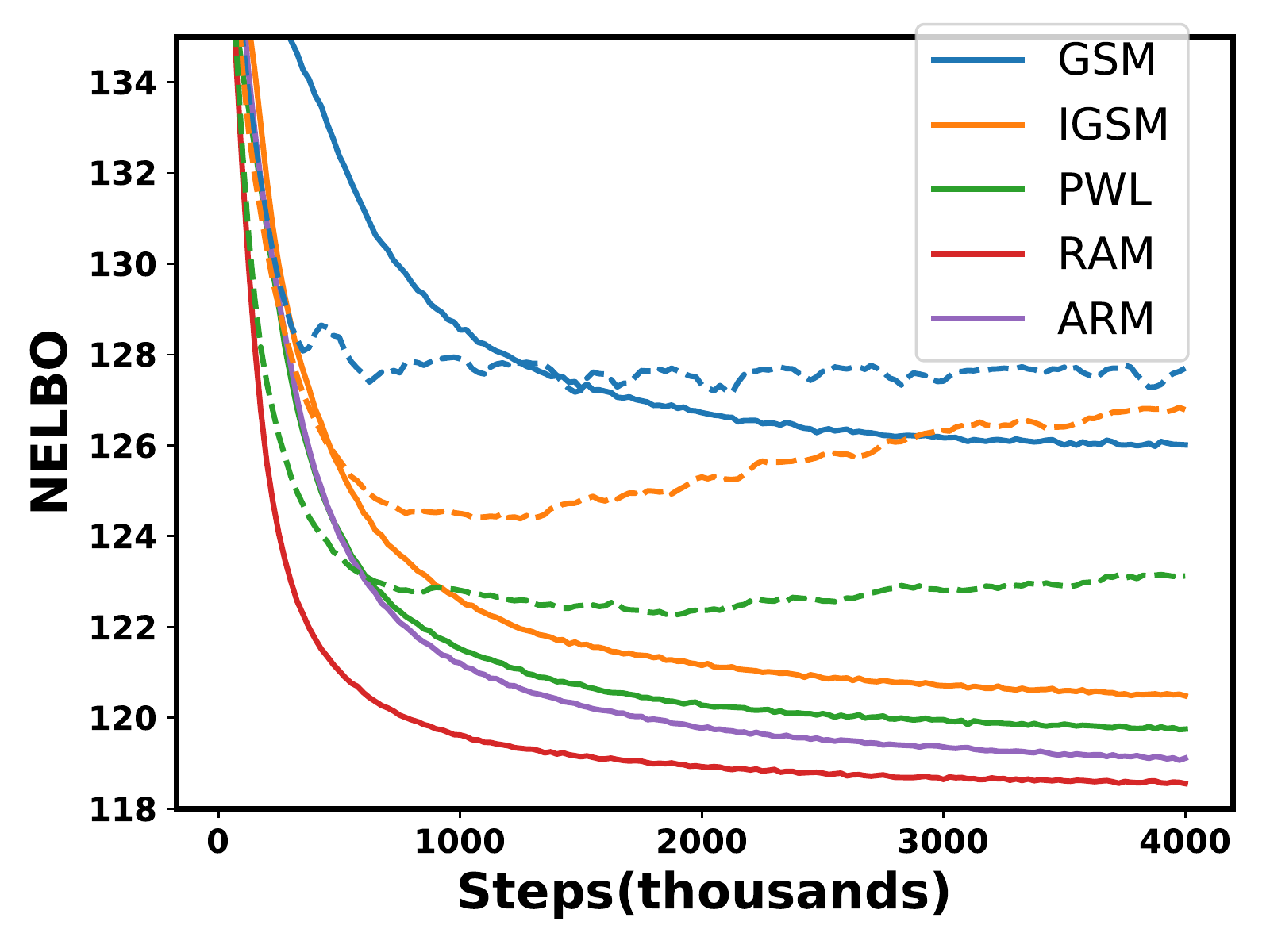}}
  \subfloat[$200{\rm H}-200{\rm H}-784{\rm V}$]{\includegraphics[scale=0.25]{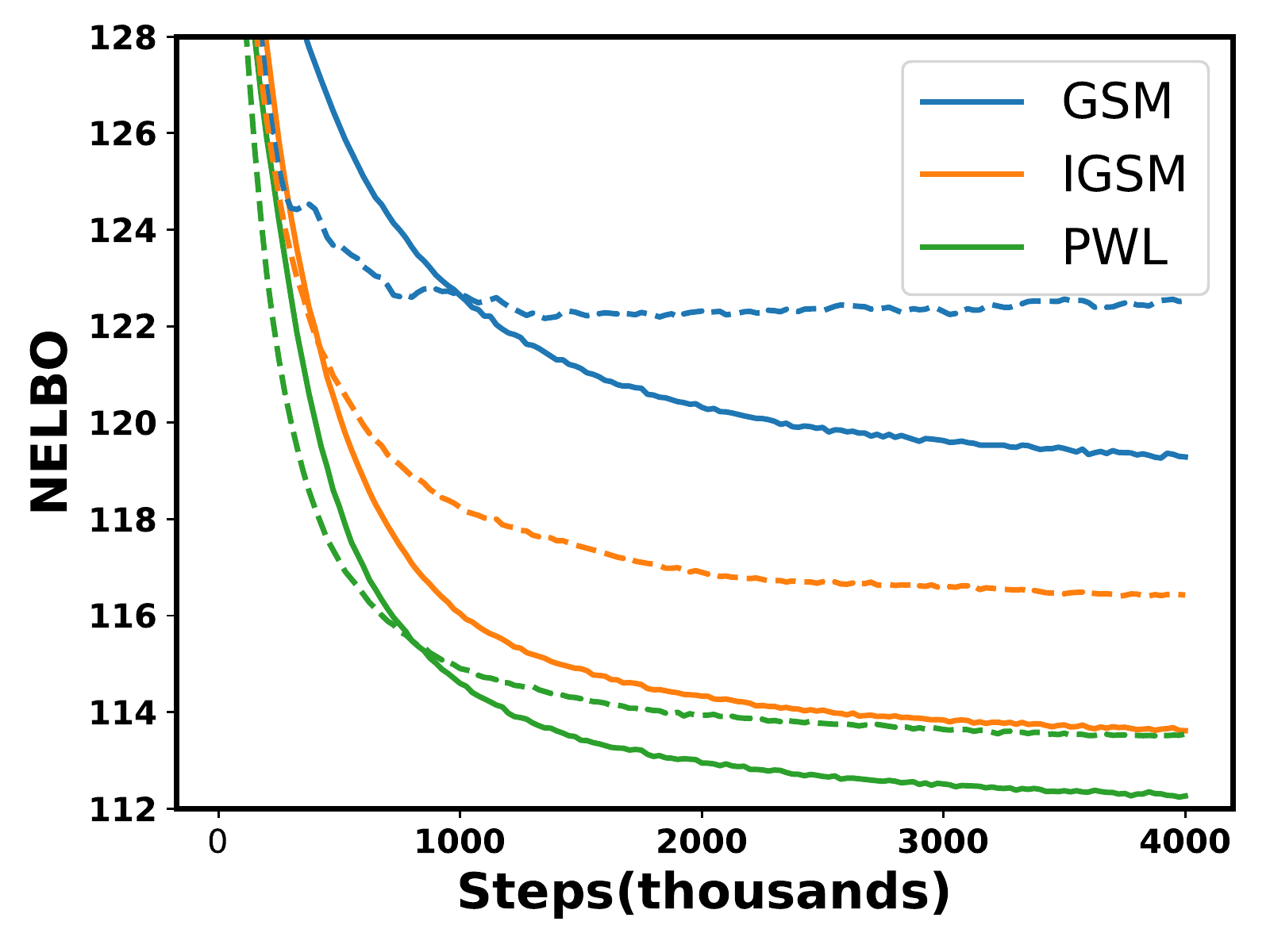}}
  \vspace{-0.4cm}
    \subfloat[$200{\rm H}\sim784{\rm V}$]{\includegraphics[scale=0.25]{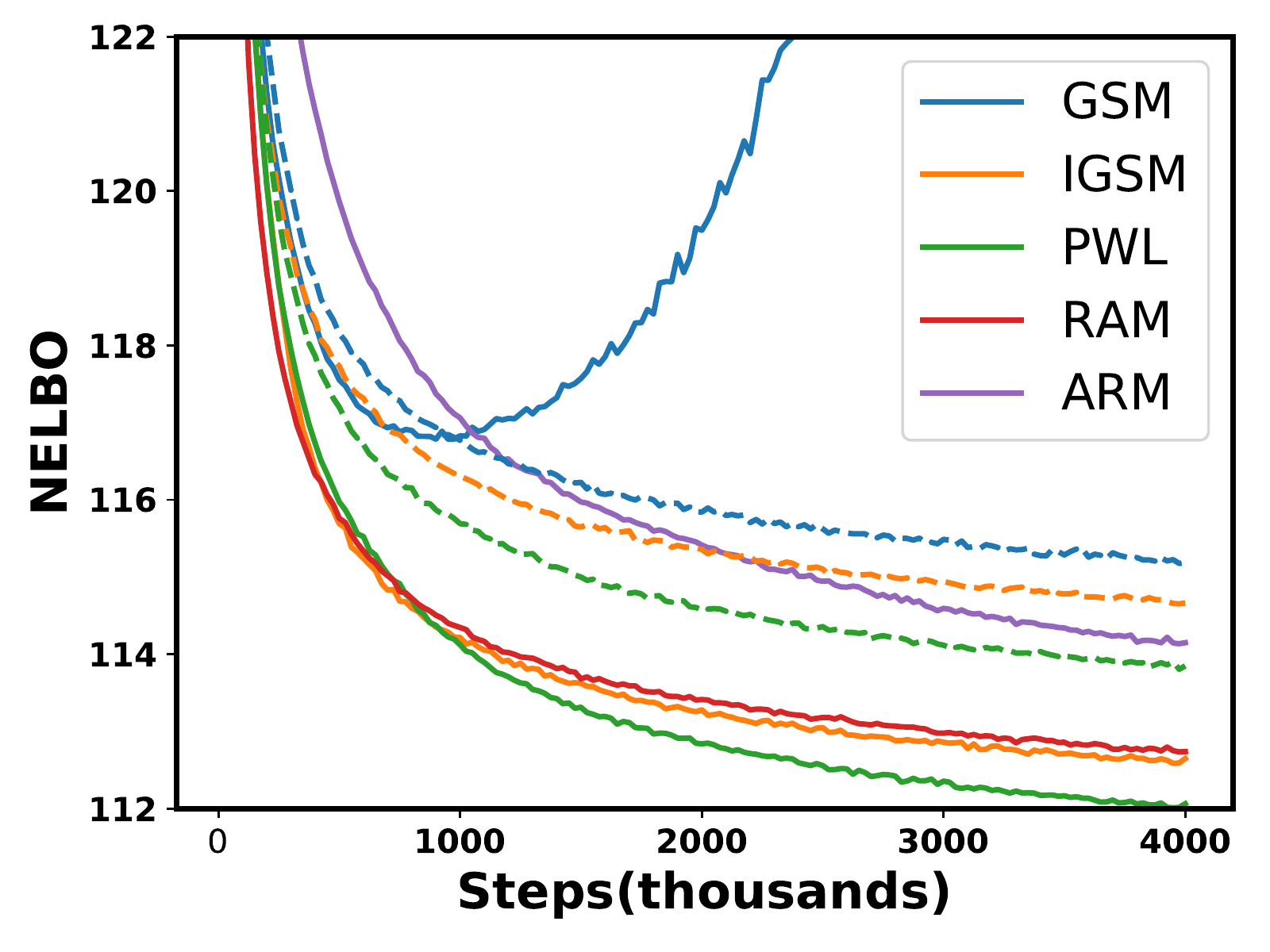}}
      \subfloat[$200{\rm H}\sim200{\rm H}\sim784{\rm V}$]{\includegraphics[scale=0.25]{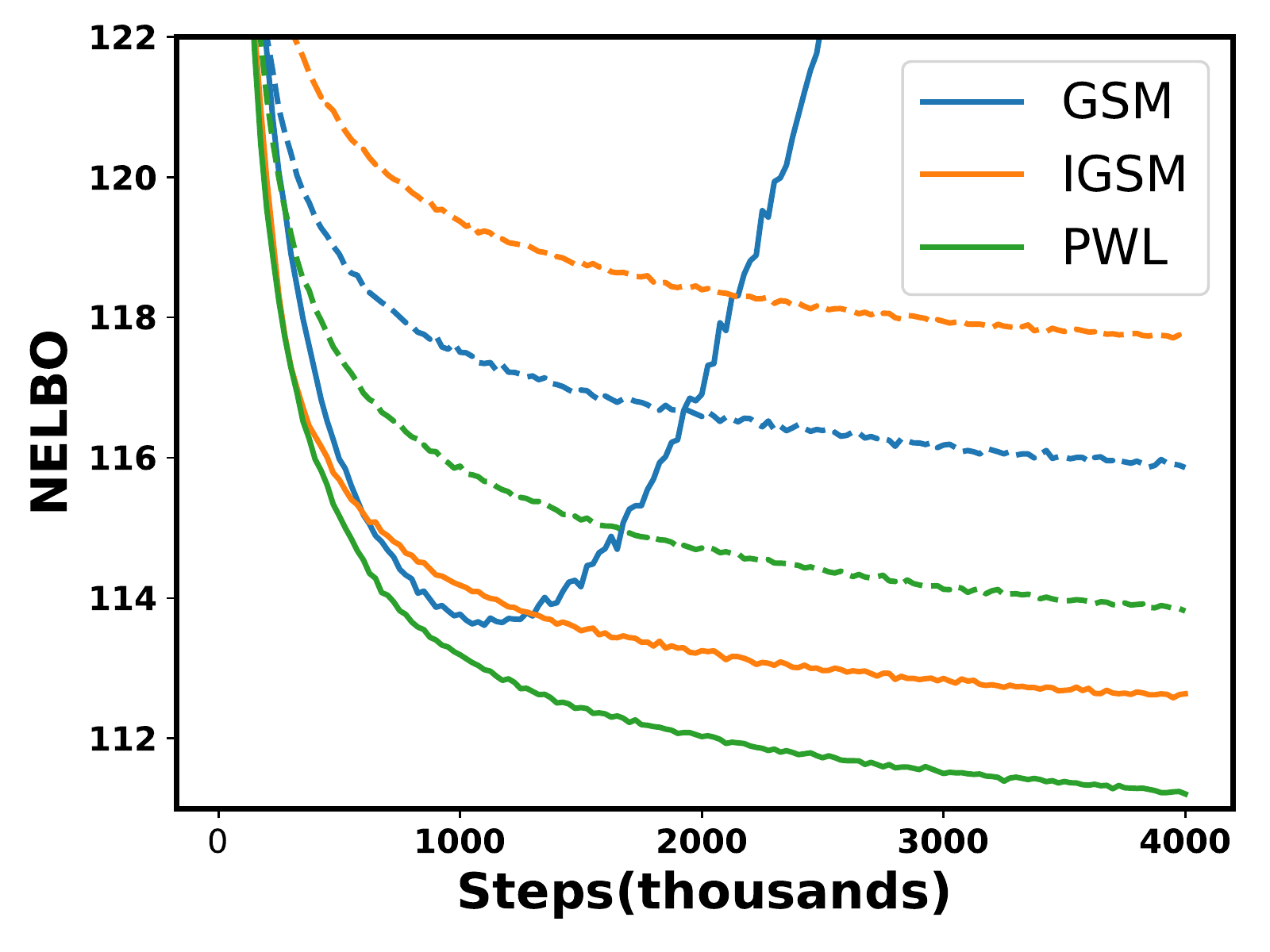}}
 \caption{OMNIGLOT training (CR estimators use $\beta=2$). Solid/dashed lines correspond to one/two-pass training.}
    \label{fig:bin_vae_elbo_omniglot_beta2}
\end{figure}

For completeness we also show the training curves for OMNIGLOT dataset at $\beta=2$ in Fig~\ref{fig:bin_vae_elbo_omniglot_beta2}. Interestingly, one-pass training performs better in all cases. PWL estimator performs best among considered CR estimators.

\end{document}